\documentclass[twoside,11pt]{article}

\usepackage{blindtext}

\usepackage[preprint]{jmlr2e}

\usepackage{lastpage}
\jmlrheading{23}{2022}{1-\pageref{LastPage}}{1/21; Revised 5/22}{9/22}{21-0000}{Author One and Author Two}

\firstpageno{1}

\usepackage{amssymb}            
\usepackage{mathtools}          
\usepackage{mathrsfs}           
\usepackage{graphicx}           
\usepackage{subcaption}         
\usepackage[space]{grffile}     
\usepackage{url}                
\usepackage{lipsum}             

\usepackage{booktabs}
\usepackage{cleveref}
\usepackage{longtable}
\usepackage{wrapfig}
\usepackage{placeins}

\DeclareMathOperator{\btheta}{\boldsymbol{\theta}}
\DeclareMathOperator{\bpsi}{\boldsymbol{\psi}}

\crefname{appendix}{appendix}{appendices}
\Crefname{appendix}{Appendix}{Appendices}

\begin{document}

\title{Performance Variation in Deep Reinforcement Learning}

\author{\name Haruto Tanaka \email haruto@ualberta.ca \\
      \addr Department of Computing Science\\
      University of Alberta\\
      Alberta Machine Intelligence Institute (Amii)
      \AND
      \name A. Rupam Mahmood \email armahmood@ualberta.ca \\
      \addr Department of Computing Science \\
      University of Alberta\\
      Alberta Machine Intelligence Institute (Amii)\\
      CIFAR AI Chair
      }

\editor{My editor}

\maketitle

\begin{abstract}
Deep reinforcement learning (RL) algorithms often suffer from low run-to-run robustness, manifesting as significant performance variation across independent runs of identically configured agents.
Although this issue poses a spectrum of challenges across research and practice, relatively few studies develop methods to evaluate it; RL research instead often reports uncertainty in the estimated mean performance.
In this paper, we outline the limitations of conventional uncertainty and variation estimates, particularly their misalignment with purpose and the risk of underreporting.
We then propose an alternative percentile-based statistic and visualization method, \textit{min-max IPR} and \textit{run-wise percentile highlighting}, respectively.
These percentile-based tools are easy to interpret and rely on standard properties of sample percentiles, providing rich information about run-to-run performance variation.
We demonstrate this through three case studies.
First, we show that LayerNorm and penultimate-layer normalizations narrow performance variation in PPO, whereas the variation is mostly unchanged in SAC.
Second, we compare PPO, SAC, TD-MPC, and TD-MPC2, and show TD-MPC exhibits the least variation while being the most data efficient among the four.
Finally, in a comparison of DQN and Rainbow on five Atari environments, we show that both algorithms exhibit similar levels of performance variation.
\footnote{The code and data are available at \url{https://github.com/WINUprj/eval-perf-variation}}
\end{abstract}


\section{Introduction}
\label{sec:introduction}

Although deep reinforcement learning (RL) algorithms are known to learn complex behaviors (e.g., \citealp{mnih_2015_HumanlevelControlDeep, bellemare_2020_balloon, wurman_2022_granturismo, haarnoja_2024_LearningAgileSoccer}), they are also notorious for performing differently with slight changes to their settings.
Here, we specifically consider online learning variants of deep RL algorithms and let \textit{performance} denote the average of online episodic returns across all episodes in a single run.
The spectrum of components can trigger performance differences, including the design of deep neural networks, independent runs, hyperparameter configurations, stochasticity in the environment or learning dynamics, hardware specifications \citep{hausknecht_2015_DeepRecurrentQLearning, henderson_2018_DRLMatters}, and delays in real-time learning systems \citep{mahmood_2018_setupRealRL}.

\begin{figure*}[t!]
    \centering
    \begin{minipage}[t]{0.45\textwidth}
        \centering
        \subcaption*{Supervised Learning}
        \begin{subfigure}[t]{0.48\textwidth}
            \centering
            \includegraphics[width=\textwidth, height=\textwidth]{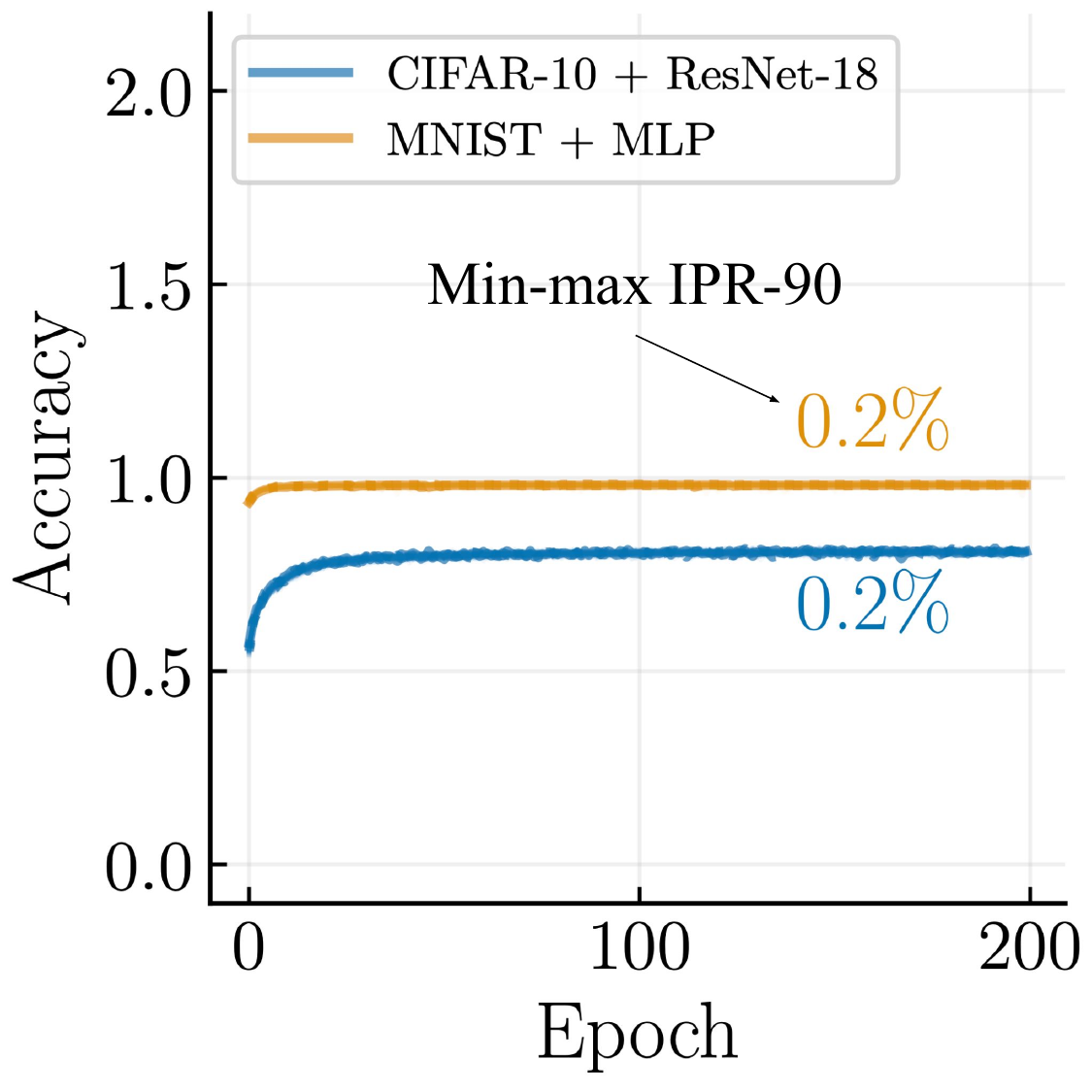}
            \caption{Test Accuracies}
            \label{fig:sl_varioation_main:sl_acc}
        \end{subfigure}\hfill
        \begin{subfigure}[t]{0.48\textwidth}
            \centering
            \includegraphics[width=\textwidth, height=\textwidth]{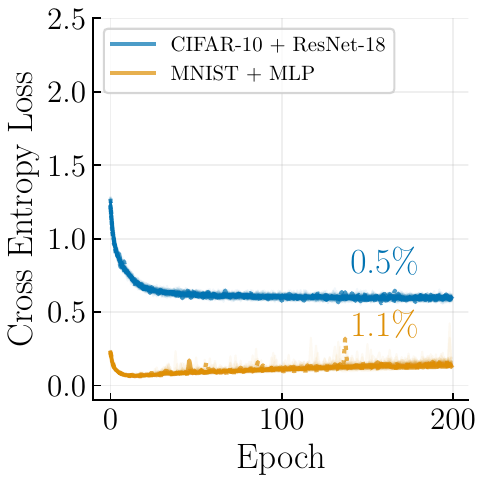}
            \caption{Test Losses}
            \label{fig:sl_variation_main:sl_loss}
        \end{subfigure}
    \end{minipage}\hspace{0.06\textwidth}
    \begin{minipage}[t]{0.45\textwidth}
        \centering
        \subcaption*{Reinforcement Learning}
        \begin{subfigure}[t]{0.48\textwidth}
            \centering
            \includegraphics[width=\textwidth, height=\textwidth]{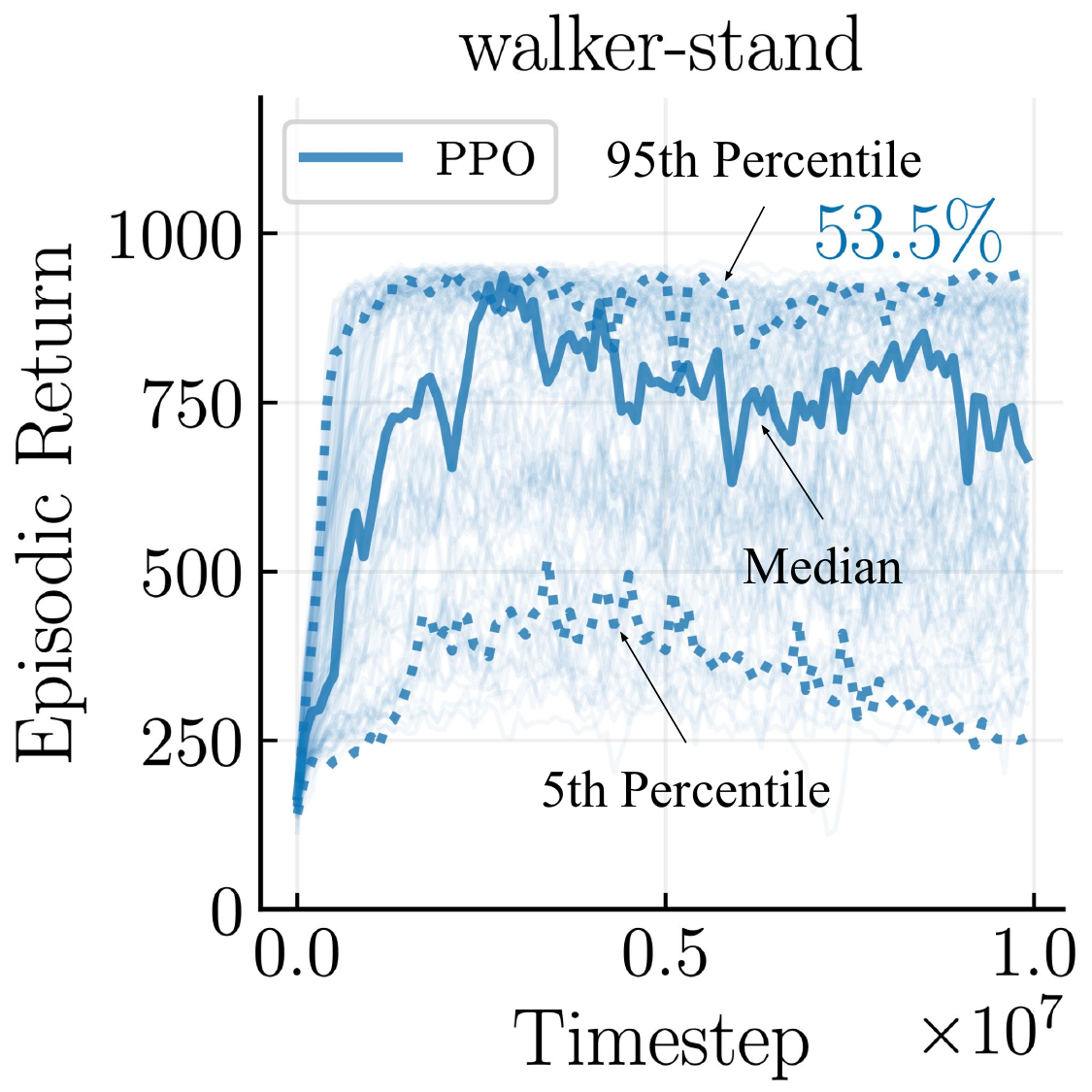}
            \caption{DMC}
            \label{fig:sl_variation_main:continuous}
        \end{subfigure}\hfill
        \begin{subfigure}[t]{0.48\textwidth}
            \centering
            \includegraphics[width=\textwidth, height=\textwidth]{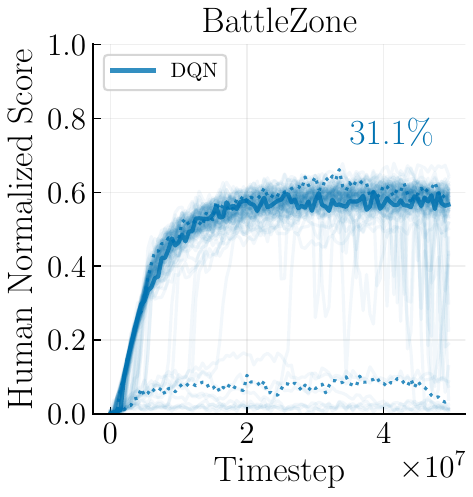}
            \caption{ALE}
            \label{fig:sl_variation_main:ale}
        \end{subfigure}
    \end{minipage}
    \caption
    {
        Variation of standard supervised learning (SL) (a \& b), continuous deep RL (c), and discrete deep RL (d) settings.
        First two plots presents \textbf{(a)} test accuracy and \textbf{(b)} test loss of \(100\) independent runs of MLP on MNIST and ResNet-\(18\) on CIFAR-\(10\).
        The step-size of Adam optimizer is decayed from \(3\times 10^{-4}\) to \(3\times 10^{-5}\) at the \(100\)th epoch.
        Third and fourth plots depict \textbf{(c)} episodic return of PPO on \texttt{pendulum-swingup} and \textbf{(d)} Human Normalized Score of DQN on \texttt{BattleZone}.
        The number beside each set of curves is min-max IPR-\(90\), our proposed measurement of spread (smaller the better. See \Cref{sec:measurement}).
        All curves are plotted with RPH (see \Cref{sec:rph}). 
        Curves related to SL settings exhibit lower performance variation than those for deep RL settings.
    }
    \label{fig:sl_variation_main}
\end{figure*}

In particular, the performance sensitivity across independent runs profoundly hinders progress in deep RL, both in research and in practical applications.
Within the context of research, such brittleness causes difficulty in reproducing results \citep{islam_2017_benchmarkedDRL}, fair comparison between algorithms \citep{clary_2019_variabilityOfDRL}, and hyperparameter tuning \citep{eimer_2023_hyperparameterInRL, hertel_2020_qVsQ}.
The current best practice to ensure rigor in these aspects is to conduct a sufficiently large number of independent trials \citep{colas_2018_howManySeed, eggensperger_2019_pitfalls, patterson_2024_EmpiricalDesignReinforcement}.
However, such a procedure demands substantial computational resources, creating a high barrier to producing scientifically sound evidence from large-scale experiments.
The lottery-like behavior over independent runs also severely undermines the practicality of deep RL-driven systems in real-world tasks.
Ineffective behaviors are not only futile but can also pose a safety concern if the task requires stringent safety measures.
Such real-world risk is one of the primary reasons why performance variability in online deep RL is important in comparison to, for example, the variability of training results in supervised learning (SL) or offline RL.
SL training is often performed offline, effectively minimizing the risks associated with poor training outcomes.
Additionally, SL tends to exhibit smaller variations, which reduces the importance of investigating its performance variability across runs, as we show in our results in \Cref{fig:sl_variation_main} (see \Cref{sec:appendix:sl_learning_curves} for more examples and experiment details).
Nevertheless, performance sensitivity across independent runs on a single task, which we refer to as \textit{performance variation}, warrants further investigation for the development of deep RL algorithms.

Despite its importance, performance variation is often overlooked in empirical deep RL studies.
This is attributed to the focus of many recent works on overall improvements in the aggregated performance of proposed methods relative to baselines.
Such a trend encourages many works to merely report uncertainty in aggregated performance across multiple tasks.
One of the most popular uncertainty measures is the bootstrapped confidence interval of the interquartile mean (IQM) \citet{agarwal_2021_statisticalPrecipice}.
Although similar in appearance, confidence intervals and other uncertainty estimates are not measures of variation (\Cref{sec:std_error}).
These also appear in a rigorous comparison of algorithm performance over many tasks.
As a result, many empirical deep RL studies neglect the performance variation in a single task.
Some work reports standard deviation as a measure of performance variation (e.g., \citealp{liang_2016_sotaAtari,bjorck_2022_high_variance}).
However, we argue that reporting data variation with standard deviation entails the risk of underreporting and yields an inaccurate summary of the data due to the specific characteristics of the performance distributions of control policies learned by modern deep RL algorithms (\Cref{sec:limit_of_std}).
Furthermore, statistically rigorous methods have been proposed to measure performance variation, such as performance profiles and the tolerance intervals (TIs)  \citep{agarwal_2021_statisticalPrecipice,patterson_2024_EmpiricalDesignReinforcement}.
While statistically rigorous methods are often robust and accurate, they are also often costly and difficult to interpret (\Cref{sec:measurement}).
Therefore, there is a need for evaluation tools that accurately capture variability of performance on a single task, while remaining easily interpretable.

In this paper, we propose quantification and visualization methods that achieve an appropriate balance between accuracy and interpretability in capturing performance variation in deep RL.
Particularly, we claim that \textit{min-max IPR-\(90\)}, a min-max normalized interpercentile range (IPR) from \(5\)th to \(95\)th percentiles, is a practical and reliable quantification of performance variation.
We discuss that min-max IPR-\(90\) captures performance variation more robustly than the standard deviation, while it is more interpretable than rigorous options, such as TI (\Cref{sec:measurement}).
In parallel, we also propose a learning curve visualization method called \textit{run-wise percentile highlighting} (RPH).
The core idea of this visualization technique is to highlight the individual learning curves corresponding to the \(5\)th, \(50\)th, and \(95\)th percentiles of performance.
We show that RPH further clarifies how each individual learning curve behaves and allows investigators to readily examine performance variability across runs (\Cref{sec:rph}).
Then, we demonstrate the use cases of min-max IPR-\(90\) and RPH on three case studies (\Cref{sec:case_studies}).
In the first case study, we analyze changes in performance variation after applying LayerNorm, penultimate layer normalization, or both to PPO and SAC \citep{bjorck_2022_high_variance}.
Using our proposed methods, we show that normalization techniques narrow performance variation in PPO, whereas it remains mostly unchanged in SAC. 
In the second case study, we use all DeepMind Control Suite (DMC) tasks to systematically compare four deep RL algorithms: PPO, SAC, TD-MPC, and TD-MPC2 \citep{schulman_2017_ppo,haarnoja_2018_sac,hansen_2022_TDMPC,hansen_2024_TDMPC2}.
Through comparisons using our proposed methods, we find that TD-MPC exhibits the lowest performance variation and the highest data efficiency.
The third case study repeats the same process as the second case study for two discrete control algorithms: DQN and Rainbow \citep{mnih_2015_HumanlevelControlDeep,hessel_2018_RainbowCombiningImprovements} on Atari-5 tasks \citep{aitchison_2023_Atari5DistillingArcade}.
With our methods, we find that both algorithms exhibit significant performance variation in some tasks.

\section{Experiment Settings}
\label{sec:exp_settings}

To illustrate and examine the performance variation issue, we mainly use PPO and SAC algorithms on two robotic control suites \citep{schulman_2017_ppo, haarnoja_2018_sac}.
For both algorithms, our implementation is based on CleanRL \citep{huang_2022_cleanrl}.
We use \(59\) task environments selected from Gymnasium's MuJoCo environments and the DeepMind Control Suite (DMC) as the testbed \citep{todorov_2012_mujoco, tassa_2018_dmc, towers_2025_Gymnasium}.
Specifically, we choose \(11\) tasks from MuJoCo and \(48\) tasks from DMC.
The time limit per episode for all tasks is $1000$ steps.
In addition to the time limit, each MuJoCo environment has its own termination condition.
The details of the tasks used in this paper are summarized in \Cref{table:rl_tasks}.
Note that we do not use parallelized environments, as is sometimes done (e.g., \citealt{stooke_2019_AcceleratedMethodsDeepRL, li_2023_PQL,lee_2025_simba}) to better align with real-world RL settings.
For each task, we run \(100\) independent runs with different random seeds.
Each run lasts \(10\) million environment steps for PPO and \(1\) million environment steps for SAC.
Other hyperparameters of the algorithms are given in \Cref{table:ppo_configs,table:sac_configs} in \Cref{sec:appendix:hp}.
All learning curves are binned for visual clarity (see \Cref{sec:appendix:binning} for the details).
Visualizations of all binned learning curves for PPO and SAC are provided in \Cref{sec:appendix:learning_curves}.

\begin{figure*}[t!]
    \centering
    \begin{subfigure}[b]{0.32\textwidth}
        \centering
        \includegraphics[width=\textwidth]{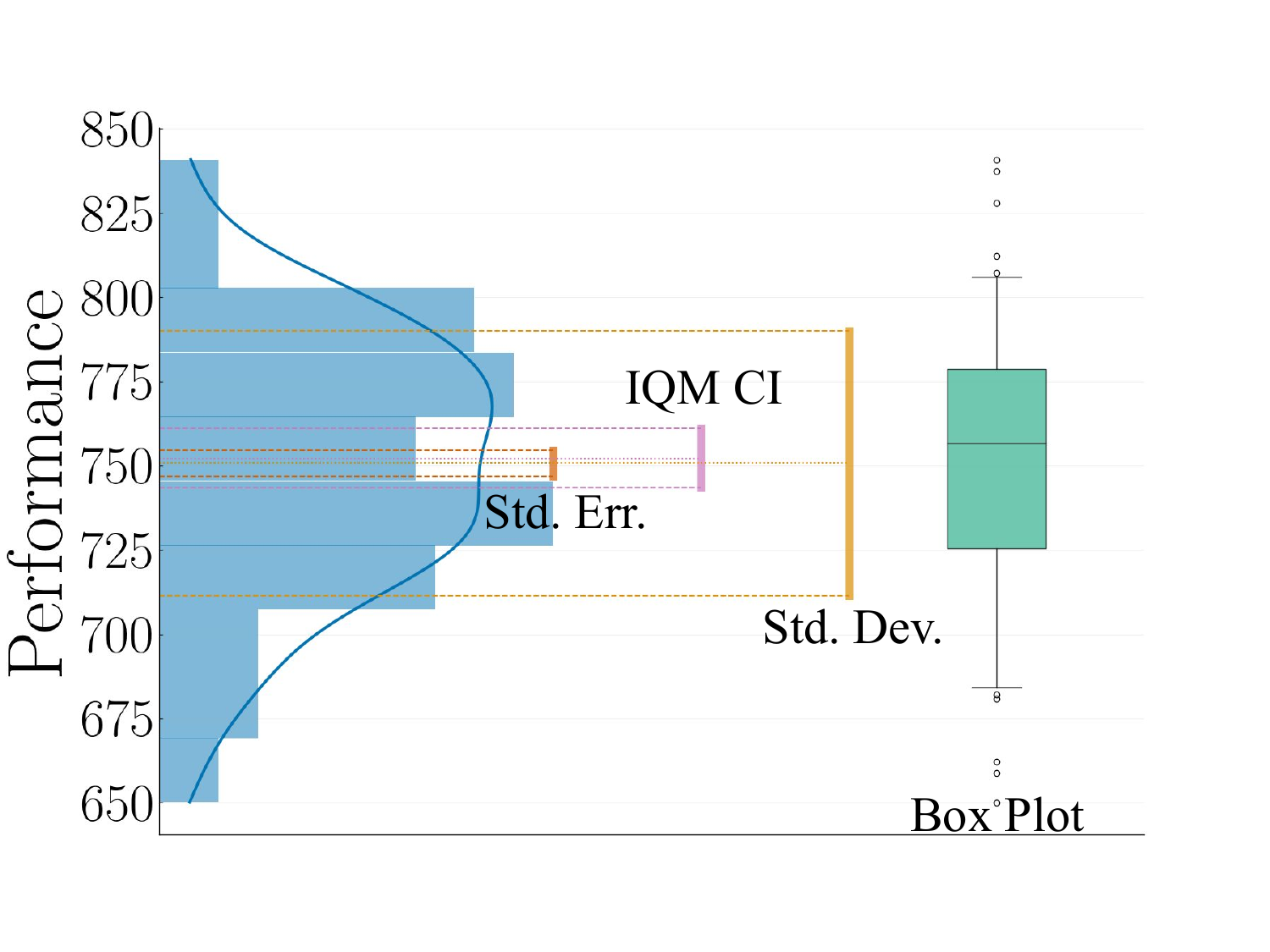}
        \caption{\texttt{reacher-hard}}
        \label{fig:perf_dist_var:reacher}
    \end{subfigure}
    \hfill
    \begin{subfigure}[b]{0.32\textwidth}
        \centering
        \includegraphics[width=\textwidth]{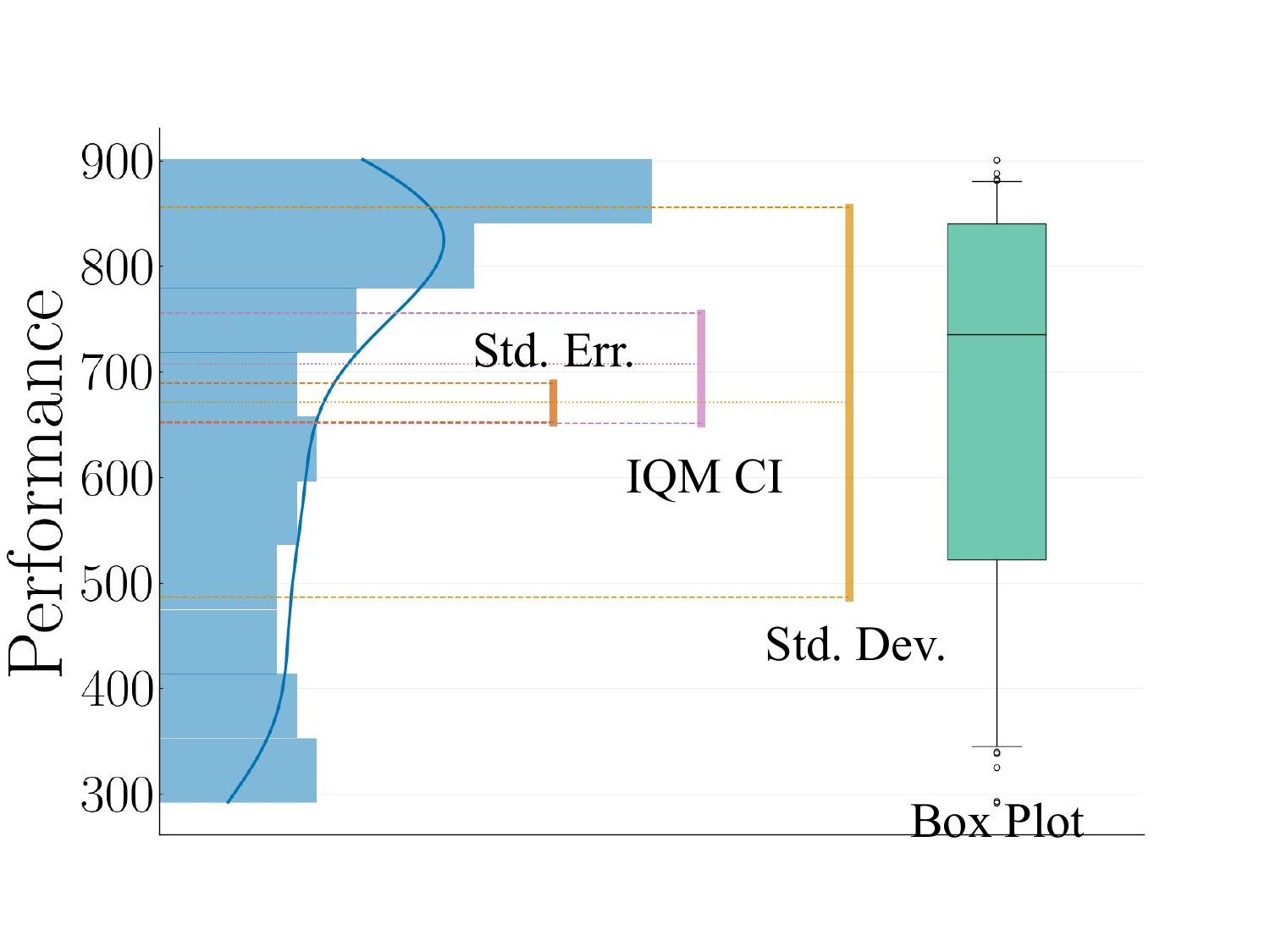}
        \caption{\texttt{walker-stand}}
        \label{fig:perf_dist_var:walker}
    \end{subfigure}
    \hfill
    \begin{subfigure}[b]{0.32\textwidth}
        \centering
        \includegraphics[width=\textwidth]{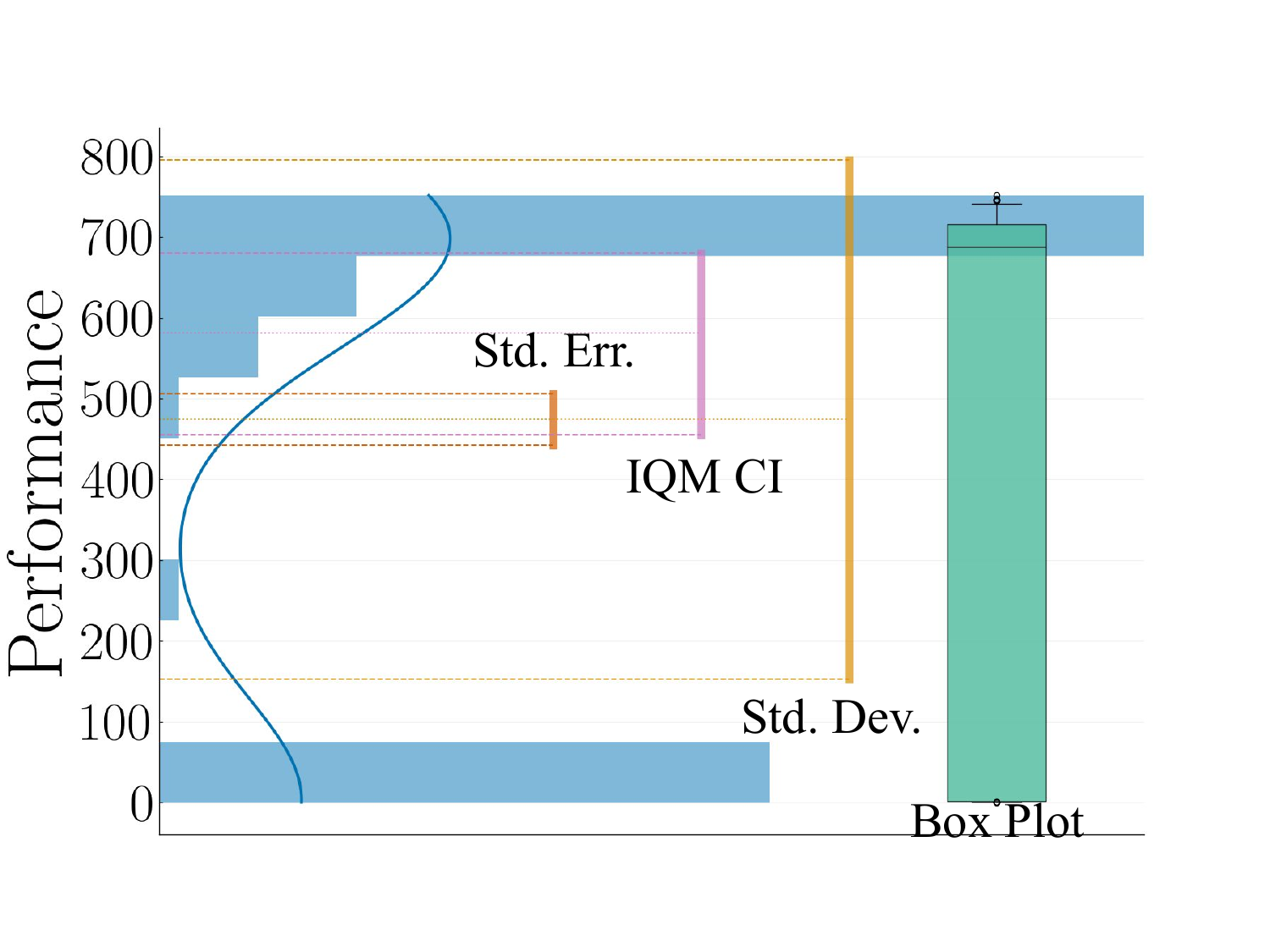}
        \caption{\texttt{pendulum-swingup}}
        \label{fig:perf_dist_var:pendulum}
    \end{subfigure}

    \caption[Visualization of performance distribution and variations on exemplar learning curves from default PPO experiments.]
    {
        Visualization of performance distribution and variations of some PPO experiments.
        Each performance distribution exhibits roughly Gaussian, unimodal-skewed, and bimodal shapes.
        Vertical red, pink, and orange lines represent the range where the average performance \(\pm\) standard error, stratified bootstrapped 95\(\%\) CI of IQM, and the average performance \(\pm\) standard deviation cover, respectively.
        The green boxplots depict the IQR, and the whiskers represent the range between \(5\)th and \(95\)th percentiles.
        The boxplots robustly cover most of the data range, unlike the other options.
    }
    \label{fig:perf_dist_var}
\end{figure*}

\section{Uncertainty Estimates Do Not Capture Performance Variation}
\label{sec:std_error}

A large volume of deep RL studies focus on the aggregated performance of their proposed algorithm relative to a baseline.
This naturally led many RL studies to report the uncertainty estimates of the aggregated performance.
For instance, confidence intervals or standard errors are often utilized \citep{henderson_2018_DRLMatters, agarwal_2021_statisticalPrecipice, tang_2024_Churn}.
Despite their popularity, these uncertainty estimates are not suitable for capturing performance variation.

Intuitively, uncertainty estimates reflect the probable discrepancy between the sample and the true statistics.
For example, the \(95\%\) stratified bootstrap confidence interval for IQM by \citet{agarwal_2021_statisticalPrecipice} is an interval estimate of the range within which the population IQM lies with a certain confidence.
Although these values provide additional information about the performance sample statistics, they do not reflect the spread of the performance distribution itself.
Thus, by design, the uncertainty estimates do not reflect the variation.

Furthermore, uncertainty estimates vanish as the number of independent runs increases.
For instance, standard error eventually converges to zero at a rate of \(\mathcal{O}(1/\sqrt{n})\).
This is an undesirable property as a measure of variation because a degree of spread in data is independent of the number of data points.
Also, due to this property, uncertainty estimators often mark values smaller than variation measurements, as shown in \Cref{fig:perf_dist_var}.
Each solid red, pink, and orange vertical line represents the range covered by the standard error, the \(95\%\) stratified bootstrap confidence interval of the IQM, and the standard deviation, respectively.
Visually, the standard error and confidence interval merely cover a small subinterval of the range covered by the standard deviation.
In contrast, the standard deviation covers a relatively wide range of performance distributions.
Henceforth, reporting uncertainty as performance variation is a misuse and risks obscuring high performance variation.

\section{Limitations of Standard Deviation as a Measurement of Performance Variation}
\label{sec:limit_of_std}

The standard deviation is a popular measure of the variability of performance in a single task (e.g., \citealt{liang_2016_sotaAtari,bjorck_2022_high_variance}).
However, deep RL returns often have distributional features that can make the standard deviation a misleading summary of performance variation.
In this section, we analyze its sensitivity to return distribution and show how it can understate risk in common deep RL evaluation settings.

\begin{figure*}[htb!]
    \centering
    \begin{subfigure}[b]{0.44\textwidth}
        \centering
        \includegraphics[width=\textwidth]{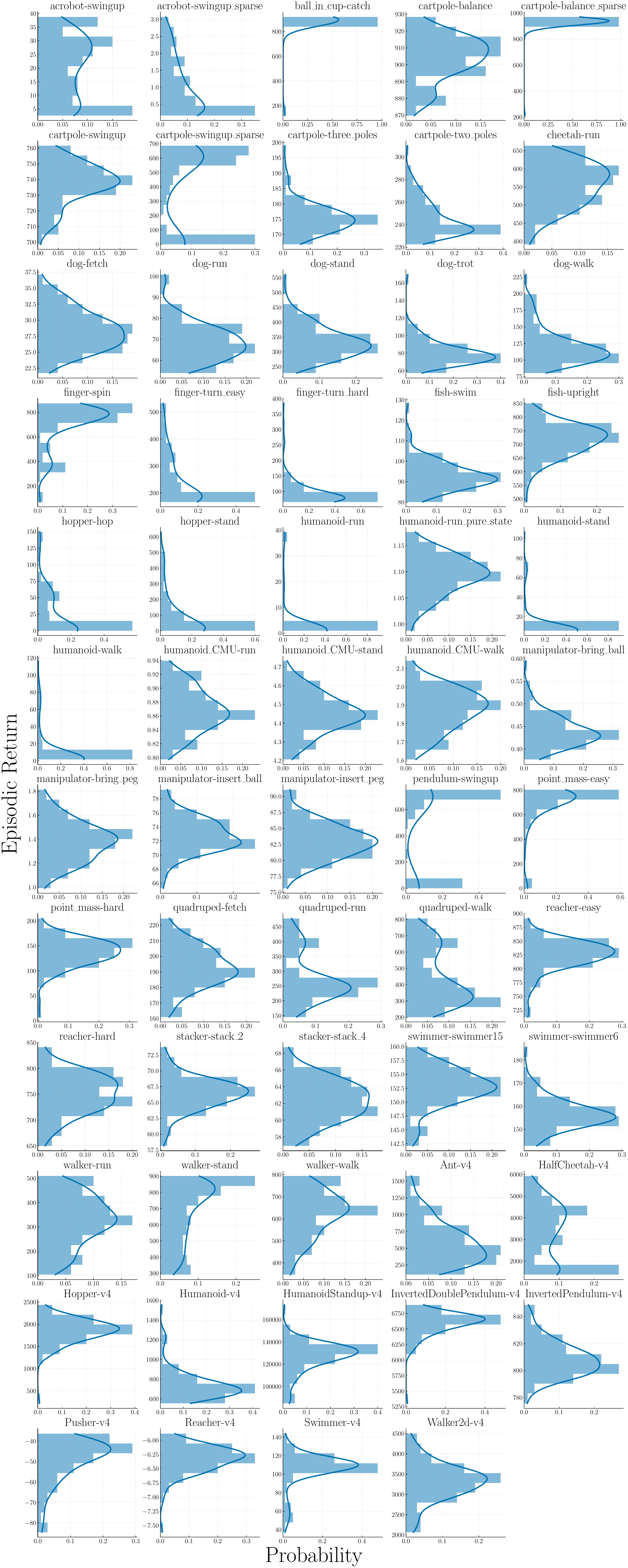}
        \caption{PPO}
        \label{fig:baseline_perf_summary:ppo:dist}
    \end{subfigure}\hspace{0.05\textwidth}
    \begin{subfigure}[b]{0.44\textwidth}
        \centering
        \includegraphics[width=\textwidth]{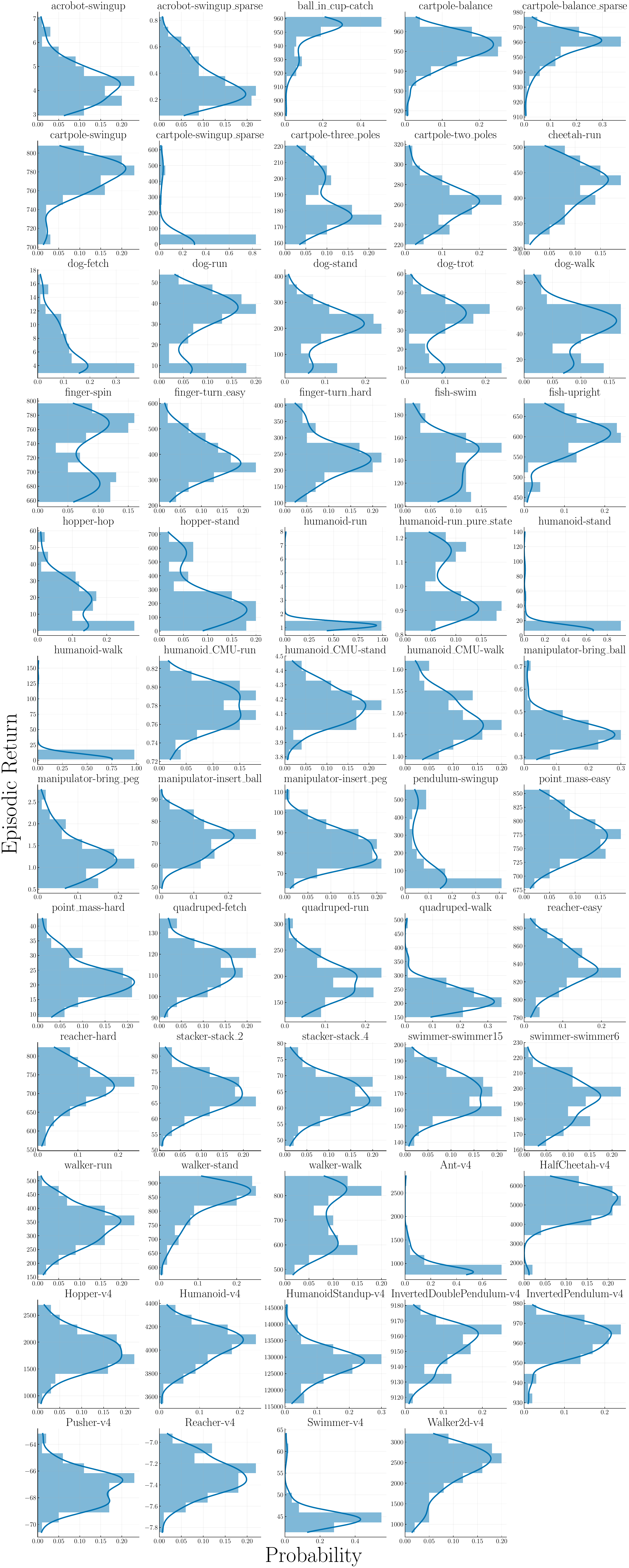}
        \caption{SAC}
        \label{fig:baseline_perf_summary:sac:dist}
    \end{subfigure}

    \caption
    {
        Histogram and kernel density estimations (KDEs) of performance distribution for PPO and SAC experiments with a default configuration.
        Each subplot (the small boxes) corresponds to a control task.
        Results are obtained by running \(100\) independent runs, and the number of histogram bins is \(10\).
        For each KDE, a Gaussian KDE is employed with Scott's rule for bandwidth selection \citep{scott_1992_multivariate}.
        Different instance of experiments exhibits different performance distributions, many of which are not necessarily Gaussian.
    }
    \label{fig:baseline_perf_summary}
\end{figure*}

\subsubsection*{Performance Distribution is Often Not Gaussian}
\label{sec:limit_of_std:insights}
The standard deviation provides comprehensive information about the spread of samples when the distribution is close to Gaussian \citep{altman_2005_StandardDA, campbell_2021_StatAtSquareOne, green_2026_meansAndStd}.
However, in non-Gaussian distributions, it may provide misleading information.
Thus, it is important to verify that the data roughly conform to the Gaussian assumption when applying the standard deviation.

Our empirical results show that the Gaussian assumption is hardly satisfied by the PPO and SAC algorithms.
\Cref{fig:baseline_perf_summary} depicts the histograms and kernel density estimation (KDE) of the performance from the PPO and SAC experiments described in \Cref{sec:exp_settings}.
Visually, performance distributions are either \((1)\)\ roughly Gaussian, \((2)\)\ unimodal but skewed, or \((3)\)\ bimodal.
Visual taxonomization already suggests the existence of non-Gaussian performances.
To solidify this discovery, we conduct a Shapiro-Wilk test on each sample distribution, which is the hypothesis test of normality \citep{shapiro_1965_ShapiroWilkTest}.
The null hypothesis of the Shapiro-Wilk test states that the underlying distribution of the given data points is normal.
With the significance level of \(\alpha = 0.05\), we find that only about \(32\%\) of the results shown in \Cref{fig:baseline_perf_summary:ppo:dist} and \Cref{fig:baseline_perf_summary:sac:dist} have no significant evidence to be classified as a non-Gaussian distribution.
In addition to these insights, \citet{mathieu_2024_Adastop} also notes the non-Gaussianity of the performance distribution.
Overall, the performance distribution does not conform to the Gaussian assumption, making the standard deviation an insufficient measure of variation.

\subsubsection*{Risk of Underreporting}
\label{sec:limit_of_std:underreporting}
Standard deviation is also prone to underreporting, making it insufficient for measuring variation.
This is illustrated in \Cref{fig:perf_dist_var}, which shows three performance distributions of PPO.
The orange vertical lines show the range of the sample mean \(\pm\) sample standard deviation.
The underreporting by standard deviation is prominent in \Cref{fig:perf_dist_var:pendulum}.
There, the standard deviation does not cover the mode lying around the performance of \(0\).
This is highly problematic because it implies that the standard deviation does not capture \(30\%\) failed PPO runs in \texttt{pendulum-swingup} task.
Underreporting can also be seen in \Cref{fig:perf_dist_var:walker}. 
In \Cref{fig:perf_dist_var:walker}, the range covered by the standard deviation is shifted towards the mode of distribution, resulting in insufficient coverage over the heavy tail of the distribution.
Therefore, due to the high risk of underreporting, the standard deviation is not suitable to capture performance variation.

\section{Inter-percentile Range as a Measurement of Performance Variation}
\label{sec:measurement}

Nonparametric statistics are preferred when it is infeasible to make distributional assumptions.
Here, we consider using percentiles as a measure of performance variation in a single task.
We first discuss why percentile-based statistics could be a suitable choice.
Then, we discuss how the existing percentile-based method, the tolerance interval (TI), can be further simplified.
Lastly, we formalize a performance variation metric: min-max normalized inter-percentile range (IPR).

\subsubsection*{Suitability of Percentile}
\label{sec:measurement:percentile}
An alternative measure of variation to standard deviation is range estimation using percentiles.
The advantage of percentiles is that they do not require assumptions about the data distribution.
Hence, it often captures the characteristics of different distributions more robustly.
For instance, the box plots in \Cref{fig:perf_dist_var} describe the essential features of the performance distributions, including symmetry, skewness, and variability.
The robust ability to capture data variability makes percentile-based variation measurement well-suited for our study.

Furthermore, the sample percentile is known to be a consistent estimator of the true population percentile.
One reason for the frequent adoption of the sample mean and standard deviation is that they are asymptotically guaranteed to match their population values.
This shared theoretical property further supports IPR as a measure of performance spread (see \Cref{sec:appendix:os_consistency} for mathematical details).

\subsubsection*{Tolerance Interval and Performance Profile}
\label{sec:measurement:ti}

Although IPR appears to be a favorable option, choosing a suitable percentile range is challenging.
This is because the choice of the range essentially decides the portion of outliers that do not contribute to representing the performance variation.
As a solution to this problem, \citet{patterson_2024_EmpiricalDesignReinforcement} suggests using a TI.
Formally, \((\alpha, \beta)\)-TI is an interval that captures the center \(\beta\) proportion of the population with a nominal error of \(\alpha\).
In other words, TI takes an IPR that covers a center \(\beta \times 100\%\) proportion of the distribution and expands that range based on the number of data points.
If the number of data points is small, the TI broadens further.
In contrast, if the number of data points is sufficiently large, the TI coincides with IPR-\((\beta \times 100)\).
In this way, TI aims to estimate the population IPR-\((\beta \times 100)\).
While TI rigorously reasons about the population IPR-\((\beta \times 100)\), this rigor adds an additional layer of complexity that may not be necessary only to capture performance variation.
In the worst case, such complexity can reduce the interpretability of the results.
Hence, a simple IPR that covers a sufficiently large portion of the data is well-suited for analyzing performance variability.

The issue of choosing the IPR range can also be avoided by plotting the entire empirical distribution function, namely a performance profile, as proposed by \citet{agarwal_2021_statisticalPrecipice}.
The performance profile provides a comprehensive visual of the performance variation at different IPRs.
Although it provides a comprehensive view, the method is specifically designed for cross-algorithm comparisons.
In other words, it primarily captures how the performance of different deep RL algorithms are distributed when deployed over multiple tasks.
This fails to capture our desired information on performance variation in a single task.
Indeed, it is possible to restrict to a single task by plotting a performance profile on a per-task basis.
However, since such a process generates a cumulative distribution curve per algorithm-task pair, it lacks scalability against the number of algorithms and tasks.
For example, in our PPO and SAC experiments, we run both algorithms on \(59\) tasks.
In our case, the analysis of the performance profile per task results in investigation of \(108\) plots, which is laborious and thus lacks scalability.
In contrast, IPR provides a representative summary of the performance profile for each algorithm-task pair with a single quantity, which is easy to interpret.
Overall, in a single-task analysis, the IPR is a more suitable option than the performance profile.

\subsubsection*{Formalization of Performance Variation as Min-max Normalized IPR}
\label{sec:measurement:mmipr90}

As a simple variation metric, we propose a min-max normalized IPR of the given data.
We normalize IPR using min-max normalization to allow the direct comparison of variation across different algorithm-task pairs.
For all DMC tasks, the min-max values are \(0\) to \(1000\) by their reward design and termination condition.
In contrast, estimating episodic returns analytically for MuJoCo tasks is difficult.
Instead of analytically deriving them, we set the min–max values to the minimum and maximum episodic returns observed across all experiments in this paper.
See \Cref{sec:appendix:env_spec} for a complete list of min-max values.
Min-max normalized IPR can be mathematically formalized as
\begin{align*}
    \text{Min-max Normalized IPR}(\mathcal{U}_{T}, X) := \frac{\mathcal{U}_T^{\left(50 + \frac{X}{2}\right)} - \mathcal{U}_T^{\left(50 - \frac{X}{2}\right)}}{M_{T} - m_{T}} \text{ } (\%),
\end{align*}
where \(\mathcal{U}_T\) is the set of performances on task \(T\), \(\mathcal{U}_T^{(x)}\) is the \(x\)-th percentile value in \(\mathcal{U}_T\), \(X \in [0, 100]\) is the range of central region of data to cover, \(M_{T}\) (\(m_{T}\)) is the theoretical/empirical maximum (minimum) performance of task \(T\).

We choose to cover the central \(90\%\) portion, since it allows us to use IPR for the risk assessment of the algorithm.
Although our primary goal is to estimate performance variation, one underlying motivation for variation estimation is to assess the risk of using a deep RL algorithm, as outlined in \Cref{sec:introduction}.
To achieve this purpose, the min-max normalized IPR must cover a sufficiently wide range of performance distributions.
On the other hand, it is undesirable for IPR to rely on extremes, as a single run can drastically distort the resulting quantity.
Such high sensitivity undermines the reliability of the resulting statistics as a measure of spread.
Coverage of central \(90\%\) is a result of taking a trade-off between these two contradictory perspectives.
We refer to this specific variant of min-max normalized IPR as a \textit{min-max IPR-\(90\)}.
Note that we omit the term ``normalized" for conciseness.
As a heuristic, we choose \(5\%\) as a reasonable value of min-max IPR-\(90\), which is the maximum min-max IPR-\(90\) value in our SL experiments (see \Cref{sec:appendix:sl_learning_curves}).

\begin{figure*}[htb!]
    \centering
    \begin{subfigure}[b]{0.495\textwidth}
        \centering
        \includegraphics[width=\textwidth]{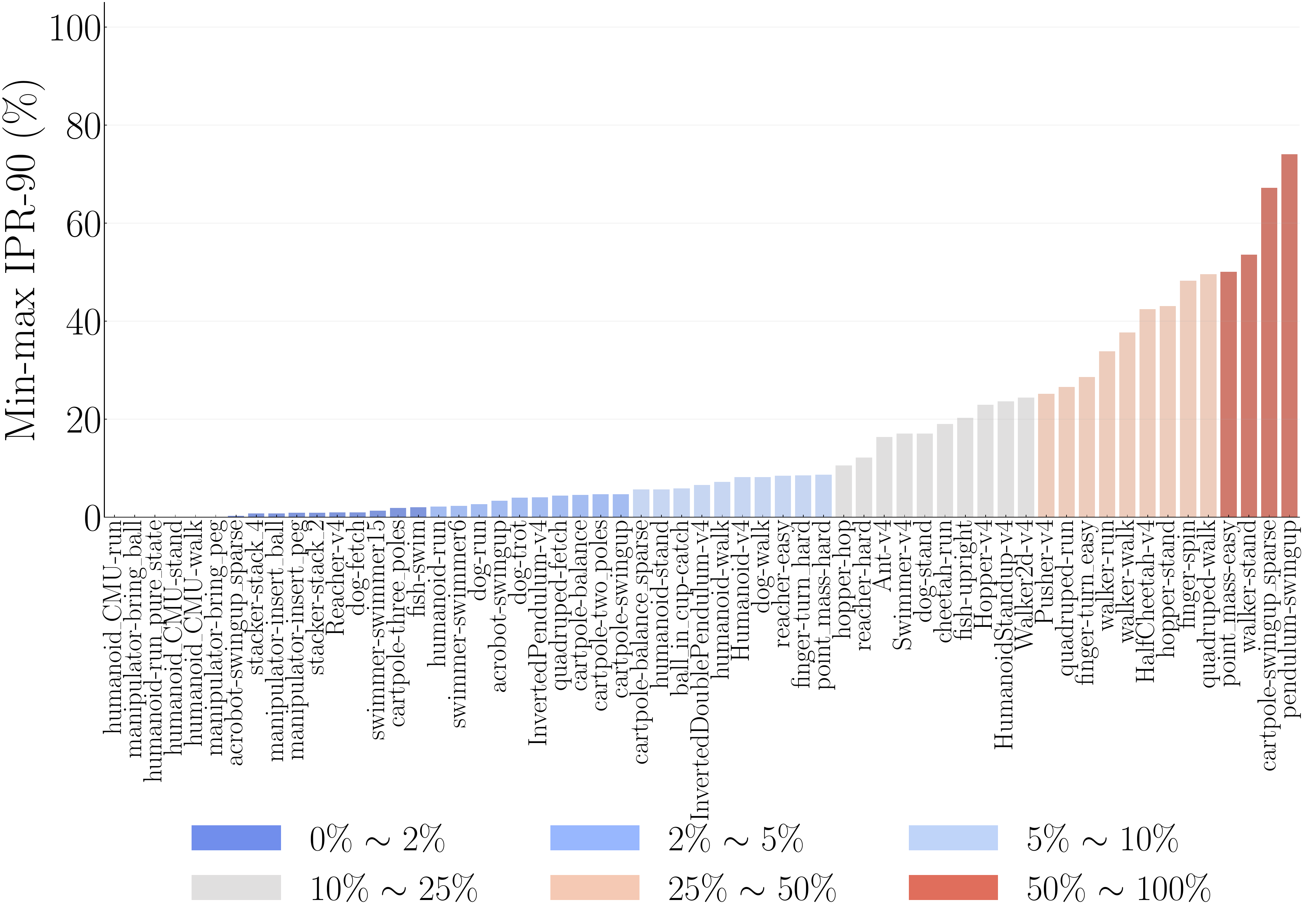}
        \caption{PPO IPR-\(90\)}
        \label{fig:perf_var_default:ppo_ipr}
    \end{subfigure}\hfill
    \begin{subfigure}[b]{0.495\textwidth}
        \centering
        \includegraphics[width=\textwidth]{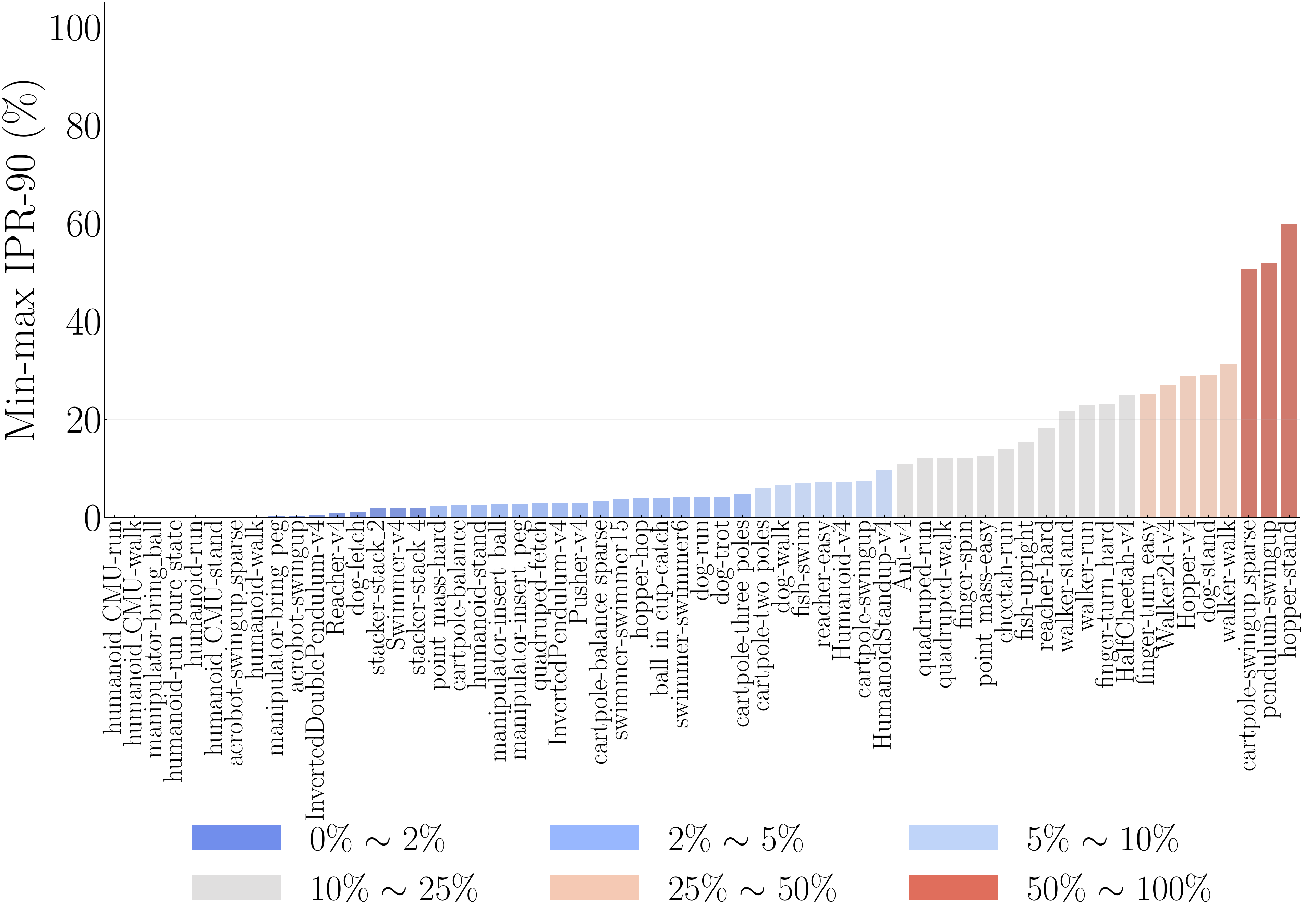}
        \caption{SAC IPR-\(90\)}
        \label{fig:perf_var_default:sac_ipr}
    \end{subfigure}
    \caption
    {
        Task-wise min-max IPR-\(90\) and median for the PPO and SAC experiments.
        In each plot, the bars are sorted by IPR-\(90\), and the units are \(\%\).
        The color of each bar represents non-overlapping range of percentages with cut off points of \(2\%\), \(5\%\), \(10\%\), \(25\%\), and \(50\%\).
        Warmer colors represent the higher variation, and vice versa.
        Both PPO and SAC exhibit high performance variation on some tasks.
        Moreover, SAC generally shows lower performance variation than PPO over the \(59\) robotic control tasks.
    }
    \label{fig:perf_var_default}
\end{figure*}

Despite its simplicity, min-max IPR-\(90\) provides clear insights into the performance variation.
For example, the bar plots in \Cref{fig:perf_var_default} show the min–max IPR-\(90\) of PPO and SAC for each task, from which we can readily see that SAC exhibits lower performance variation than PPO.

\section{Reporting Learning Curve Variability Time-wise vs.\ Run-wise}
\label{sec:rph}
Visualization of learning curves is often achieved by plotting a representative statistic aggregated across runs at each timestep, with a shaded region indicating the associated measure of variability or uncertainty.
Here, we emphasize that this time-wise learning curve aggregation is not necessarily suitable for visualizing performance variation.
We then propose a visualization method that takes a run-wise perspective and demonstrate its suitability for presenting performance variation.

\subsubsection*{Limitations of Time-wise Approach}
\label{sec:rph:limitation}

\begin{figure*}[t!]
    \centering
    \includegraphics[width=0.95\textwidth]{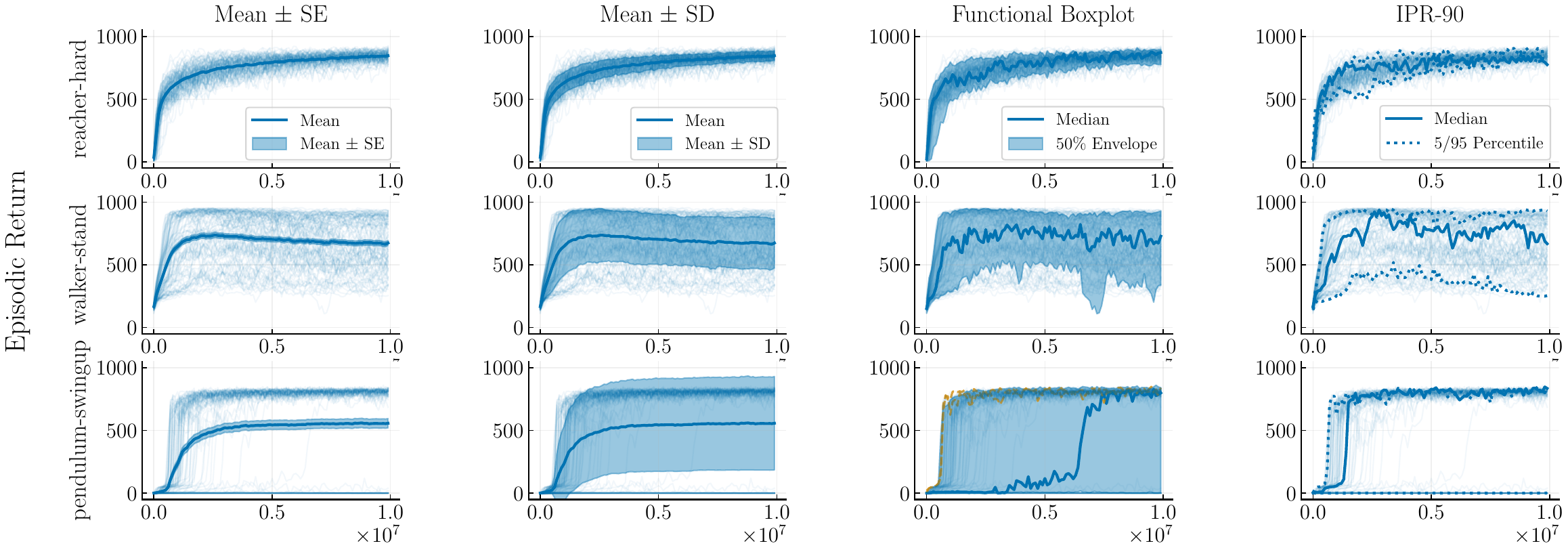}
    \caption[Examples of different visualization methods applied to the default PPO's learning curves for the \texttt{reacher-hard}, \texttt{walker-stand}, and \texttt{pendulum-swingup} tasks.]
    {
        Examples of different visualization methods applied to the PPO's learning curves for the \texttt{reacher-hard}, \texttt{walker-stand}, and \texttt{pendulum-swingup} tasks.
        Each column from left corresponds with the mean \(\pm\) standard error, mean \(\pm\) standard deviation, functional boxplot, and RPH, respectively.
        All learning curves are preprocessed with binning (details in \Cref{sec:appendix:binning}).
        Unlike the other three, RPH provides a simple run-wise perspective, which aligns with the notion of performance variation.
    }
    \label{fig:visualization_example}
\end{figure*}

A popular way to visualize learning curves is to plot the sample mean aggregated across runs at each timestep, along with the associated standard deviation/error as a shaded region.
Despite its popularity, this time-wise format of mean \(\pm\) standard error/deviation potentially leads to inaccurate conclusions, due to the risk of underreporting the variability by standard error/deviation (\Cref{sec:limit_of_std}) and the misrepresentation of actual learning curves.
A prominent case where two of these issues arises is presented in the bottom plots of the first two columns in \Cref{fig:visualization_example}.
Here, the shaded region underestimates the variation, and both the mean curves and the shaded regions represent trends not followed by any individual learning curve, making these plots an inaccurate summary of learning curves as a whole.
Even in a case with a skewed, unimodal performance distribution, the mean \(\pm\) standard deviation band leaves many individual learning curves outside the shaded region, illustrating that point-wise aggregation is not a reliable run-wise visualization of performance variation.
Due to the tendency of underreporting and the risk of misrepresentation of learning curves, the use of mean \(\pm\) standard deviation/error with time-wise aggregation is not suitable for visualizing performance variation.

\begin{figure*}[h!]
    \centering
    \begin{subfigure}[b]{0.49\textwidth}
        \centering
        \includegraphics[width=\textwidth]{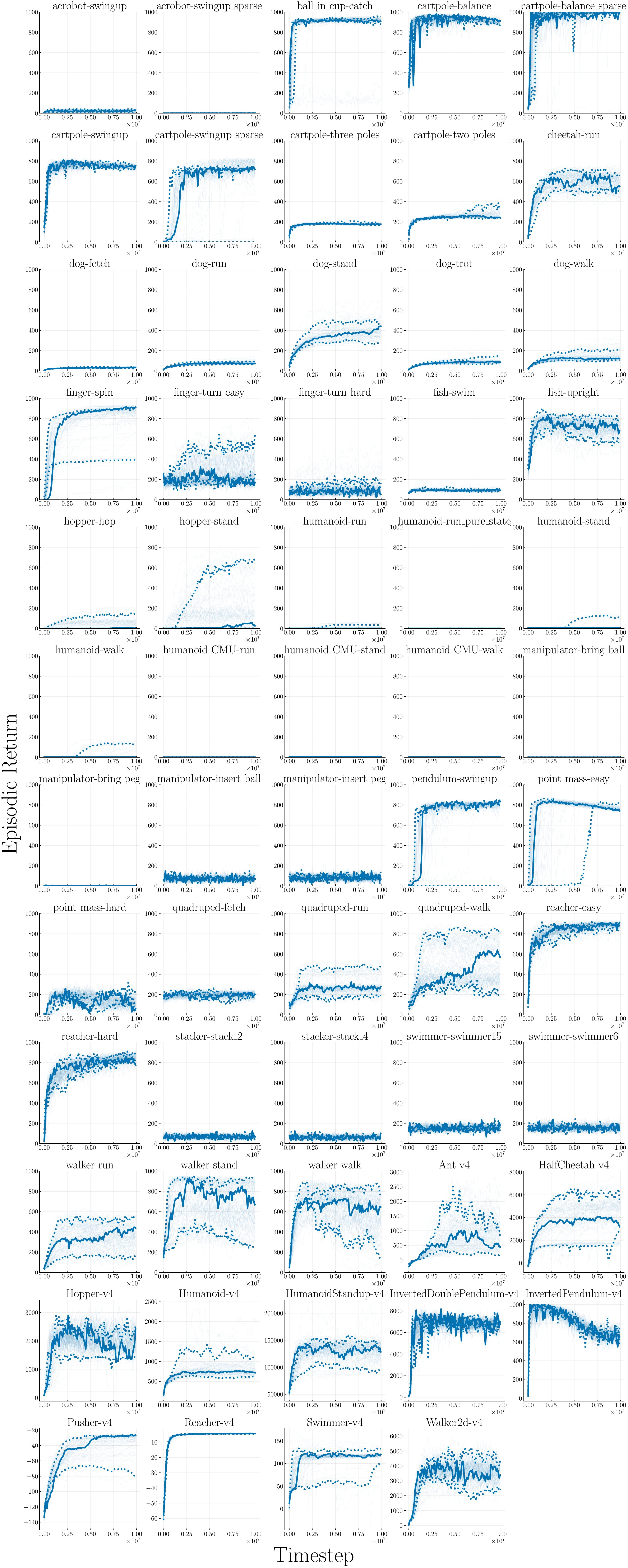}
        \caption{PPO Learning Curves}
        \label{fig:baseline_perf_with_var:ppo}
    \end{subfigure}
    \hfill
    \begin{subfigure}[b]{0.49\textwidth}
        \centering
        \includegraphics[width=\textwidth]{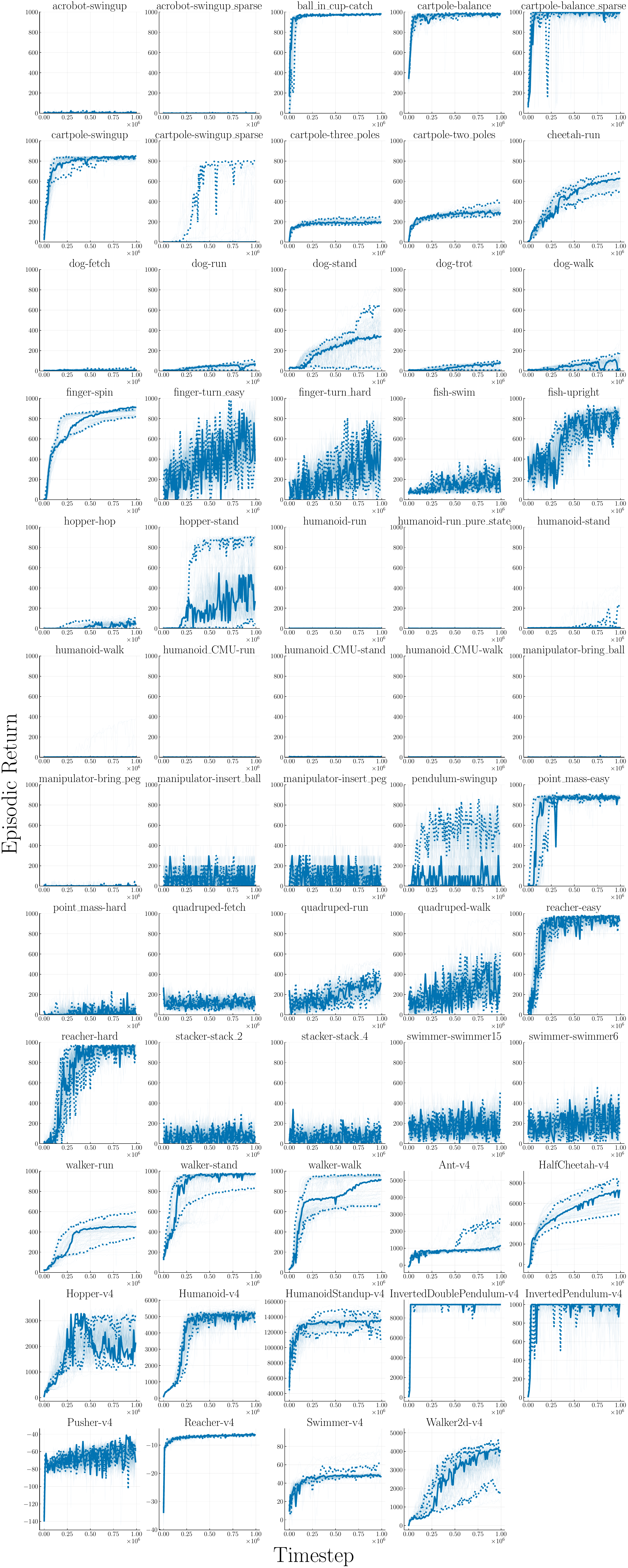}
        \caption{SAC Learning Curves}
        \label{fig:baseline_perf_with_var:sac}
    \end{subfigure}

    \caption[Learning curves with the visualization of variation.]
    {
        Learning curves for PPO and SAC across all \(59\) continuous control tasks, with variation visualized using RPH.
    }
    \label{fig:baseline_perf_with_var}
\end{figure*}

\subsubsection*{Run-wise Percentile Highlighting}
\label{sec:rph:run_wise}

Inheriting the idea of using percentiles from \Cref{sec:measurement}, the most straightforward method is to highlight the \(5\)th, \(50\)th, and \(95\)th percentile learning curves.
We highlight those with high opacity, and colorize the rest with transparency.
The \(5\)th and \(95\)th percentile curves are depicted with a line style different from others to emphasize the variation (we choose a dotted line in this paper).
Because we highlight learning curves at a particular percentile, we refer to this visualization protocol as run-wise percentile highlighting (RPH).

The main competitor to RPH is the functional boxplot (FB), a percentile-based visualization method considered rigorous \citep{sun_2011_functionalBoxplot}.
FB constructs the error band by ranking the closeness of all the curves from the center, which is robust against the distribution of learning curves.
In fact, the third column of \Cref{fig:visualization_example} shows a wide range of area covered by FB, and less misrepresentation compared to the mean \(\pm\) standard error/deviation by highlighting the actual curve that reside in the center of distribution.
Note that the ``center" in FB is not necessarily equivalent to the median, but rather derived from their ranking of closeness (see \Cref{chap:appendix:func_boxplot} for the technical details).
Although FB seems compelling enough, RPH is advantageous compared to FB for visualization of performance variation.

The first advantage of RPH over FB is its computational efficiency and simplicity.
Even without costly and complex operations in FB, RPH is more representative of the actual learning curve distribution, as shown in the examples in the right-most column of \Cref{fig:visualization_example}.
RPH also retains the shape of individual learning curves, effectively avoiding the shortcomings of time-wise aggregation.
In contrast, FB utilizes time-wise aggregation to compute a band, which includes the risk of misrepresenting the learning curves.
Finally, RPH only requires the definition of a percentile as prior knowledge to interpret the results, while FB requires knowledge of band depth and other technical details described in \Cref{chap:appendix:func_boxplot}.
Overall, RPH aligns with our purpose of capturing performance variation while achieving high interpretability without advanced statistical knowledge.
All curves from the PPO and SAC runs, annotated by RPH, are shown in \Cref{fig:baseline_perf_with_var} for reference.

\section{Case Studies}
\label{sec:case_studies}

\begin{wrapfigure}{r}{0.27\textwidth}
    \centering
    \includegraphics[width=0.3\textwidth]{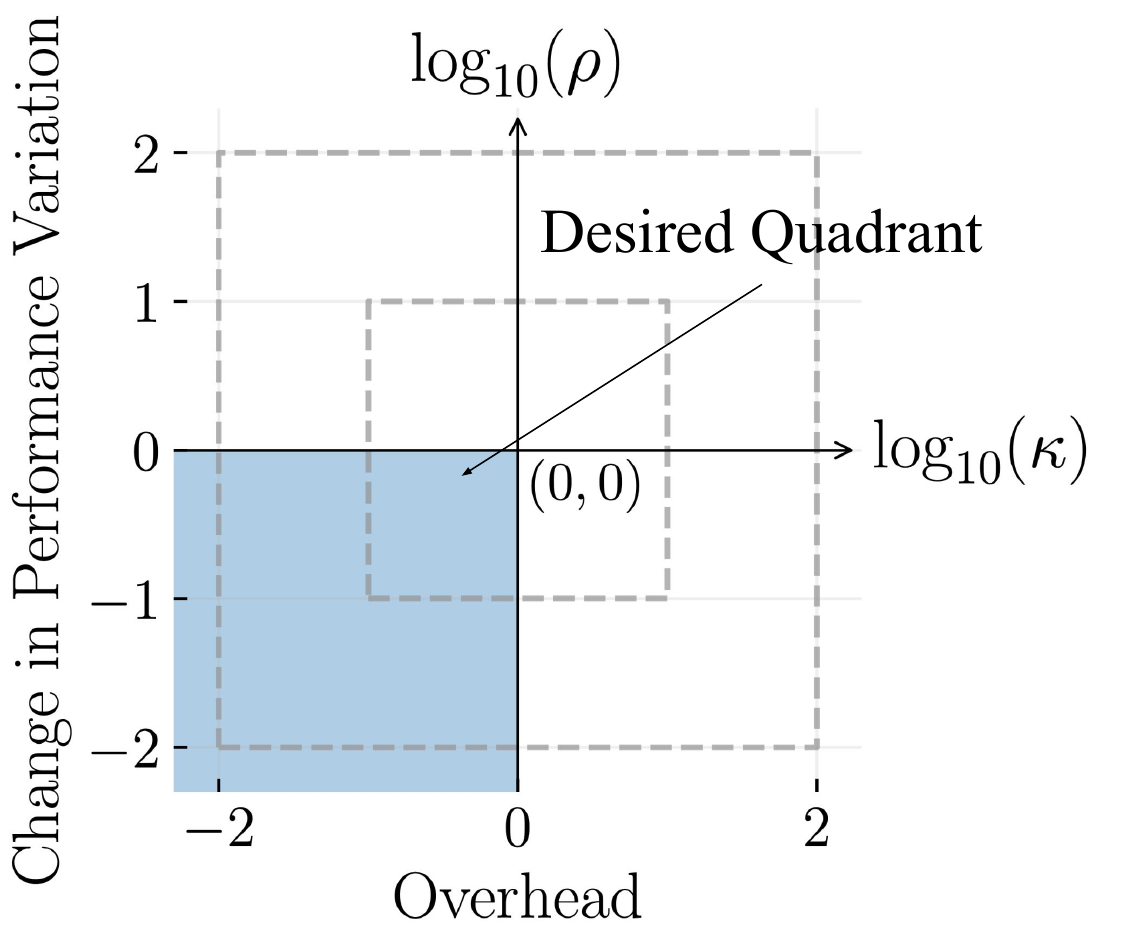}
    \caption[Visualization of the plotting format for variation change and its associated overhead.]
    {
        Plotting format for variation change and overhead.
        The plot takes \(\log_{10}\)-scaled values of overhead (\(\kappa\)) and change in variation (\(\rho\)) on its \(x\) and \(y\) axis, respectively.
        The modification is ideal if most or all points fall within the blue-shaded region, as this indicates reduced performance variation and higher overall performance.
    }
    \label{fig:effectiveness_plot}
    \vspace{-10pt}
\end{wrapfigure}
So far, we have shown that both the min–max IPR-\(90\) and RPH are relatively suitable tools for analyzing performance variation.
The natural next question is their effectiveness in practice.
To address this, we present three case studies.
First, we examine whether LayerNorm and penultimate layer normalization (PNorm) reduce a performance variation in PPO and SAC \citep{ba_2016_layerNorm, bjorck_2022_high_variance}.
Second, we compare the performance distributions of popular continuous control algorithms, including PPO, SAC, TD-MPC, and TD-MPC2.
Lastly, we repeat a similar comparative analysis for discrete control algorithms, namely DQN and Rainbow.

\subsubsection*{Case Study 1: Effect of Normalization on Performance Variation}
\label{sec:case_studies:relative}

As a first case, we examine the change in performance variation after applying LayerNorm and PNorm to the default PPO/SAC settings described in \Cref{sec:exp_settings}.
In this experiment, LayerNorm is applied to pre-activations of each layer in actor and critic networks.
PNorm applies \(L^2\) normalization to penultimate layer outputs of both networks.
Otherwise, the configurations remain the same.
We apply both methods individually and jointly.
Variants with either modification are referred to as LayerNorm PPO/SAC and PNorm PPO/SAC, and the variant with both modifications as Normalized PPO/SAC.

\begin{figure*}[t!]
    \centering
    \begin{subfigure}[b]{0.33\textwidth}
        \centering
        \includegraphics[width=\textwidth]{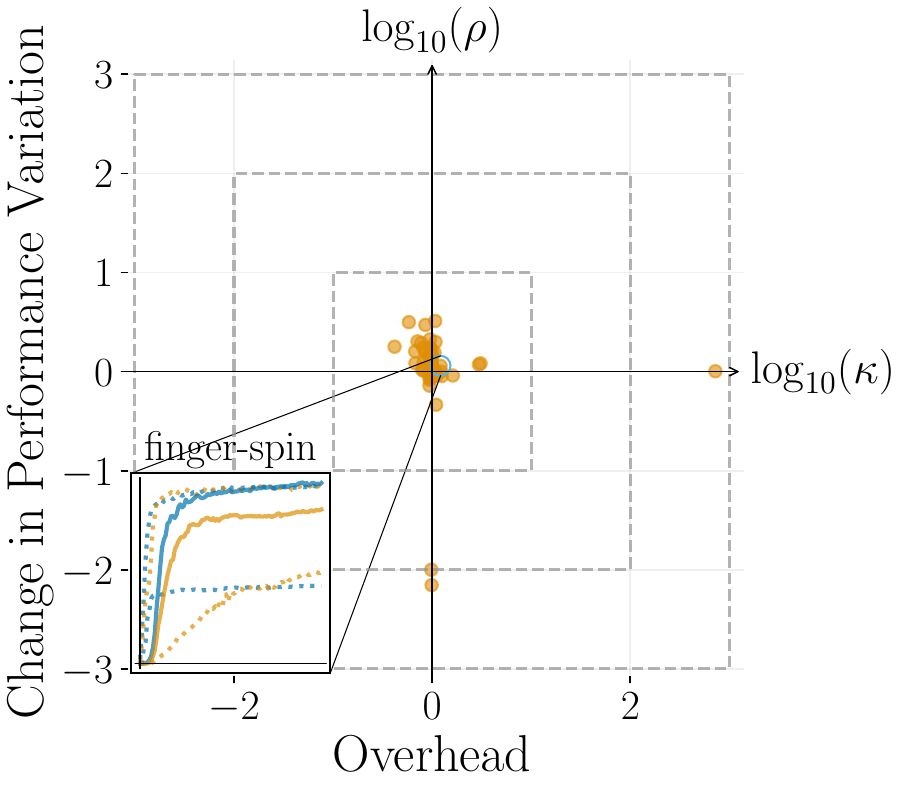}
        \caption{\footnotesize PPO vs. LayerNorm PPO}
        \label{fig:perf_comp:ppo_layernorm}
    \end{subfigure}
    \hfill
    \begin{subfigure}[b]{0.32\textwidth}
        \centering
        \includegraphics[width=\textwidth]{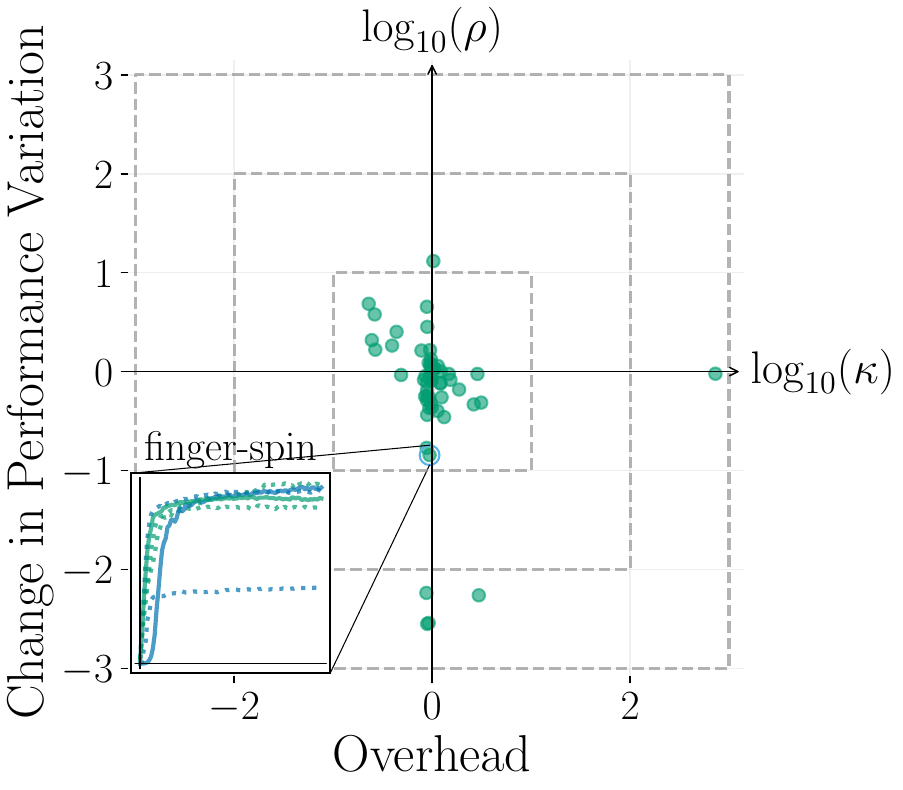}
        \caption{\footnotesize PPO vs. PNorm PPO}
        \label{fig:perf_comp:ppo_pnorm}
    \end{subfigure}
    \hfill
    \begin{subfigure}[b]{0.32\textwidth}
        \centering
        \includegraphics[width=\textwidth]{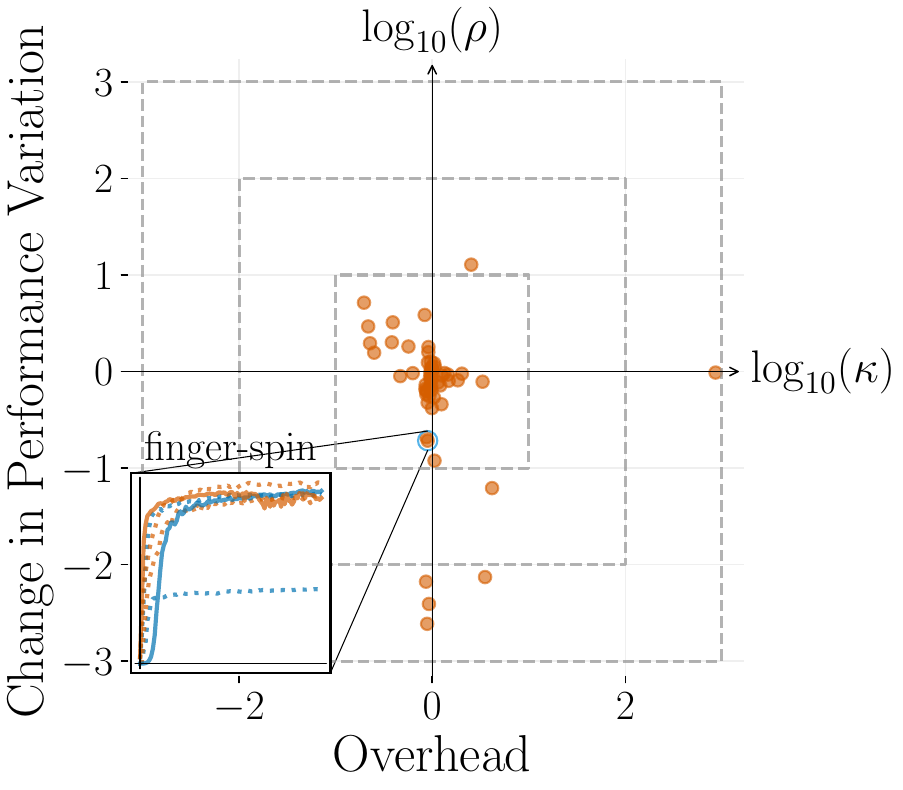}
        \caption{\footnotesize PPO vs. Normalized PPO}
        \label{fig:perf_comp:ppo_normalized}
    \end{subfigure}\\
    \begin{subfigure}[b]{0.32\textwidth}
        \centering
        \includegraphics[width=\textwidth]{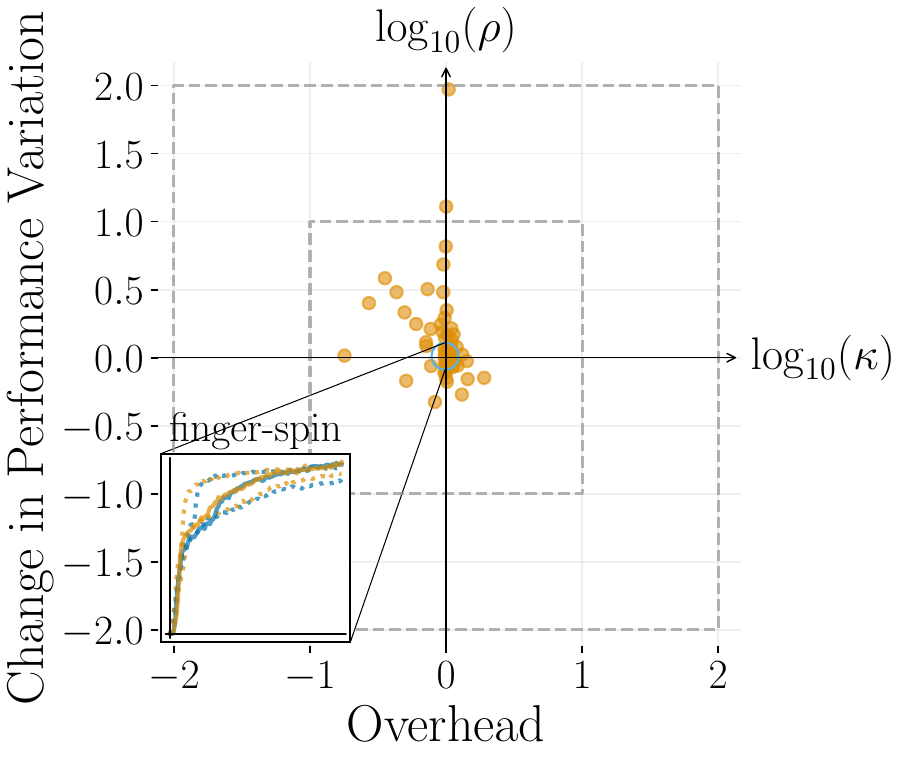}
        \caption{\footnotesize SAC vs. LayerNorm SAC}
        \label{fig:perf_comp:sac_layernorm}
    \end{subfigure}
    \hfill
    \begin{subfigure}[b]{0.32\textwidth}
        \centering
        \includegraphics[width=\textwidth]{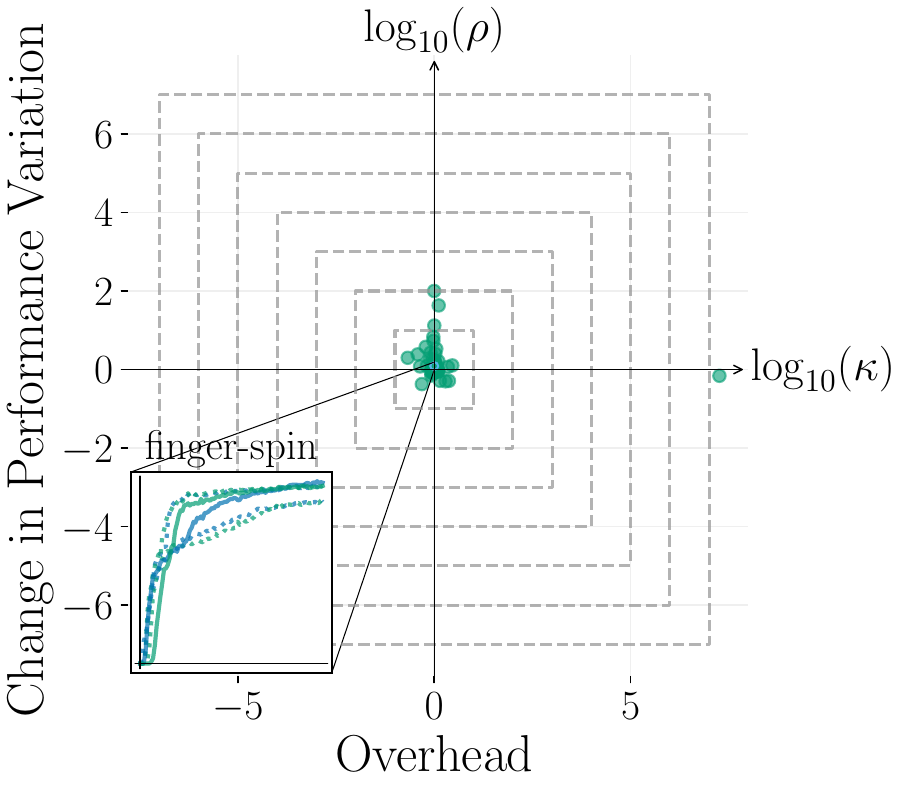}
        \caption{\footnotesize SAC vs. PNorm SAC}
        \label{fig:perf_comp:sac_pnorm}
    \end{subfigure}
    \hfill
    \begin{subfigure}[b]{0.32\textwidth}
        \centering
        \includegraphics[width=\textwidth]{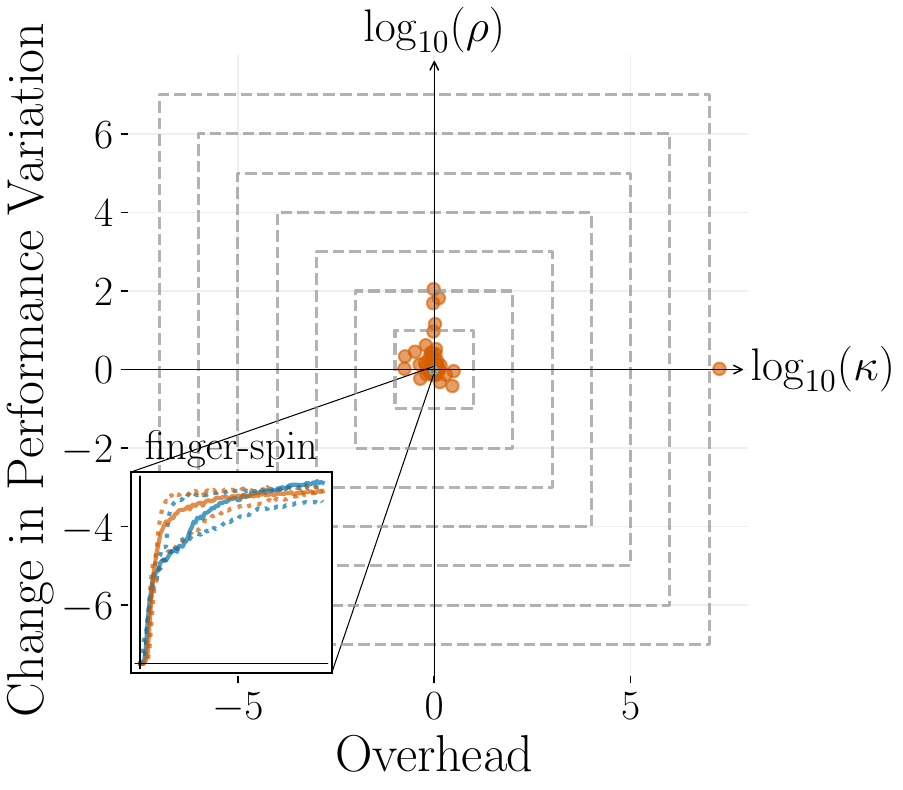}
        \caption{\footnotesize SAC vs. Normalized SAC}
        \label{fig:perf_comp:sac_normalized}
    \end{subfigure}
    \caption
    {
        Change of performance variation and median after applying modifications to PPO and SAC.
        Zoomed-in plots compare RPH learning curves for default and modified algorithms on \texttt{finger-spin}.
        The positive effect of normalization techniques, especially their combination, is evident for PPO.
        In contrast, modifications to SAC generally leave performance variation unchanged or increase it.
    }
    \label{fig:perf_comp}
\end{figure*}

For each task, we report the ratio of performance variation between the variants of algorithms as a change in performance variation and denote with \(\rho\).
Mathematically, \(\rho\) is defined as
\begin{align}
    \rho(\mathcal{U}_{\textnormal{baseline}}, \mathcal{U}_{\textnormal{modified}}) := \frac{\textnormal{Min-max IPR}(\mathcal{U}_{\textnormal{modified}}, 90)}{\textnormal{Min-max IPR}(\mathcal{U}_{\textnormal{baseline}}, 90) + \epsilon},
    \label{eq:rho}
\end{align}
\noindent
where \(\mathcal{U}_{\textnormal{baseline}}\) is a set of performance for running algorithm with default configuration, \(\mathcal{U}_{\textnormal{modified}}\) is a set of performance for running the same algorithm with either of normalization techniques, and \(\epsilon > 0\) is a small constant to avoid zero division.
While \Cref{eq:rho} captures the change in performance variation, it cannot identify the cause of the change.
For instance, \Cref{eq:rho} cannot identify whether improvement is sourced by collapse of all learning curves to the minimum episodic returns or improvement of lower performing learning curves.
To disambiguate the cause of change in performance variation, we also obtain the overhead, the ratio between the median performances.
The overhead, denoted by \(\kappa\), is formulated as
\begin{align}
    \kappa(\mathcal{\widehat{U}}_{\textnormal{baseline}}, \mathcal{\widehat{U}}_{\textnormal{modified}}) := \frac{\mathcal{\widehat{U}}_{\textnormal{baseline}}^{(50)}}{\mathcal{\widehat{U}}_{\textnormal{modified}}^{(50)} + \epsilon}.
    \label{eq:kappa}
\end{align}
\noindent
Note that we use \(\mathcal{\widehat{U}}_{\textnormal{baseline}}\) and \(\mathcal{\widehat{U}}_{\textnormal{modified}}\), which are \(\mathcal{U}_{\textnormal{baseline}}\) and \(\mathcal{U}_{\textnormal{modified}}\) shifted by \(-\min(\{\min(\mathcal{U}_{\textnormal{baseline}}), \min(\mathcal{U}_{\textnormal{modified}}), 0\})\).
This is done to satisfy a condition of \(\kappa \geq 0\).
For a compact visualization, we report both values of \(\rho\) and \(\kappa\) by plotting them as a scatter plot in \(\log_{10}\) scale, and treat points in the third quadrant as an instance where modifications improve the performance without much overhead (see \Cref{fig:effectiveness_plot}).

\Cref{fig:perf_comp} depicts a comparison plot between the default and normalized variants of PPO and SAC.
Graphically, normalization techniques either reduce performance variation with less overhead or increase the overall performance for PPO.
On the other hand, performance variation generally stays the same or increases after applying normalization techniques to SAC.
These results imply that, at least when one inherits hyperparameter from the default configuration, normalization techniques benefit PPO by a considerable amount, while they do not for SAC.
Notice that these results are easily captured by application of min-max IPR-\(90\).

Improvements in performance variation is also presented by overlaying the RPH curves.
Each subfigure in \Cref{fig:perf_comp} includes a zoomed-in view of the RPH-visualized learning curves for \texttt{finger-spin}.
Blue curves represent the learning curves from default runs, and alternative colors represent those from modified runs.
We readily observe how PNorm, especially when combined with LayerNorm, narrows the gap between PPO learning curves.
Also, the gap between learning curves remains unchanged in the case of SAC.
The learning curve comparisons for all algorithm-task pairs are shown in \Cref{sec:appendix:use_case_comp}, and bar plots for performance variation and median performance are in \Cref{sec:appendix:comp_bar_plots}.

\subsubsection*{Case Study 2: Cross-algorithm Comparative Study}
\label{sec:case_studies:cross_algo_comp}

Another use case of our proposed methods is on a comparative study of different algorithms.
Here, we consider two deep RL algorithms, TD-MPC and TD-MPC2, in addition to PPO and SAC.
TD-MPC is a major model-based deep RL algorithm for continuous control, which combines a short-term reward estimate from a latent dynamics model with a long-term return estimate from a value function.
TD-MPC2 is an upgraded version of TD-MPC that offers greater scalability and improved robustness across tasks.
In this paper, we adopt the original implementation and run \(100\) independent runs with hyperparameters given by \citet{hansen_2022_TDMPC} and \citet{hansen_2024_TDMPC2} (also listed in \Cref{table:tdmpc_configs,table:tdmpc2_configs}).
Performance variation and median performance of TD-MPC/TD-MPC2 are given in \Cref{fig:tdmpc_var_and_med}.
Now, we consider the following question: among PPO, SAC, TD-MPC, and TD-MPC2, which algorithm is superior in average sample mean, sample standard deviation, IPR-\(90\), \(5\)th percentile, median, and \(95\)th percentile performance?
We answer this question by comparing the histogram of performance statistics in the \(48\) DMC tasks.

\begin{figure*}[htb!]
    \centering
    \begin{subfigure}[b]{0.48\textwidth}
        \centering
        \includegraphics[width=\textwidth]{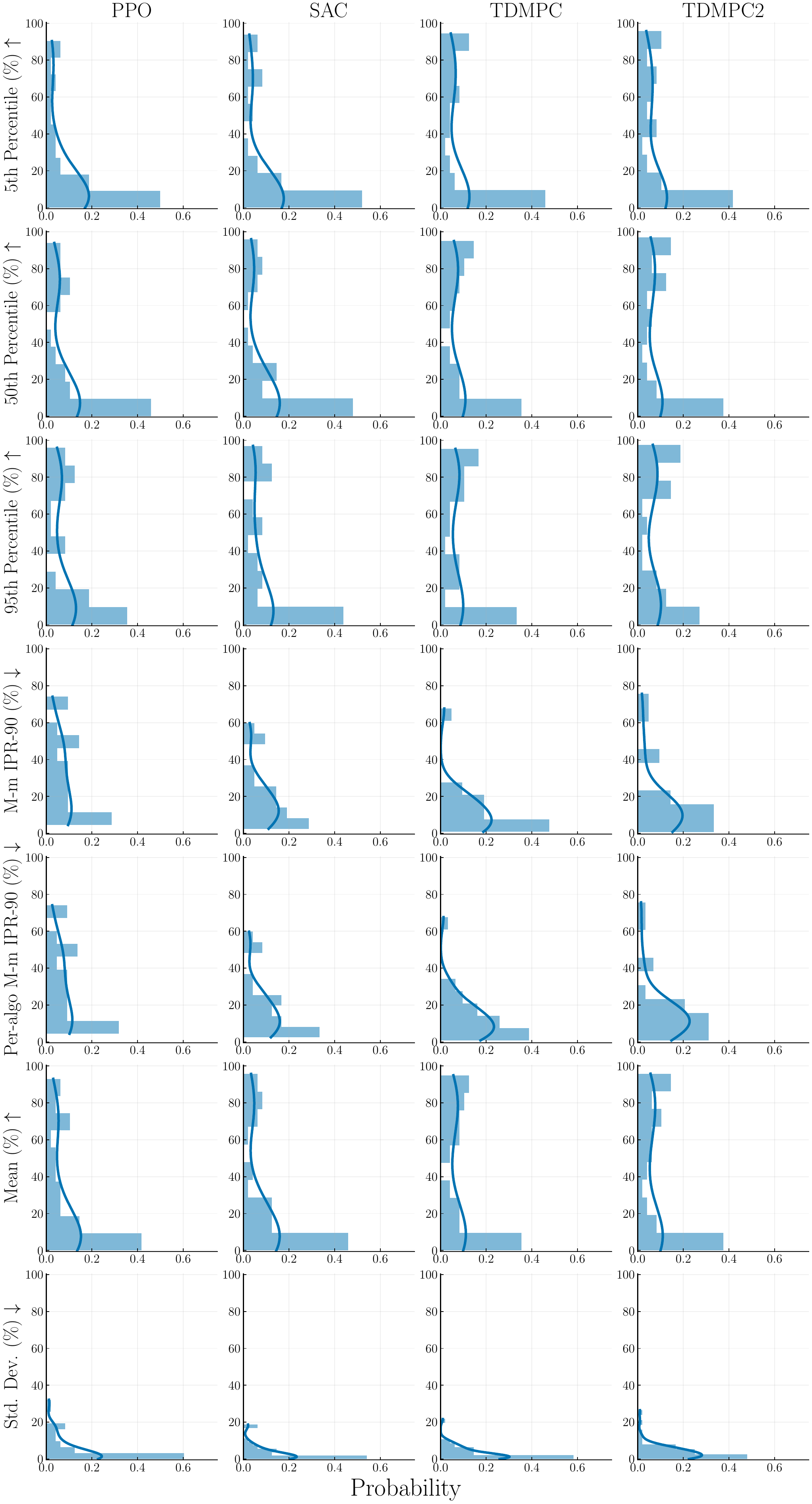}
        \caption{Full Timestep}
        \label{fig:stat_summary:full}
    \end{subfigure}
    \hfill
    \begin{subfigure}[b]{0.48\textwidth}
        \centering
        \includegraphics[width=\textwidth]{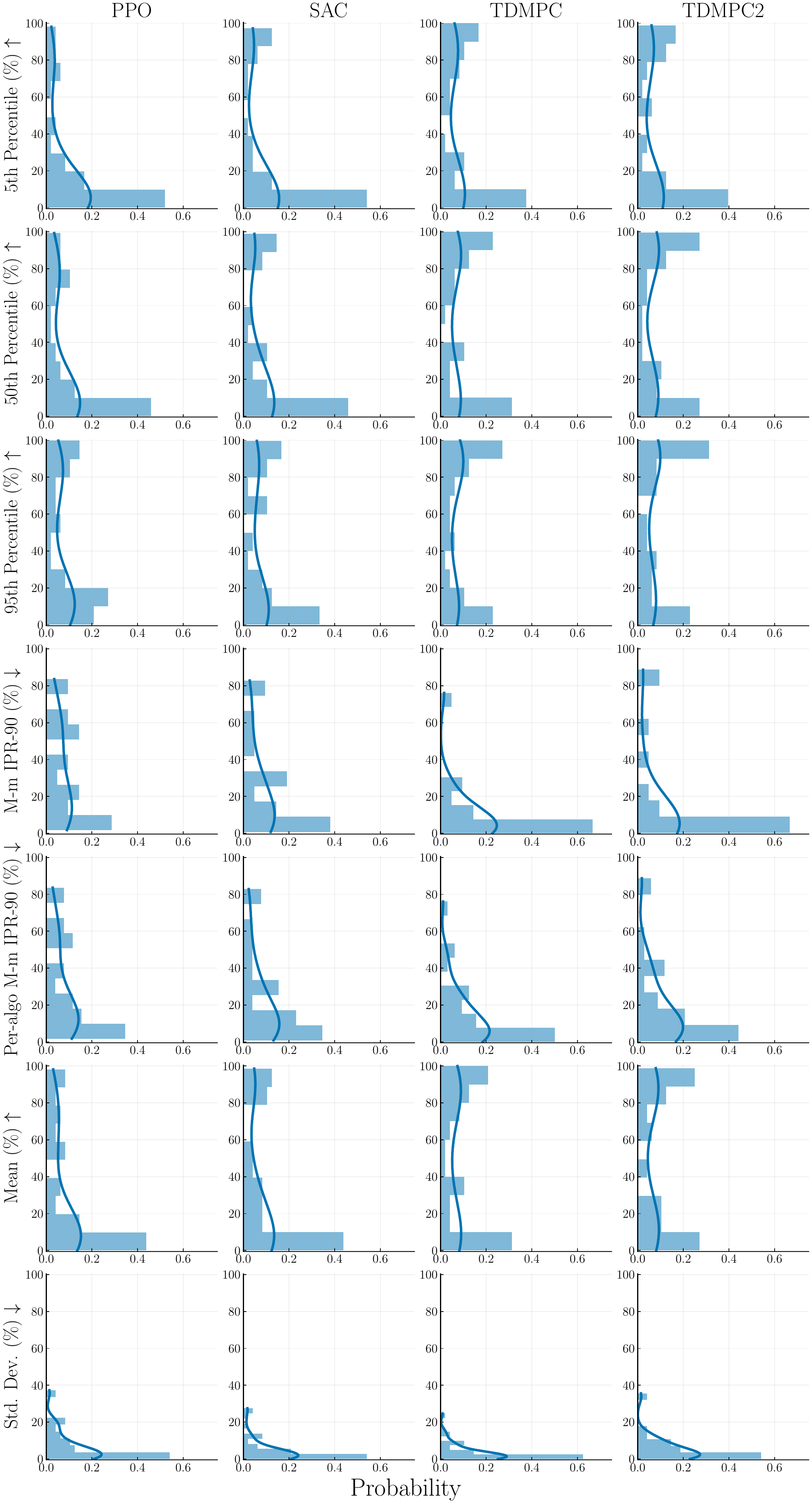}
        \caption{Last \(100\)k Steps}
        \label{fig:stat_summary:last}
    \end{subfigure}

    \caption
    {
        Distribution of task-level performance statistics computed from the corresponding algorithm's performance across multiple DMC tasks.
        Performance in \textbf{(a)} considers episodic returns over entire training steps, whereas \textbf{(b)} only considers episodic returns over the last \(100\)k training steps.
        Row and column in each subfigure correspond to the statistic and the algorithm, respectively.
        Each subplot shows a histogram with KDE, where the x-axis represents probability and the y-axis shows min-max normalized statistics.
        Histograms for Min-max IPR-\(90\) (labeled as M-m IPR-\(90\)) uses performances from \(21\) DMC tasks where none of algorithms' \(95\)th percentile performance are under \(200\).
        As a variant, histograms labeled as Per-algo M-m IPR-\(90\) uses performances from tasks where the corresponding algorithm's \(95\)th percentile performance is above \(200\).
        All other histograms are produced from the performances on \(48\) tasks.
        TD-MPC tends to exhibit less performance variation while marking relatively higher performance.
    }
    \label{fig:stat_summary}
\end{figure*}

\Cref{fig:stat_summary} shows the histogram of the performance statistics for each task.
Each row and column correspond to a statistic and an algorithm, respectively, and all quantities are min-max normalized.
Note that in computing the average min–max IPR-\(90\) (referred to as M-m IPR-\(90\) in \Cref{fig:stat_summary}), we exclude \(27\) tasks where at least one algorithm marks the performance below \(200\) in its \(95\)th percentile run.
This is because failed runs often have low performance spread, which can obscure comparisons of performance variation among algorithms.
Of the \(27\) excluded tasks, the \(95\)th percentile performance of PPO, SAC, TD-MPC, and TD-MPC2 is under \(200\) in \(26\), \(24\), \(17\), and \(19\) tasks, respectively.
For completeness, we also report the distribution of min-max IPR-\(90\) for each algorithm without the tasks they failed to surpass the performance of \(200\) (referred to as Per-algo M-m IPR-\(90\) in \Cref{fig:stat_summary}).
Additionally, \Cref{table:avg_stats} summarize the sample mean of histograms in \Cref{fig:stat_summary}.
Both \Cref{fig:stat_summary} and \Cref{table:avg_stats} indicates the dominance of TD-MPC and TD-MPC2 against the other two in all aspects.
The average median performance of the TD-MPC algorithms is almost \(1.5\) times that of the PPO/SAC.
Similar patterns also hold for mean, \(5\)th percentile, and \(95\)th percentile performance.
Furthermore, min-max IPR-\(90\) histogram for TD-MPC concentrates more around the lower values.
In fact, the average min-max IPR-\(90\) for TD-MPC is around \(10\%\), which is surprisingly low.
Thus, when all four algorithms are trained on \(48\) DMC tasks, TD-MPC outperforms the others in a data efficient manner, while exhibiting a lower or at least equivalent variation.

\begin{table}[t]
\caption{
    Average statistics of performance of continuous control algorithms over multiple DMC tasks.
    All the statistics are min-max normalized, and hence the units of value are percentages.
    Mean, standard deviation, median, and \(5\)/\(95\)th percentiles are calculated with all \(48\) tasks.
    Min-max IPR-\(90\) (labeled as M-m IPR-\(90\)) is calculated with \(21\) tasks where none of algorithms' \(95\)th percentile performance are under \(200\).
    As a variant of M-m IPR-\(90\), we also report Min-max IPR-\(90\) with tasks where each algorithm's \(95\)th percentile performance is above \(200\), which is labeled as Per-algo M-m IPR-\(90\). 
    Parentheses in each column name indicate the range of final timesteps used to compute each statistic, separated by slash.
    Each entry in table is separated by slash, where each value indicates statistic computed with corresponding number of final timesteps.
    TD-MPC achieves the lowest average performance variation while being more data efficient than PPO/SAC.
}
\label{table:avg_stats}
\tiny
\centering
\begin{tabular}{l c c c c c}
\toprule
 Aggregate over \(48\)/\(21^*\)/\\varying \#\(^\dagger\) of tasks
 & PPO (\(10\)M/\(100\)k)
 & SAC (\(1\)M/\(100\)k)
 & TD-MPC (\(500\)k/\(100\)k)
 & TD-MPC2 (\(500\)k/\(100\)k)\\
\midrule
{\(95\)th Percentile (\%) \(\uparrow\)} & \(35.05 / 39.23\) & \(32.58 / 40.68\) & \(43.16 / 52.98\) & \(\mathbf{44.82} / \mathbf{54.0}\) \\
{Median (\%) \(\uparrow\)} & \(28.93 / 30.45\) & \(26.42 / 32.07\) & \(39.82 / 48.60\) & \(\mathbf{40.72} / \mathbf{50.23}\) \\
{\(5\)th Percentile (\%) \(\uparrow\)} & \(20.62 / 21.06\) & \(22.2 / 25.93\) & \(\mathbf{33.68} / \mathbf{41.59}\) & \(32.61 / 38.69\) \\
{M-m IPR-\(90^*\) (\%) \(\downarrow\)}  & \(29.28 / 33.37\) & \(19.94 / 25.30\) & \(\mathbf{11.98} / \mathbf{10.54}\) & \(18.17 / 17.45\) \\
{Per-algo M-m IPR-\(90^\dagger\) (\%) \(\downarrow\)}  & \(28.15 / 29.20\) & \(18.78 / 23.74\) & \(\mathbf{13.40} / \mathbf{15.04}\) & \(17.67 / 19.89\) \\
\midrule
{Mean (\%) \(\uparrow\)} & \(28.09 / 29.9\) & \(26.77 / 32.48\) & \(39.23 / 48.13\) & \(\mathbf{39.99} / \mathbf{48.69}\) \\
{Std. Dev.\(^*\) (\%) \(\downarrow\)} & \(11.16 / 13.29\) & \(6.55 / 8.49\) & \(\mathbf{4.03} / \mathbf{3.40}\) & \(6.31 / 7.15\) \\
\bottomrule
\end{tabular}
\end{table}

\begin{figure*}[htb]
    \centering
    \includegraphics[width=0.95\textwidth]{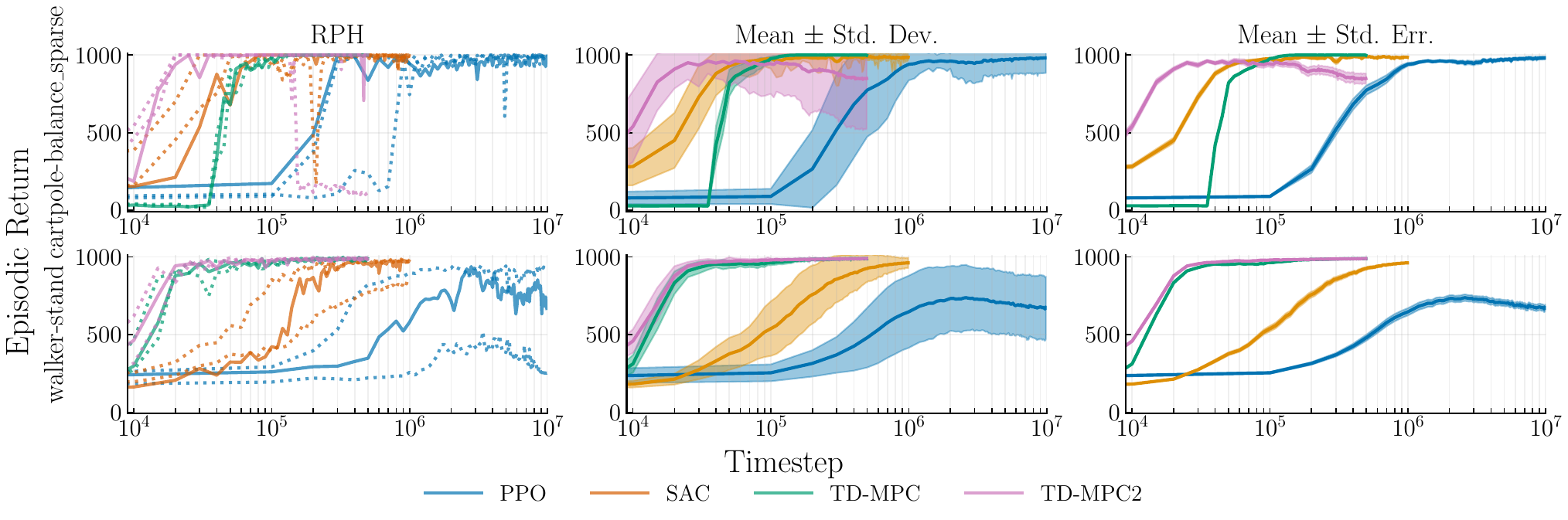}
    \caption{
        Comparison of learning curves of four continuous control algorithms on two selected DMC tasks in different plotting styles.
        For visual clarity, RPH curves are plotted without non-highlighted curves.
        The x-axis of each subfigure is log-scaled.
        RPH most accurately represents the fact that TD-MPC/TD-MPC2 learns more rapidly than PPO/SAC, while exhibiting less performance variation.
    }
    \label{fig:learning_curve_comparison_rph_compact}
\end{figure*}

Although TD-MPC achieves a relatively low min-max IPR-\(90\) of around \(10\%\), it is still higher than the reasonable value of \(5\%\) (see \Cref{sec:measurement}).
This implies that, even with modern advanced methods, deep RL algorithms have yet to achieve sufficiently low performance variation for practical use. 
Also, recall that TD-MPC does not achieve the performance above \(200\) in \(17\) out of \(48\) tasks with its \(95\)th percentile performance.
This is around \(35\%\) of the failure rate.
In contrast, none of the performance of our SL experiments collapses, while achieving low variation (see \Cref{sec:appendix:sl_learning_curves}).
These insights pose another important challenge for modern advanced deep RL methods: consistently learning successfully with low performance variation.

Note that similar insights gained from \Cref{fig:stat_summary} and \Cref{table:avg_stats} can also be drawn by observing \Cref{fig:learning_curve_comparison_rph_compact}, thanks to RPH.
For instance, on the \texttt{walker-stand} task, \Cref{fig:learning_curve_comparison_rph_compact} clearly shows a wide range of performance variation in the PPO algorithm, from an episodic return of $200$ to $900$.
It also captures the remarkably small performance variation of the TD-MPC and TD-MPC2 algorithms on the same task.
Furthermore, the information that is not fully described in \Cref{fig:stat_summary} and \Cref{table:avg_stats} is perceivable from the RPH curves in \Cref{fig:learning_curve_comparison_rph_compact}.
For example, some failure modes of TD-MPC2 can be observed from the \(5\)th percentile curve of \texttt{cartpole-balance\_sparse} task.
These insights cannot even be drawn from the standard error/deviation plots in \Cref{fig:learning_curve_comparison_rph_compact}, where the shaded regions representing the standard error are narrow or barely visible.
For a full comparison between different plotting styles, see \Cref{fig:learning_curve_comparison_rph,fig:learning_curve_comparison_std,fig:learning_curve_comparison_stderr}.
Nevertheless, RPH is advantageous and practical for visualizing a performance variation.

\subsubsection*{Case Study 3: Cross-algorithm Comparative Study with Discrete Control Algorithms}
\label{sec:case_studies:atari}

The empirical results in previous sections focus solely on continuous control settings.
For completeness, we repeat the cross-algorithm comparative study from the previous section for discrete control deep RL algorithms, particularly for Deep Q-Network (DQN) and Rainbow \citep{mnih_2015_HumanlevelControlDeep,hessel_2018_RainbowCombiningImprovements}.
We compare the performance statistics of these two algorithms by running them for \(50\) million steps on five Arcade Learning Environments (ALE) \citep{bellemare_2013_ALE}.
We adopted the CleanRL implementation, and ran \(100\) independent runs per environment \citep{huang_2022_cleanrl}.
The five environments have been claimed to be essential among all ALEs by \citet{aitchison_2023_Atari5DistillingArcade}.
To ensure stochasticity in environments, we reset all environments with a random no-op period and enable sticky actions \citep{machado_2018_RevisitingArcadeLearning}.  
Hyperparameter details are provided in \Cref{table:dqn_configurations,table:rainbow_configurations}. 

\begin{figure*}[t!]
    \centering
    \includegraphics[width=0.9\textwidth]{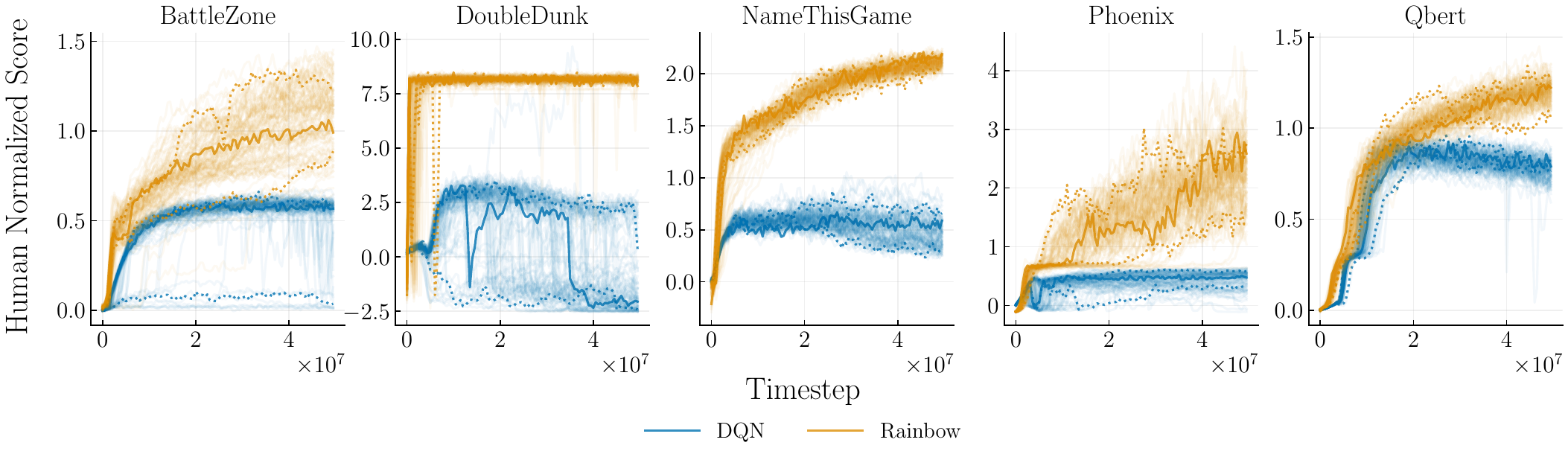}
    \caption{
        Comparison of learning curves of discrete control algorithms on five Atari tasks.
        Each subfigure shows the RPH Human Normalized Score curves for DQN and Rainbow on corresponding task.
        Rainbow generally outperforms DQN, and both algorithms exhibit significant performance variation in some tasks.
    }
    \label{fig:learning_curve_comparison_dqn_rainbow}
\end{figure*}

\begin{table}[t!]
\caption{
    Average statistics of performance of discrete control algorithms over \(5\) Atari tasks.
    All the statistics are min-max normalized, and hence the units of value are percentages.
    Parentheses in each column name indicate the range of final timesteps used to compute each value, separated by slash.
    Each entry in table is separated by slash, where each value indicates statistic computed with corresponding number of final timesteps.
    In all aspects, Rainbow outperforms DQN.
    Also, both algorithms exhibit similar levels of performance variation.
}
\label{table:avg_stats_discrete}
\scriptsize
\centering
\begin{tabular}{l c c }
\toprule
 Aggregate over \(5\) tasks
 & DQN {\scriptsize (\(50\)M/\(500\)k)}
 & Rainbow {\scriptsize (\(50\)M/\(500\)k)} \\
\midrule
{\(95\)th Percentile (\%) \(\uparrow\)} & \(28.44 / 31.78\) & \(\mathbf{63.15} / \mathbf{81.52}\) \\
{Median (\%) \(\uparrow\)} & \(23.24 / 20.94\) & \(\mathbf{56.71} / \mathbf{71.78}\) \\
{\(5\)th Percentile (\%) \(\uparrow\)} & \(11.05 / 8.15\) & \(\mathbf{48.84} / \mathbf{60.70}\) \\
{Min-max IPR-\(90\) (\%) \(\downarrow\)}  & \(17.39 / 23.63\) & \(\mathbf{14.31} / \mathbf{20.81}\) \\
\midrule
{Mean (\%) \(\uparrow\)} & \(22.46 / 20.96\) & \(\mathbf{56.47} / \mathbf{71.24}\) \\
{Std. Dev. (\%) \(\downarrow\)} & \(5.33 / 7.82\) & \(\mathbf{4.32} / \mathbf{6.56}\) \\
\bottomrule
\end{tabular}
\end{table}

Statistics in \Cref{table:avg_stats_discrete} indicate the dominance of Rainbow over DQN.
Thanks to the learning curves with RPH in \Cref{fig:learning_curve_comparison_dqn_rainbow}, we readily observe that this dominance arises because the near-worst runs of Rainbow outperform the near-best runs of DQN.
Additionally, min-max IPR-\(90\) in \Cref{table:avg_stats_discrete} shows that both algorithms experience a similar level of high performance variation.
Together, Rainbow indeed improves overall performance in comparison to DQN, but does not solve the problem of high performance variation.

\section{Limitations}
\label{sec:limitations}

One of the prominent limitations of IPR-\(90\) is the choice of \(90\%\) as the range of variation.
Although we discuss this number strikes a reasonable tradeoff between the range it covers and the sensitivity against extrema, it still lacks theoretical motivation.
A potentially superior approach is to report barplot for various IPR ranges, but this can decrease interpretability by increasing the amount of information (but see  \Cref{sec:appendix:multi_level_barplots}).
The pursuit of a concise multi-level IPR is certainly an exciting direction.

Another limitation is the number of independent runs for which our proposed method functions well.
Since both min-max IPR-\(90\) and RPH naively use the \(5\)-th and \(95\)-th percentile data, the number of independent runs is a deciding factor of accuracy.
This property restricts the range of applicable settings.
Especially, it is less applicable when a single run leverages substantial computational resources (e.g., large language models).
Although, even with fewer runs, our methods are beneficial compared to conventional methods.
Regardless of the smaller number of samples (e.g., \(5\) runs), IPR is more faithful to the observed spread compared to parametric statistics, such as standard deviation.
RPH provides a clear and accurate depiction of all the learning curves, which offers richer information of how the learning curves behave than in plots with shaded region.
Although plotting all the learning curves may raise concerns about a drop in visual clarity, the visualizations remain sufficiently clear.
In fact, while \Cref{fig:learning_curve_comparison_dqn_rainbow} depicts all \(100\) independent runs in RPH format, the visual clarity is retained.

\section{Conclusion}
\label{sec:conclusion}

In this paper, we proposed min-max IPR-\(90\) and RPH as intuitive, yet robust methods to capture performance variation in a single task, compared to previous approaches such as standard error/deviation.
These characteristics enable the broad applicability of our proposed methods to deep RL research.
In fact, using our proposed methods, we have shown that normalizations reduce performance variation in PPO, TD-MPC exhibits surprisingly low performance variation, and DQN and Rainbow experience a similar level of performance variation.
Although effective, because both proposed methods are based on specific statistics, relying on them alone can mislead insights.
To avoid such a situation and provide complementary insights, our methods should be used alongside other statistical tools.
We hope that our methods will aid future deep RL research by providing additional insights into empirical results.


\subsubsection*{Acknowledgments}
\label{sec:ack}
The authors sincerely thank John D. Martin and Levi H. S. Lelis for their thorough feedback and comments on this work.
The authors also thank the University of Alberta, Amii, the Natural Sciences and Engineering Research Council of Canada (NSERC), the Canada CIFAR AI Chairs program, and the Digital Research Alliance of Canada for the funding and computational resources.


\bibliography{main}

\appendix
\crefalias{section}{appendix}

\section{Binning Operation}
\label{sec:appendix:binning}

\begin{figure*}[h]
    \centering
    \begin{subfigure}[t]{0.35\textwidth}
        \centering
        \includegraphics[width=\textwidth]{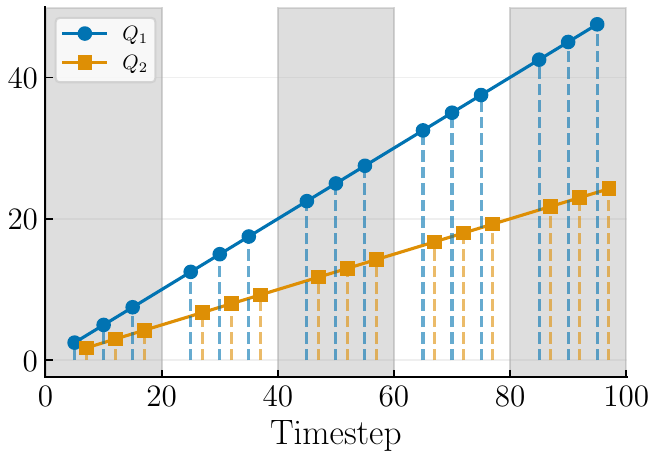}
        \caption{Pre-binned}
        \label{fig:synth_bin:pre_bin}
    \end{subfigure}
    \hspace{0.05\textwidth}
    \begin{subfigure}[t]{0.35\textwidth}
        \centering
        \includegraphics[width=\textwidth]{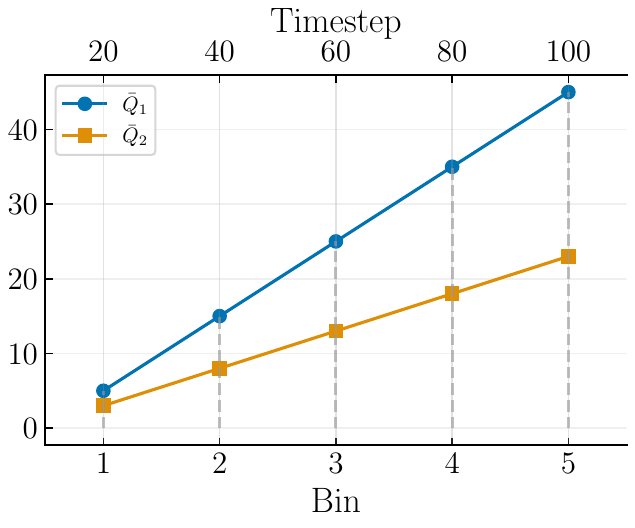}
        \caption{Binned}
        \label{fig:synth_bin:binned}
    \end{subfigure}
    \caption[Exemplar binning on synthetic data.]
    {
        Exemplar binning process for synthetic data.
        Datapoint in different time-series data can be collected at different timesteps, as shown in \textbf{(a)}.
        By aggregating the data on a bin basis (where each bin is represented as a shaded/unshaded region in \textbf{(a)}), one can compare two time-series data more easily as shown in \textbf{(b)}.
    }
    \label{fig:synth_bin}
\end{figure*}

By the nature of learning through interaction, a lot of data/statistics collected from RL experiments have a chronological structure.
Because such time-series data are challenging to handle, we often employ multiple preprocessing procedures to simplify them.
Here, we cover one instance of such methods that are employed throughout this paper, the \textit{binning} operation.

The primary motivation for binning is to provide a common basis for the distinct time-series data.
As a premise, a pair of scalar time-series data is not guaranteed to align in time.
For example, consider two independent runs of the RL agent in an episodic environment with an explicit termination condition other than the time limit.
Depending on the termination condition, the timesteps at which the episodic return is received may differ across runs.
Such misalignment in time hinders time-wise comparisons between pairs of time series, underscoring the necessity of a common basis for comparison.
Binning provides this basis by dividing the time space into non-overlapping intervals (bins) and aggregating each time series data on a per-bin basis.
Formally, let us define a scalar time series data collected over discrete timesteps \(1\) through \(N\) as \(\mathcal{Q} = \{q_t \in \mathbb{R} | t \in \mathcal{I}\}\), where \(\mathcal{I} \subseteq [N]\).
Note that here we denote the set of positive integers \(\{1, \dots, N\}\) for some positive integer \(N\) as \([N]\).
Then, the binned statistics of \(\mathcal{Q}\) is given as \(\bar{\mathcal{Q}} = \left\{f\left(\{q_i\}_{i \in \mathcal{I} \cap \{(b-1) \cdot C + 1, \dots, b \cdot C\}}\right)  | q_i \in \mathcal{Q}, b \in [B]\right\}\).
Here, \(B\) denotes the number of bins, \(C = \left\lceil \frac{N}{B} \right\rceil\) is the size of each bin, and \(f: 2^\mathcal{Q} \rightarrow \mathbb{R}\) is the aggregation function.
The default value of \(B\) is \(100\) and the default for the function \(f\) is a sample mean, if not explicitly specified.

\Cref{fig:synth_bin} shows the benefit of binning via a synthetic example.
Suppose we collect two series of scalar data over the timesteps \(1\) through \(100\).
Assume we collect each series of data \(\mathcal{Q}_1\) and \(\mathcal{Q}_2\) at timesteps \(\mathcal{I}_1\) and \(\mathcal{I}_2\), respectively.
For example, let \(\mathcal{Q}_1 := \{ 0.5 t | t \in \mathcal{I}_1\}\), where \(\mathcal{I}_1 := \{20 x + 5 y | x \in \{0, \dots, 4\}, y \in [3]\}\).
Similarly, let \(\mathcal{Q}_2 := \{ 0.25 t | t \in \mathcal{I}_2\}\), where \(\mathcal{I}_2 := \{20 x + 5 y + 2 | x \in \{0, \dots, 4\}, y \in [3]\}\).
\Cref{fig:synth_bin:pre_bin} depicts both \(\mathcal{Q}_1\) and \(\mathcal{Q}_2\).
Since both timesteps of when the datapoints are collected do not align, comparing a single datapoint from \(\mathcal{Q}_1\) to another in \(\mathcal{Q}_2\) (and vice versa) is non-trivial.
Now, consider dividing the whole timeline from \(1\) to \(100\) into \(5\) non-overlapping bins, as presented by the alternating shaded regions in \Cref{fig:synth_bin:pre_bin}.
By taking an average within each bin, we get \(\bar{\mathcal{Q}}_1 := \left\{10 b - 5 | b \in [5]\} \right\}\) and \(\bar{\mathcal{Q}}_2 := \left\{5 b - 2 | b \in [5]\right\}\).
As it is observable from \Cref{fig:synth_bin:binned} and the range of values the variable \(b\) takes in both \(\bar{\mathcal{Q}}_1\) and \(\bar{\mathcal{Q}}_2\), binning process provides a common basis to compare different time-series data in side-by-side manner.

\section{Order Statistics is a Consistent Estimator}
\label{sec:appendix:os_consistency}

In the main text, we mentioned that the sample percentile is a consistent estimator.
Here, we provide further mathematical details on how the sample percentile (or, more broadly, order statistics) are consistent with their population values.
Let \(X_1, X_2, \dots, X_n\) be a sequence of \(n\) random variables, and its ascendedly sorted counterpart as \(X_{(1)} \leq X_{(2)} \leq \dots \leq X_{(n)}\).
Also, let \(q := \frac{p}{100}\) for the convenience.
Then, the sample \(p\)-th percentile is \(X_{\left(\left[nq\right]\right)}\), where \([\cdot]\) denotes the nearest integer of a given value.
When a target distribution is absolutely continuous, the asymptotic distribution of the sample order statistic \(X_{\left(\left[nq\right]\right)}\) at quantile \(q\) follows the Gaussian distribution:
\begin{align}
    X_{\left(\left[nq\right]\right)} \sim \mathcal{N}\left(\xi_q, \frac{q(1-q)}{n f(\xi_q)^2}\right),
\end{align}
\noindent
where \(\xi_q\) is the population quantile value at quantile \(q\), and \(f\) is a true pdf (for elementary proof, see \citealt{walker_1968_distOfQuantiles} for example).
This property also implies that the sample order statistic \(X_{\left(\left[nq\right]\right)}\) is a consistent estimator of population quantile \(\xi_{q}\) \citep{arnold_2008_orderStats}.
Together, these theoretical results strengthen the suitability of IPR to represent performance spread.

\section{Functional Boxplot}
\label{chap:appendix:func_boxplot}

Functional boxplot (FB) is an order statistics-based method for plotting time-series (functional) data \citep{sun_2011_functionalBoxplot}.
In standard practice, it highlights the median curve, shades \(50\%\) central region (called \(50\%\) envelope), and emphasizes the outliers if applicable.
An FB is a generalization of an ordinary boxplot to time-series data.
In this paper, we use FB as a competitor of the other visualization methods.

For the succeeding arguments, we first introduce the notations and basic concepts related to an FB.
Let \(\mathcal{Y} := \left\{\left\{y_i(t) | t \in \mathcal{I} \right\} | i=1, \dots, n\right\}\) denote a set of time-series data, where \(y_i\) is a real function, \(n\) is a number of time-series data and \(\mathcal{I}\) is an interval in \(\mathbb{R}\).
Given an arbitrary real function \(y(t)\), its graph is the subset of the plane \(G(y) := \left\{ (t, y(t)) | t \in \mathcal{I} \right\}\).
Also, the band in \(\mathbb{R}^2\) delimited by \(K\) curves from \(\mathcal{Y}\) is given as
\[B(y_{i_1}, \dots, y_{i_K}) := \left\{ (t, x(t)) \;\middle| t \in \mathcal{I}, \min_{k=1, \dots, K} y_{i_k}(t) \leq x(t) \leq \max_{k=1, \dots, K} y_{i_k}(t)\right\}.\]
\noindent
With these notations, we now proceed to the method of constructing an FB.

FB employs the concept of band depth (BD) to order time-series data \citep{lopez_2009_bd}.
Intuitively, BD provides a ranking of how close each time-series data in \(\mathcal{Y}\) is to the center of all data.
Hence, when time-series data is sorted according to BD, they are ordered in a center-to-outwards fashion.
For instance, the deepest data (with the highest BD) is closest to the center and therefore a median among the given set of time-series data.
Mathematically, canonical BD is defined as the fraction of bands determined by \(k\) sample curves in \(\mathcal{Y}\) that contains the whole graph of \(y(t)\):
\begin{align}
    &\textnormal{BD}_{n, K}(y) = \sum_{k=2}^{K} \textnormal{BD}^{(k)}_{n}(y),\\
    &\textnormal{where } \textnormal{BD}^{(k)}_{n}(y) = \begin{pmatrix}
        n\\
        k
    \end{pmatrix}^{-1} \sum_{1 \leq i_1 < i_2 < \dots < i_k \leq n} I\left\{ G(y) \subseteq B(y_{i_1}, \dots, y_{i_k}) \right\}, \nonumber
\end{align}
\noindent
where \(I\{\cdot\}\) is an indicator function, \(BD^{(k)}_{n}(y)\) is a band depth of given curve \(y\) derived from \(k\) curves out of \(n\) curves, and \(BD_{n,K}(y)\) is an overall band depth of given curve \(y\).
Although this is the canonical form of BD, we use a more flexible variant called modified BD (MBD).
Instead of the indicator function in BD, MBD measures the proportion of time \(t\) that \(y(t)\) resides within a band.
Suppose
\begin{align*}
    A_k(y) &\equiv A(y | y_{i_1}, y_{i_2}, \dots, y_{i_k})\\
        &\equiv \left\{ t \in \mathcal{I} \;\middle| \min_{r = i_1, \dots, i_k} y_r(t) \leq y(t) \leq \max_{r = i_1, \dots, i_k} y_r(t) \right\}.
\end{align*}
\noindent
Then, MBD is formally given as
\begin{align}
    &\textnormal{MBD}_{n, K}(y) = \sum_{k=2}^{K} \textnormal{MBD}^{(k)}_{n}(y),\\
    &\textnormal{where } \textnormal{MBD}^{(k)}_{n}(y) = \begin{pmatrix}
        n\\
        k
    \end{pmatrix}^{-1} \sum_{1 \leq i_1 < i_2 < \dots < i_k \leq n} \lambda_r\left( A_k(y | y_{i_1}, y_{i_2}, \dots, y_{i_k}) \right), \nonumber
\end{align}
\noindent
where \(\lambda_r(A_k(y)) = \frac{\lambda(A_k(y))}{\lambda(\mathcal{I})}\) and \(\lambda\) is the Lebesgue measure on \(\mathcal{I}\).
While \(K\) can take any integer from \(2\) to \(n\), we use \(K = 2\) by following \citet{sun_2011_functionalBoxplot}.

\begin{figure*}[t!]
    \centering
    \includegraphics[width=0.5\textwidth]{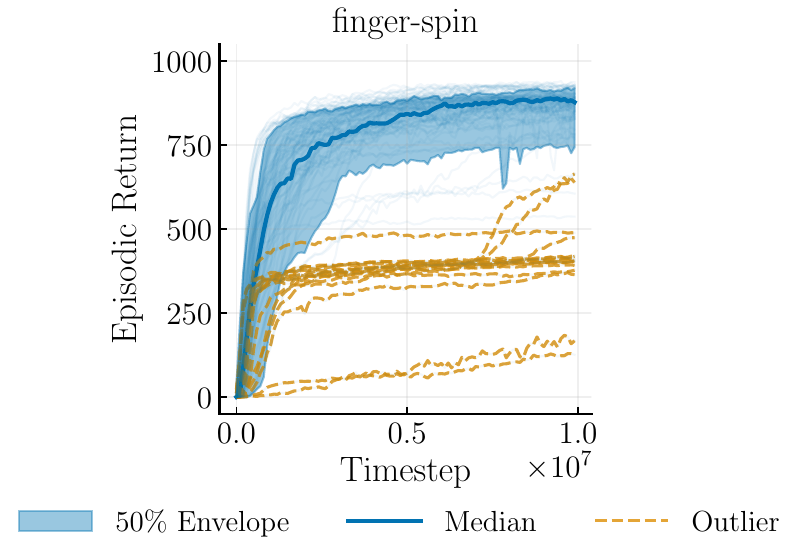}
    \caption[Exemplar FB.]
    {
        FB applied to the learning curves of PPO on \texttt{finger-spin} task.
        The necessary components of the FB are highlighted.
        Namely, the median, \(50\%\) envelope, and outlier curves.
    }
    \label{fig:func_boxplot_example}
\end{figure*}

Now, all the necessary elements that consist of an FB ---median, \(50\%\) envelope, and outliers--- can be determined by using MBD for each curve in \(\mathcal{Y}\).
Suppose all curves in \(\mathcal{Y}\) are ranked according to its corresponding \(\textnormal{MBD}_{n, 2}\), and let us denote a curve with \(j\)-th largest \(\textnormal{MBD}_{n,2}\) as \(y_{[j]}\).
Then, the median curve according to the MBD is \(y_{[1]}\) by definition.
We refer to this median curve as the FB median curve.
A \(50\%\) envelope can be constructed by taking the time-wise minimum and maximum over the \(50\%\) of deepest curves:
\[C_{0.5} := \left\{ (t, y(t)) \;\middle| t \in \mathcal{I}, \min_{r = 1, \dots, \lceil \frac{n}{2} \rceil} y_{[r]}(t) \leq y(t) \leq \max_{r = 1, \dots, \lceil \frac{n}{2} \rceil} y_{[r]}(t) \right\}.\]
Lastly, the curves are classified as outliers if the value of the curve falls outside the fence at any point in time.
The fence is determined by inflating the \(50\%\) envelope by \(1.5\) times the range of itself, which is analogous to the \(1.5\) times IQR criterion for outlier detection in a canonical boxplot.
By plotting these together with appropriate graphing styles, we get an FB (see \Cref{fig:func_boxplot_example} for example).

\newpage
\section{Hyperparameters For The Experiments}
\label{sec:appendix:hp}

\begin{table*}[htbp!]
    \footnotesize
    \centering
    \caption{\centering MNIST Configurations.}
    \label{table:mnist_config}
    \begin{tabular}{@{}lll@{}}
        \toprule
        \textbf{Parameter} & \textbf{Configuration} \\
        \midrule
        Total Epochs & \(200\) \\
        Minibatch Size & \(256\) \\
        Augmentations & \texttt{Flatten} \\
        Network Architecture & Fully-connected NN \\
        Hidden Layer Dims & \(256\) \\
        Number of Hidden Layers & \(2\) \\
        Activation Function & ReLU \\
        Optimizer & Adam \\
        Step-size & \(3 \times 10^{-4}\) \\
        \bottomrule
    \end{tabular}
\end{table*}

\begin{table*}[htbp!]
    \footnotesize
    \centering
    \caption{\centering CIFAR-\(10\) Configurations.}
    \label{table:cifar_config}
    \begin{tabular}{@{}lll@{}}
        \toprule
        \textbf{Parameter} & \textbf{Configuration} \\
        \midrule
        Total Epochs & \(200\) \\
        Minibatch Size & \(128\) \\
        Augmentations & \texttt{RandomResizedCrop(size=\(32\), scale=\((0.8, 1.0\)))} \\
            & \texttt{RandomHorizontalFlip} \\
            & \texttt{Normalize(mean=\(((0.4914, 0.4822, 0.4465)\), std=\((0.247, 0.243, 0.261)\))} \\
        Network Architecture & ResNet\(18\) \\
        Activation Function & \(ReLU\) \\
        Optimizer & Adam \\
        Step-size & \(3 \times 10^{-4}\) \\
        Weight Decay & \(10^{-4}\) \\
        Gradient Clipping & \(0.1\) \\
        \bottomrule
    \end{tabular}
\end{table*}

\begin{table*}[htbp!]
    \footnotesize
    \centering
    \caption{\centering Pascal VOC Configurations.}
    \label{table:pascalvoc_config}
    \begin{tabular}{@{}lll@{}}
        \toprule
        \textbf{Parameter} & \textbf{Configuration} \\
        \midrule
        Total Epochs & \(200\) \\
        Minibatch Size & \(32\) \\
        Network Architecture & YOLO\(11\) \citep{yolo11_ultralytics} \\
        Optimizer & Adam \\
        Step-size & \(3 \times 10^{-4}\) \\
        \bottomrule
    \end{tabular}
\end{table*}

\begin{table*}[htbp!]
    \footnotesize
    \centering
    \caption{\centering PPO Configurations.}
    \label{table:ppo_configs}
    \begin{tabular}{@{}lll@{}}
        \toprule
        \textbf{Parameter} & \textbf{Configuration} \\
        \midrule
        Total timesteps (\(T\))   & \(10^7\) \\
        Parameter Update Frequency (\(B\)) & \(8192\) \\
        Number of Epochs (\(N\))  & \(10\) \\
        Minibatch Size (\(M\))    &  \(256\) \\
        Clipping Parameter (\(\epsilon\)) & \(0.2\) \\
        Value Loss Coefficient  & \(0.5\) \\
        Maximum Gradient Norm & \(0.5\) \\
        GAE \(\gamma\)     &  \(0.99\) \\
        GAE \(\lambda\)    &  \(0.95\) \\
        Network Architecture (\(\btheta\) \& \(\bpsi\)) & Fully-connected NN \\
        Hidden Layer Dims (\(\btheta\) \& \(\bpsi\)) & \(256\) \\
        Number of Hidden Layers (\(\btheta\) \& \(\bpsi\)) & \(2\) \\
        Activation Function (Actor \& Critic) & \texttt{Tanh} \\
        Optimizer (Actor \& Critic) & Adam \\
        Step-size for Actor \& Critic (\(\eta\)) & \(3 \times 10^{-4}\) \\
        \bottomrule
    \end{tabular}
\end{table*}

\begin{table*}[htbp!]
    \footnotesize
    \centering
    \caption{\centering SAC Configurations.}
    \label{table:sac_configs}
    \begin{tabular}{@{}lll@{}}
        \toprule
        \textbf{Parameter} & \textbf{Configuration} \\
        \midrule
        Total timesteps (\(T\))   & \(10^6\) \\
        Length of Initial Sampling Phase (\(B\))        & \(5 \times 10^3\) \\
        Buffer Size & \(10^6\) \\
        Minibatch Size (\(M\))    &  \(256\) \\
        Frequency of Policy Update (\(N_p\)) & \(2\) \\
        Frequency of Target Update (\(N_t\)) & \(1\) \\
        \(\tau\)           & \(5 \times 10^{-3}\) \\
        Minimum std.\ deviation in log-scale & \(-5\) \\
        Maximum std.\ deviation in log-scale & \(2\) \\
        Network Architecture (\(\btheta\) \& \(\{\bpsi\}_{i=1,2}, \{\bpsi_{\textnormal{targ}_i}\}_{i=1,2}\)) & Fully-connected NN \\
        Hidden Layer Dims (\(\btheta\) \& \(\{\bpsi\}_{i=1,2}, \{\bpsi_{\textnormal{targ}_i}\}_{i=1,2}\)) & \(256\) \\
        Number of Hidden Layers (\(\btheta\) \& \(\{\bpsi\}_{i=1,2}, \{\bpsi_{\textnormal{targ}_i}\}_{i=1,2}\)) & \(2\) \\
        Activation Function (Actor \& Critic) & \texttt{ReLU} \\
        Optimizer (Actor \& Critic) & Adam \\
        Step-size for Actor \& Critic (\(\eta\)) & \(3 \times 10^{-4}\) \\
        Initial entropy coefficient (\(\alpha\)) & \(1.0\) \\
        Optimizer (Entropy Coefficient) & Adam \\
        Step-size (Entropy Coefficient) & \(10^{-4}\) \\
        Target Entropy (\(\mathcal{H}\))    & \(|\mathcal{A}|\)\\
        \bottomrule
    \end{tabular}
\end{table*}

\begin{table*}[htbp!]
    \footnotesize
    \centering
    \caption{\centering TD-MPC Hyperparameter Configurations (from \citealp{hansen_2022_TDMPC}).}
    \label{table:tdmpc_configs}
    \begin{tabular}{@{}lll@{}}
        \toprule
        \textbf{Parameter} & \textbf{Configuration} \\
        \midrule
        Total timesteps (\(T\))   & \(5 \times 10^5\) \\
        Discount Factor (\(\gamma\)) & \(0.99\) \\
        Seed Steps & \(5000\) \\
        Replay buffer size & Unlimited \\
        Sampling Technique & PER (\(\alpha=0.6, \beta=0.4\)) \\
        Planning Horizon (H) & \(5\) \\
        Initial Parameters (\(\mu^0, \sigma^0\)) & \((0, 2)\) \\
        Population Size & \(512\) \\
        Elite Fraction & \(64\) \\
        Iterations & \(12\) (Humanoid) \\
            & \(8\) (Dog) \\
            & \(6\) (Otherwise) \\
        Policy Fraction & \(5\%\) \\
        Number of Particles & \(1\) \\
        Momentum coefficient & \(0.1\) \\
        Temperature (\(\tau\)) & \(0.5\) \\
        MLP Hidden Size & \(512\) \\
        MLP Activation & ELU \\
        MLP Activation & ELU \\
        Latent Dimension & \(100\) (Humanoid, Dog) \\
            & \(50\) (Otherwise) \\
        Step-size & \(3 \times 10^{-4}\) (Dog) \\
            & \(10^{-3}\) \\
        Optimizer (\(\theta\)) & Adam (\(\beta_1 = 0.99, \beta_2=0.999\) \\
        Temporal Coefficient (\(\lambda\)) & \(0.5\) \\
        Reward Loss Coefficient (\(c_1\)) & \(0.5\) \\
        Value Loss Coefficient (\(c_2\)) & \(0.1\) \\
        Consistency Loss Coefficient (\(c_3\)) & \(2\) \\
        Exploration Schedule (\(\epsilon\)) & \(0.5 \to 0.05\) (\(25\)k steps) \\
        Planning Horizon Schedule & \(1 \to 5\) (\(25\)k steps) \\
        Batch Size & \(2048\) (Dog) \\
            & \(512\) (Otherwise) \\
        Momentum Coefficient (\(\zeta\)) & \(0.99\) \\
        Steps Per Gradient Update & \(1\) \\
        \(\theta^-\) Update Frequency & \(2\) \\
        \bottomrule
    \end{tabular}
\end{table*}

\begin{table}[ht]
    \footnotesize
    \centering
    \caption{\centering TD-MPC2 Hyperparameter Configurations (from \citealp{hansen_2024_TDMPC2}).}
    \label{table:tdmpc2_configs}
    \begin{tabular}{ll}
        \toprule
        \textbf{Hyperparameter} & \textbf{Value} \\
        \midrule
        \textbf{Planning} & \\
        Horizon ($H$) & 3 \\
        Iterations & 6 ($+2$ if $\|\mathcal{A}\| \geq 20$) \\
        Population size & 512 \\
        Policy prior samples & 24 \\
        Number of elites & 64 \\
        Minimum std. & 0.05 \\
        Maximum std. & 2 \\
        Temperature & 0.5 \\
        Momentum & No \\
        \addlinespace
        
        \textbf{Policy prior} & \\
        Log std. min. & $-10$ \\
        Log std. max. & 2 \\
        \addlinespace
        
        \textbf{Replay buffer} & \\
        Capacity & 1,000,000 \\
        Sampling & Uniform \\
        \addlinespace
        
        \textbf{Architecture (5M)} & \\
        Encoder dim & 256 \\
        MLP dim & 512 \\
        Latent state dim & 512 \\
        Task embedding dim & 96 \\
        Task embedding norm & 1 \\
        Activation & LayerNorm + Mish \\
        $Q$-function dropout rate & 1\% \\
        Number of $Q$-functions & 5 \\
        Number of reward/value bins & 101 \\
        SimNorm dim ($V$) & 8 \\
        SimNorm temperature ($\tau$) & 1 \\
        \addlinespace
        
        \textbf{Optimization} & \\
        Update-to-data ratio & 1 \\
        Batch size & 256 \\
        Joint-embedding coef. & 20 \\
        Reward prediction coef. & 0.1 \\
        Value prediction coef. & 0.1 \\
        Temporal coef. ($\lambda$) & 0.5 \\
        $Q$-fn. momentum coef. & 0.99 \\
        Policy prior entropy coef. & $1 \times 10^{-4}$ \\
        Policy prior loss norm. & Moving (5\%, 95\%) percentiles \\
        Optimizer & Adam \\
        Learning rate & $3 \times 10^{-4}$ \\
        Encoder learning rate & $1 \times 10^{-4}$ \\
        Gradient clip norm & 20 \\
        Discount factor & Heuristic \\
        Seed steps & Heuristic \\
        \bottomrule
    \end{tabular}
\end{table}

\begin{table*}[htbp!]
    \footnotesize
    \centering
    \caption{\centering DQN Configurations. Network architecture is written in PyTorch syntax without activation function between layers.}
    \label{table:dqn_configurations}
    \begin{tabular}{@{}lll@{}}
        \toprule
        \textbf{Parameter} & \textbf{Configuration} \\
        \midrule
        Total Timesteps (\(T\))   & \(5 \times 10^7\) \\
        Buffer Size & \(10^6\) \\
        Length of Initial Sampling Phase & \(8 \times 10^4\) \\
        Stacked Frames & \(4\) \\
        Network Architecture & \texttt{Conv2d(4, 32, 8, stride=4)}\\
            & \texttt{Conv2d(32, 64, 4, stride=4)} \\
            & \texttt{Conv2d(64, 64, 3, stride=1)} \\
            & \texttt{Flatten} \\
            & \texttt{Linear(3136, 512)} \\
            & \texttt{Linear(512, 18)} \\
        Activation Function & \texttt{ReLU} \\
        Optimizer  & Adam \\
        Step-size  & \(10^{-4}\) \\
        Parameter Update Frequency & \(4\) \\
        Minibatch Size   &  \(32\) \\
        Gamma (\(\gamma\)) & \(0.99\) \\
        Frequency of Target Update & \(10^3\) \\
        Tau (\(\tau\)) & \(1.0\) \\
        Epsilon Scheduler & Linear \\
        Initial Epsilon Value & \(1.0\) \\
        Final Epsilon Value & \(0.01\) \\
        Sticky Action Probability & \(0.25\) \\
        \bottomrule
    \end{tabular}
\end{table*}

\begin{table*}[htbp!]
    \footnotesize
    \centering
    \caption{\centering Rainbow Configurations. Network architecture is written in PyTorch syntax without activation function between layers.}
    \label{table:rainbow_configurations}
    \begin{tabular}{@{}lll@{}}
        \toprule
        \textbf{Parameter} & \textbf{Configuration} \\
        \midrule
        Total Timesteps   & \(5 \times 10^7\) \\
        Buffer Size & \(10^6\) \\
        Length of Initial Sampling Phase & \(8 \times 10^4\) \\
        Stacked Frames & \(4\) \\
        Network Architecture & \texttt{Conv2d(4, 32, 8, stride=4)}\\
            & \texttt{Conv2d(32, 64, 4, stride=4)} \\
            & \texttt{Conv2d(64, 64, 3, stride=1)} \\
            & \texttt{Flatten} \\
        (for each value and advantage estimation): & \texttt{NoisyLinear(3136, 512)} \\
            & \texttt{NoisyLinear(512, 18)} \\
        Activation Function & \texttt{ReLU} \\
        Optimizer  & Adam \\
        Step-size  & \(6.25 \times 10^{-5}\) \\
        Parameter Update Frequency & \(4\) \\
        Minibatch Size   &  \(32\) \\
        Discount \(\gamma\) & \(0.99\) \\
        Frequency of Target Update & \(10^3\) \\
        Target Update \(\tau\) & \(1.0\) \\
        Sticky Action Probability & \(0.25\) \\
        Number of Atoms & \(51\) \\
        Minimum Value & \(-10\) \\
        Maximum Value & \(10\) \\
        Steps for \(n\)-step Return & \(3\) \\
        Prioritized Experience Replay (PER) \(\alpha\) & \(0.5\) \\
        PER \(\beta\) & \(0.4\) \\
        PER \(\epsilon\) & \(10^{-6}\) \\
        \bottomrule
    \end{tabular}
\end{table*}

\clearpage
\section{Supervised Learning Tasks}
\label{sec:appendix:sl_learning_curves}

\begin{figure*}[tbh!]
    \centering
    \begin{subfigure}[tbh]{0.33\textwidth}
        \centering
        \includegraphics[width=\textwidth]{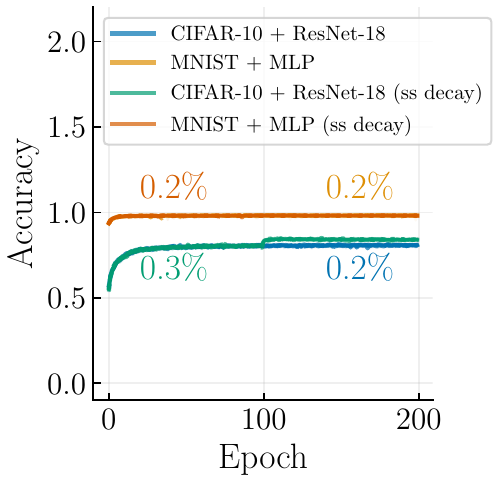}
        \caption{Test Accuracies}
        \label{fig:sl_varioation:sl_acc}
    \end{subfigure}
    \hspace{0.01\textwidth}
    \begin{subfigure}[tbh]{0.33\textwidth}
        \centering
        \includegraphics[width=\textwidth]{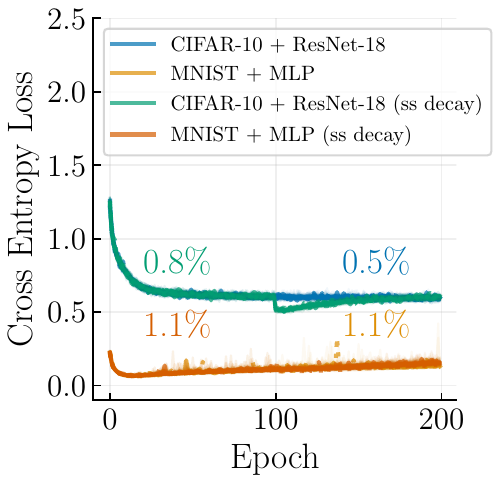}
        \caption{Test Losses}
        \label{fig:sl_variation:sl_loss}
    \end{subfigure}
    \caption
    {
        Variation of test accuracies and losses in standard supervised learning (SL) settings.
        Plots presents \textbf{(a)}.\ test accuracy and \textbf{(b)}.\ test loss of \(100\) independent runs of MLP on MNIST and ResNet-\(18\) on CIFAR-\(10\).
        By default, step-size for the optimizer does not change throughout the learning.
        The plots also provide the results with step-size decay from \(3\times 10^{-4}\) to \(3\times 10^{-5}\) at the \(100\)th epoch (labeled as ``(ss decay)" in the legends).
        While some loss curves exhibit relatively high variation, many sets of curves achieve low variation.
    }
    \label{fig:sl_variation}
\end{figure*}

\begin{figure*}[tbh!]
    \centering
    \includegraphics[width=0.95\textwidth]{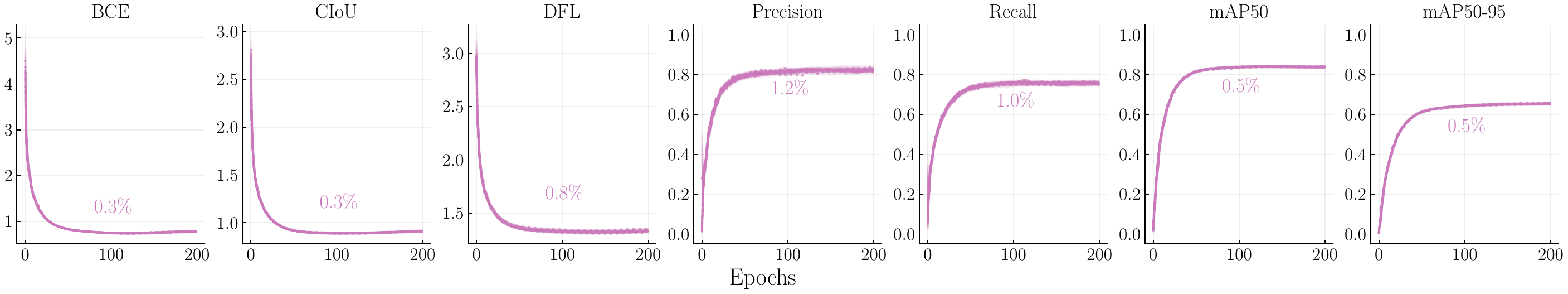}
    \caption{
        Metrics across \(100\) independent YOLO runs on Pascal VOC dataset.
        All metrics are computed on the test data at the end of each epoch, and curves are plotted using RPH.
        For all the metrics, variation over runs is mostly unrecognizable.
    }
    \label{fig:sl_variation_large}
\end{figure*}

Here, we empirically show that SL algorithms tend to exhibit only minor performance variability.
We consider three SL tasks: MNIST with an MLP, CIFAR-\(10\) with a ResNet-\(18\), and Pascal VOC with a YOLO \citep{yolo11_ultralytics}.
All experiments runs \(100\) independent runs, where each lasting \(200\) epochs.
Additionally, all the experiments use Adam optimizer for all experiments with step-size \(3 \times 10^{-4}\).
We set mini-batch size to \(256\) for the first two tasks and \(32\) for the last.
Also, for the first two tasks, we conduct two variants of the experiment, one with step-size annealing and one without.
With annealing, the step-size drops to \(10\%\) of its original value at the \(50\%\) point of the total number of epochs.
For the other hyperparameters, see \Cref{table:mnist_config,table:cifar_config,table:pascalvoc_config}.

\Cref{fig:sl_variation} reports test accuracies and losses over training epochs for the first two tasks in RPH format.
We observe that the spread in test accuracies is nearly imperceptible across all experiment configurations.
Note that the majority of the accuracy curves of each variant of the experiment overlap.
Although the accuracy is wider than the test losses, the spread of test losses is still small.
In particular, the MNIST experiments exhibit almost no fluctuation across independent runs.
While CIFAR-\(10\) experiments show relatively wider variation in test losses, its degree is relatively small when compared to those with deep RL.
Notably, step-size annealing increases the loss and reduces its variation in CIFAR-\(10\) experiments.
This suggests that step-size annealing promotes greater learning stability but does not necessarily improve the loss.
Despite such loss behavior, step-size annealing improves test accuracy in CIFAR-\(10\) experiments.
Such behavior may occur for several reasons, such as low logits for correct predictions.
Nevertheless, the variations over independent runs tend to be small in the first two SL tasks.

A similar conclusion also holds for the YOLO on the Pascal VOC dataset.
Curves in \Cref{fig:sl_variation_large} represent various test metric values.
The first three correspond to different loss functions: binary cross-entropy (BCE), complete intersection over union (CIoU), and distribution focal loss (DFL).
The remaining four are the evaluation metrics: precision, recall, mean average precision \(50\) (mAP\(50\)), and mAP\(50\)-\(95\).
Despite its task complexity, YOLO yields nearly identical values across runs in all the metrics.
This further supports the claim of lower performance variation in SL tasks.

\newpage
\section{All the Learning Curves for Default Experiments}
\label{sec:appendix:learning_curves}

\begin{figure*}[hbt!]
    \centering
    \begin{subfigure}[b]{0.46\textwidth}
        \centering
        \includegraphics[width=\textwidth]{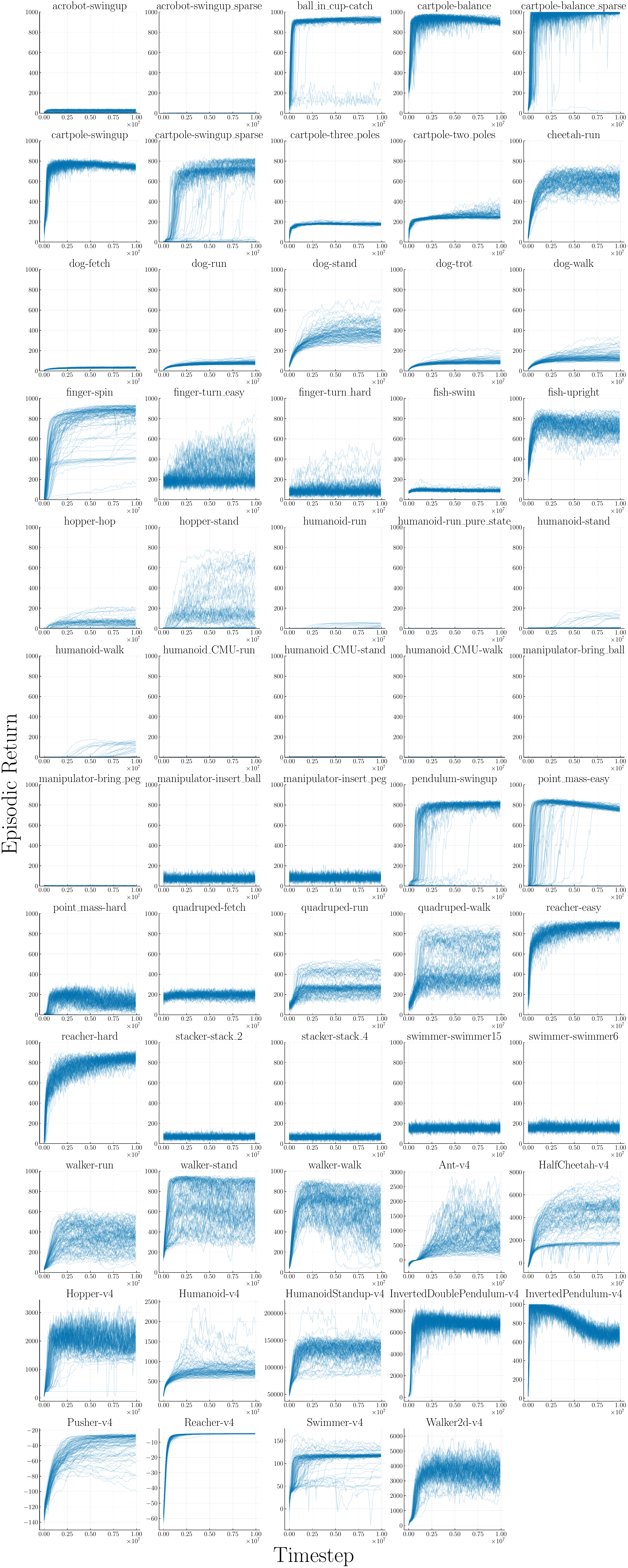}
        \caption{Default PPO Learning Curves}
        \label{fig:baseline_perf_summary:ppo:raw}
    \end{subfigure}
    \hfill
    \begin{subfigure}[b]{0.46\textwidth}
        \centering
        \includegraphics[width=\textwidth]{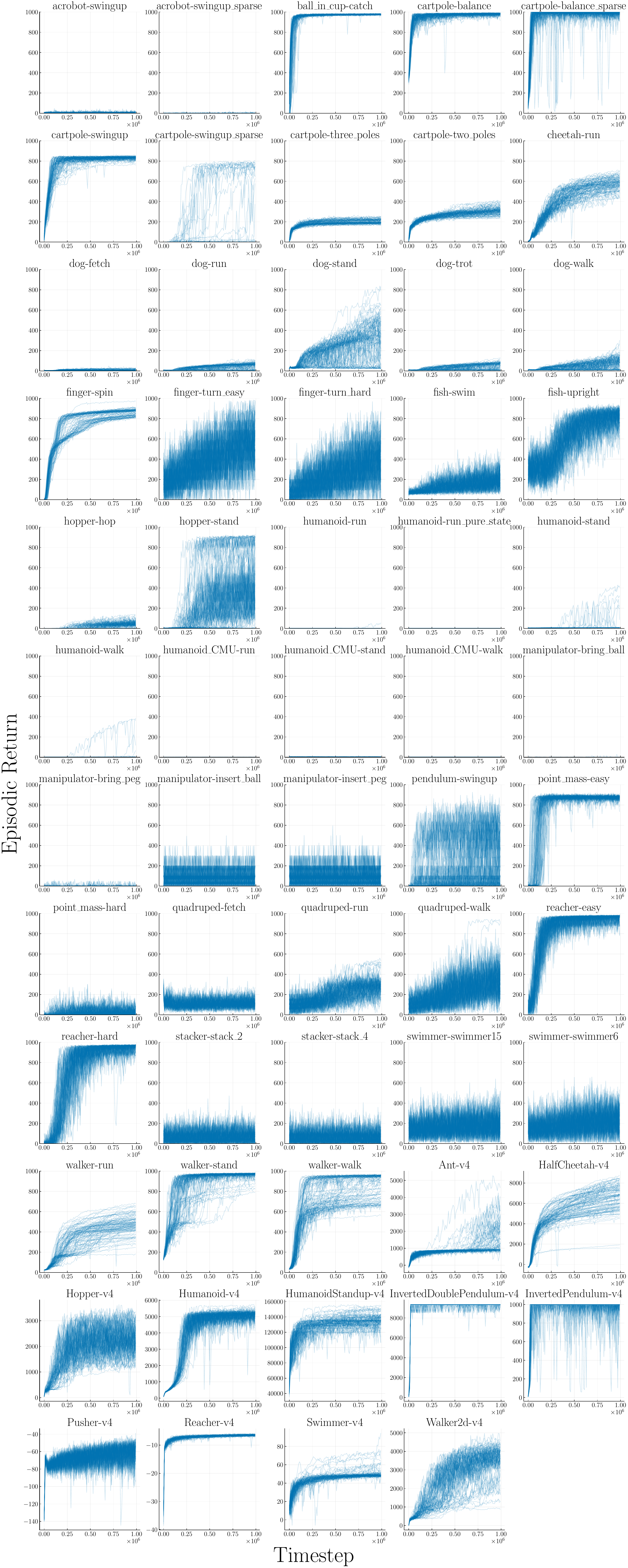}
        \caption{Default SAC Learning Curves}
        \label{fig:baseline_perf_summary:sac:raw}
    \end{subfigure}
    \caption
    {
        Learning curves for PPO and SAC.
        Each subplot (the small boxes) corresponds to one robotic control task.
        Results are obtained by running \(100\) independent runs.
    }
    \label{fig:baseline_perf_raw}
\end{figure*}

\newpage
\section{Performance Variation and Median Bar Plots from Case Studies}
\label{sec:appendix:comp_bar_plots}

\begin{figure*}[hbt]
    \centering
    \begin{subfigure}{0.48\linewidth}
        \includegraphics[width=\linewidth]{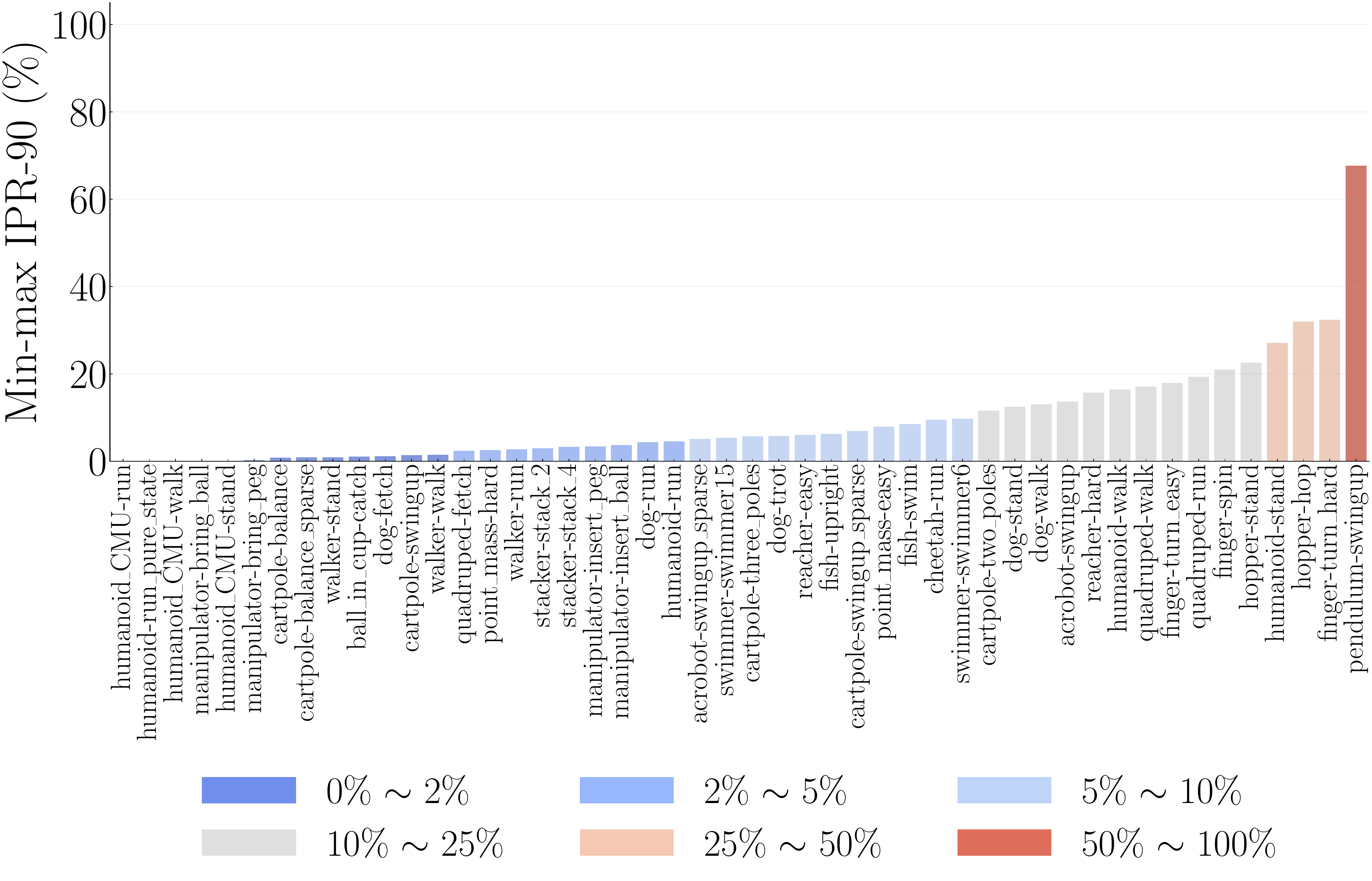}
        \caption{TD-MPC Min-max IPR-\(90\)}
    \end{subfigure}
    \hfill
    \begin{subfigure}{0.48\linewidth}
        \includegraphics[width=\linewidth]{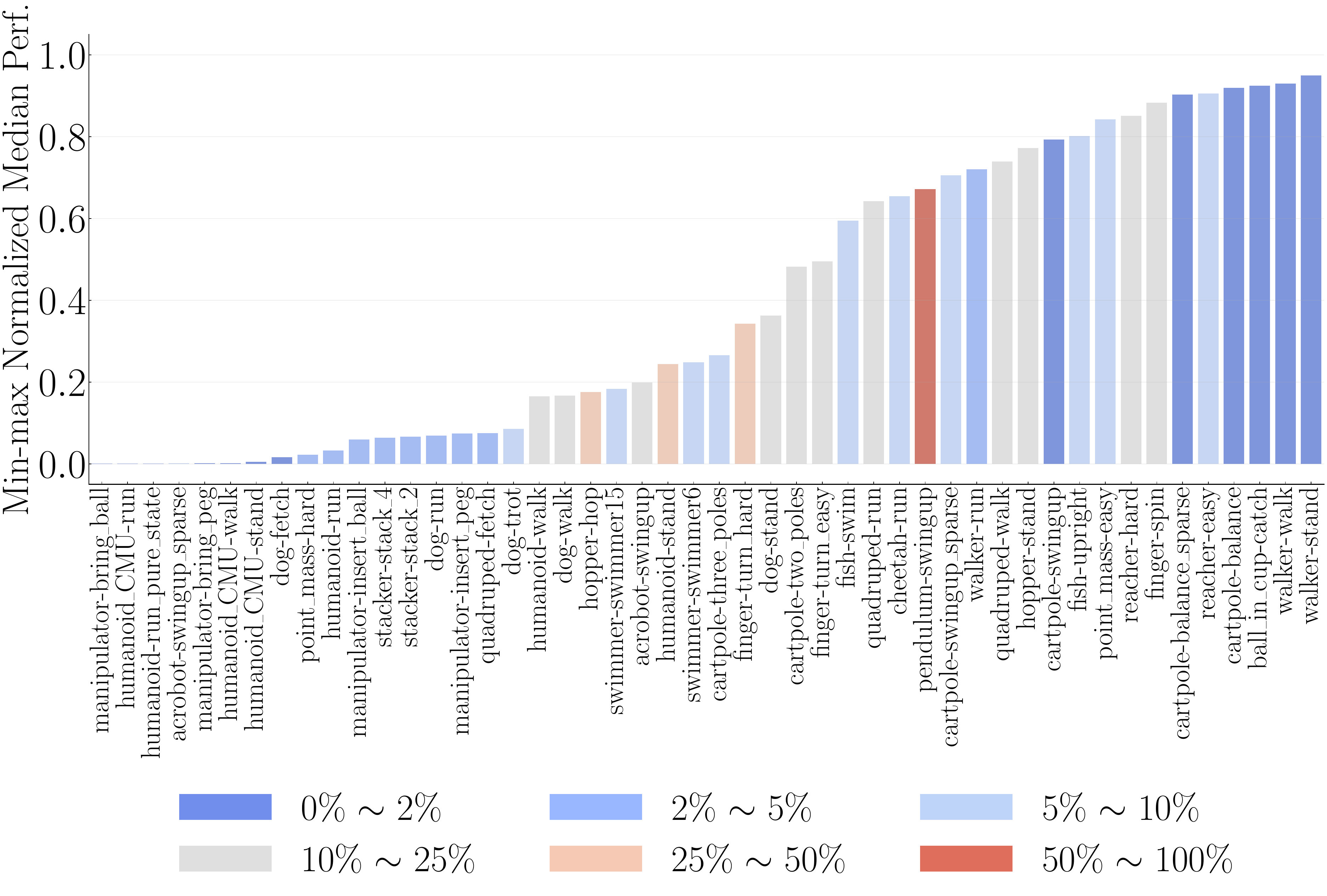}
        \caption{TD-MPC Median}
    \end{subfigure}\\
    \begin{subfigure}{0.48\linewidth}
        \includegraphics[width=\linewidth]{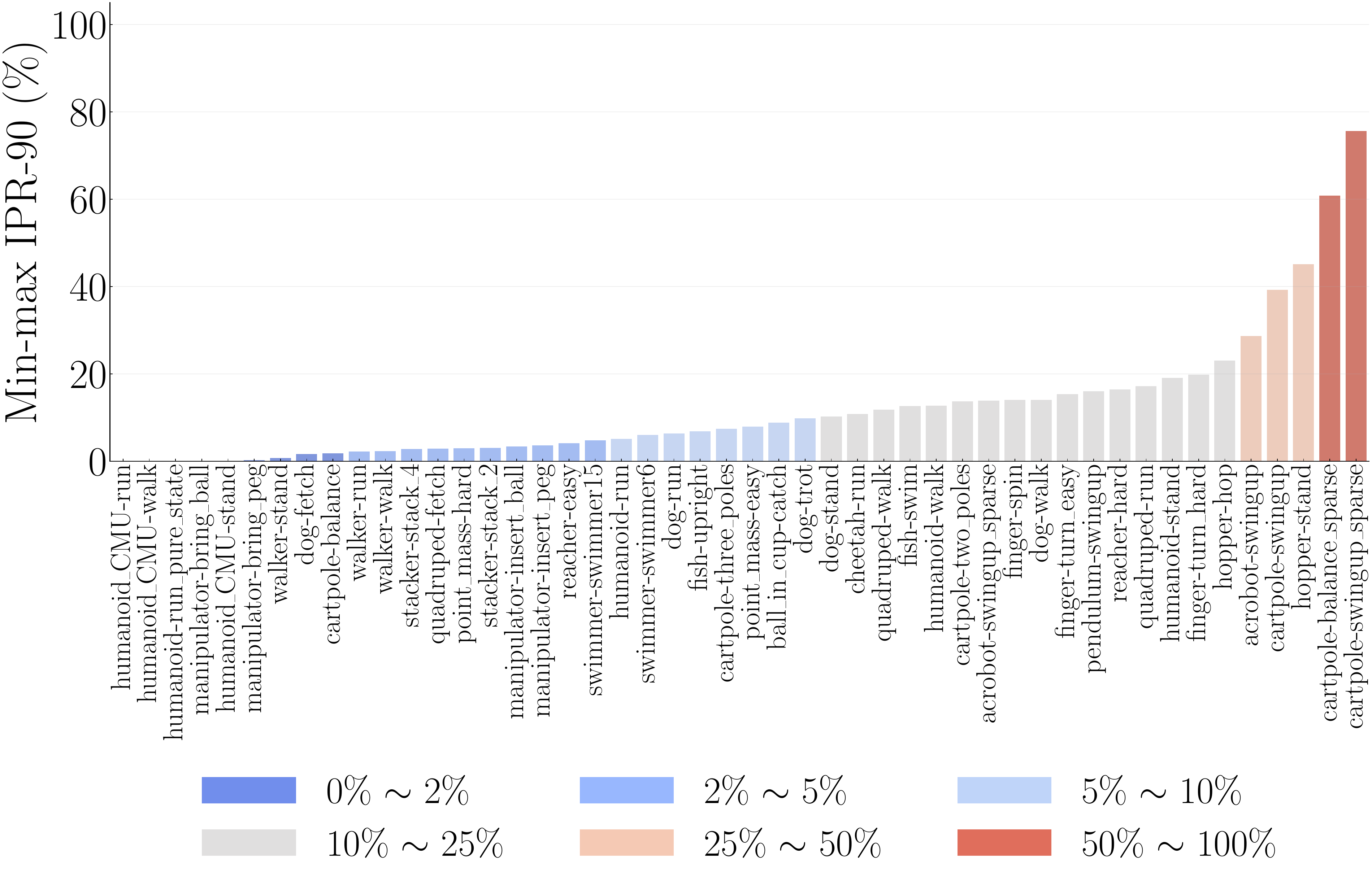}
        \caption{TD-MPC2 Min-max IPR-\(90\)}
    \end{subfigure}
    \hfill
    \begin{subfigure}{0.48\linewidth}
        \includegraphics[width=\linewidth]{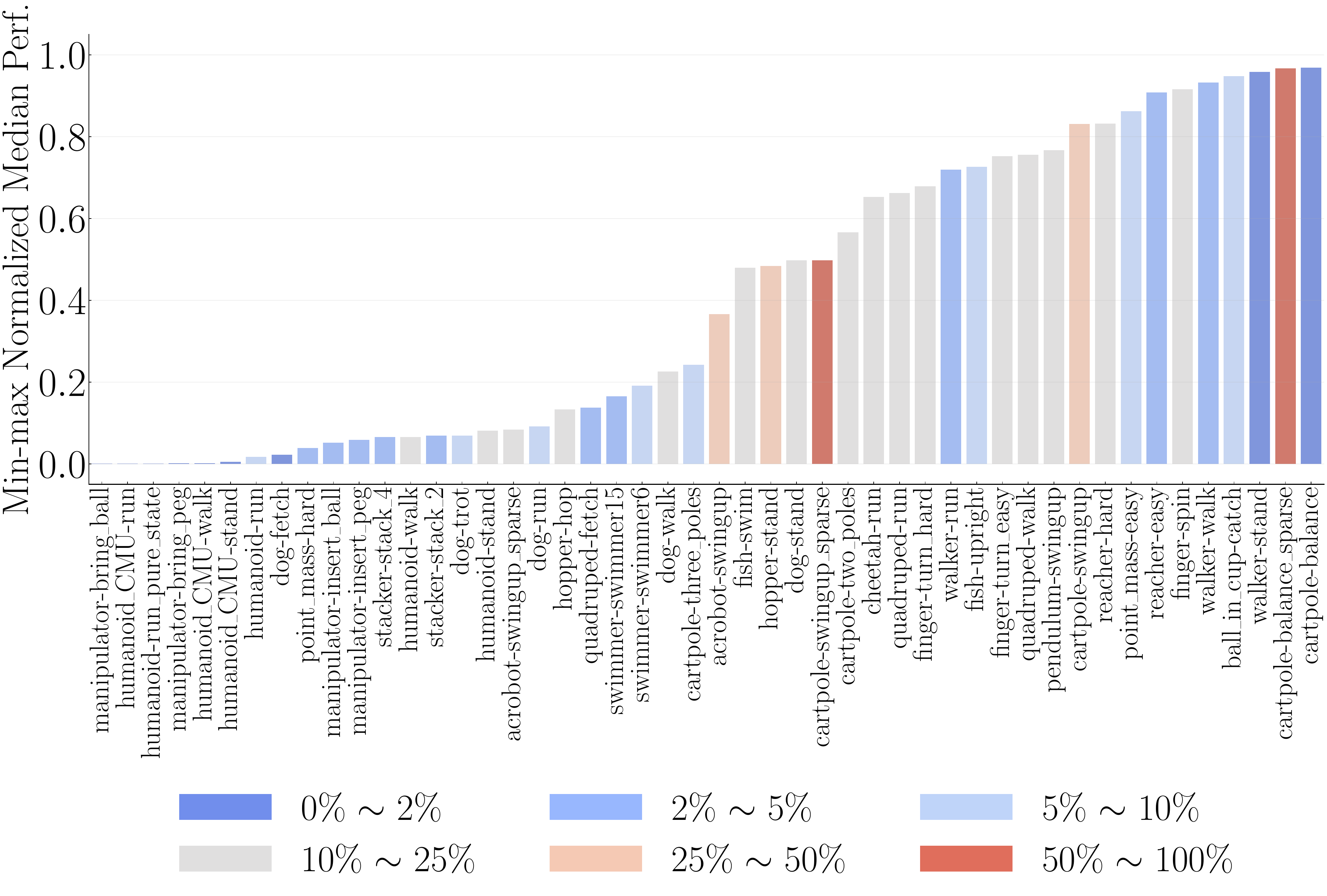}
        \caption{TD-MPC2 Median}
    \end{subfigure}
    \caption{
        Bar plots for performance variation and median performance of TD-MPC and TD-MPC2 experiments.
    }
    \label{fig:tdmpc_var_and_med}
\end{figure*}

\newpage
\begin{figure*}[hbt]
    \centering
    \begin{subfigure}[b]{0.4\textwidth}
        \centering
        \includegraphics[width=\textwidth]{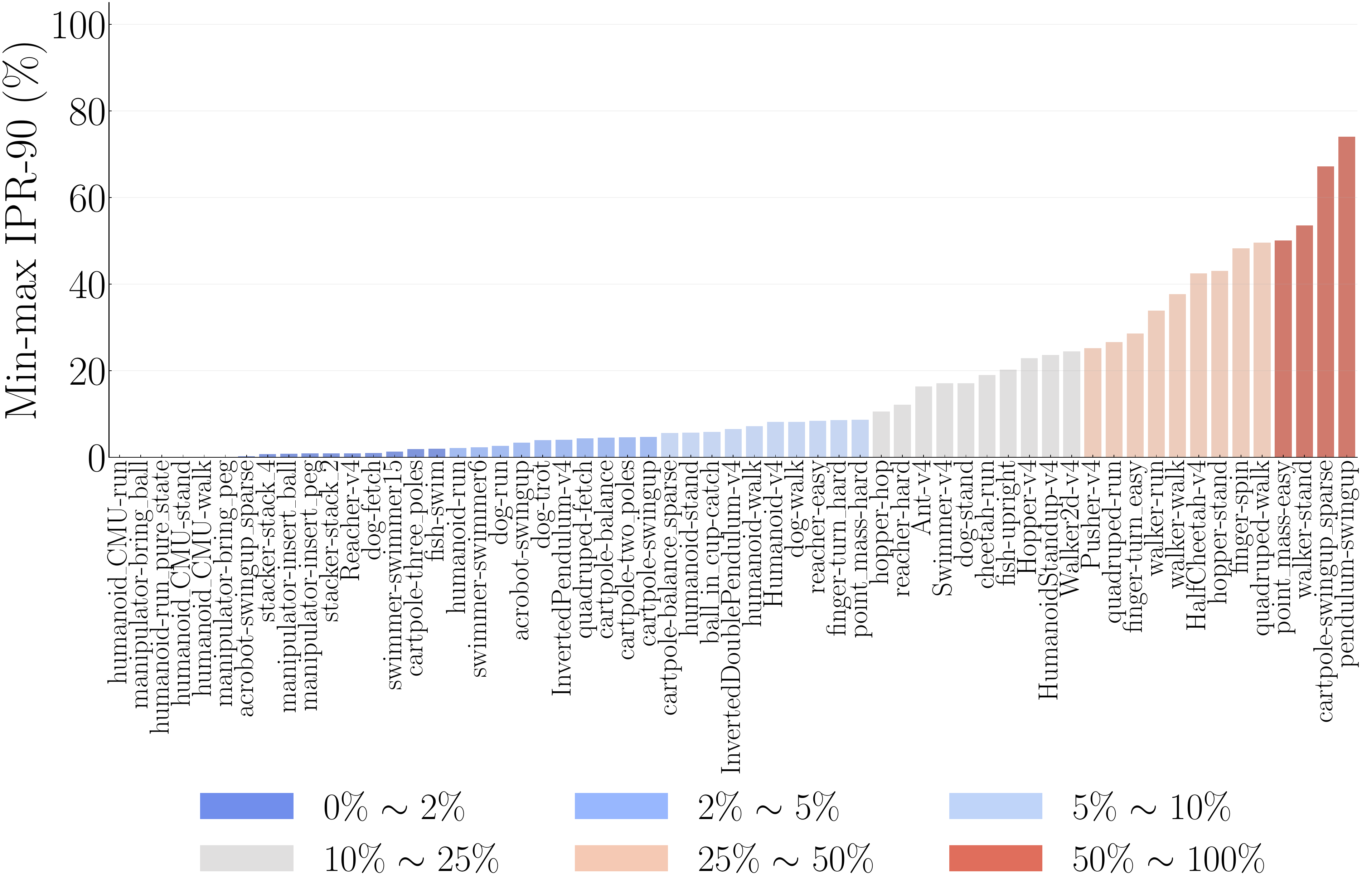}
        \caption{Default IPR-\(90\)}
        \label{fig:ppo_comp_bar_plots:default:ipr}
    \end{subfigure}
    \hspace{0.05\textwidth}
    \begin{subfigure}[b]{0.4\textwidth}
        \centering
        \includegraphics[width=\textwidth]{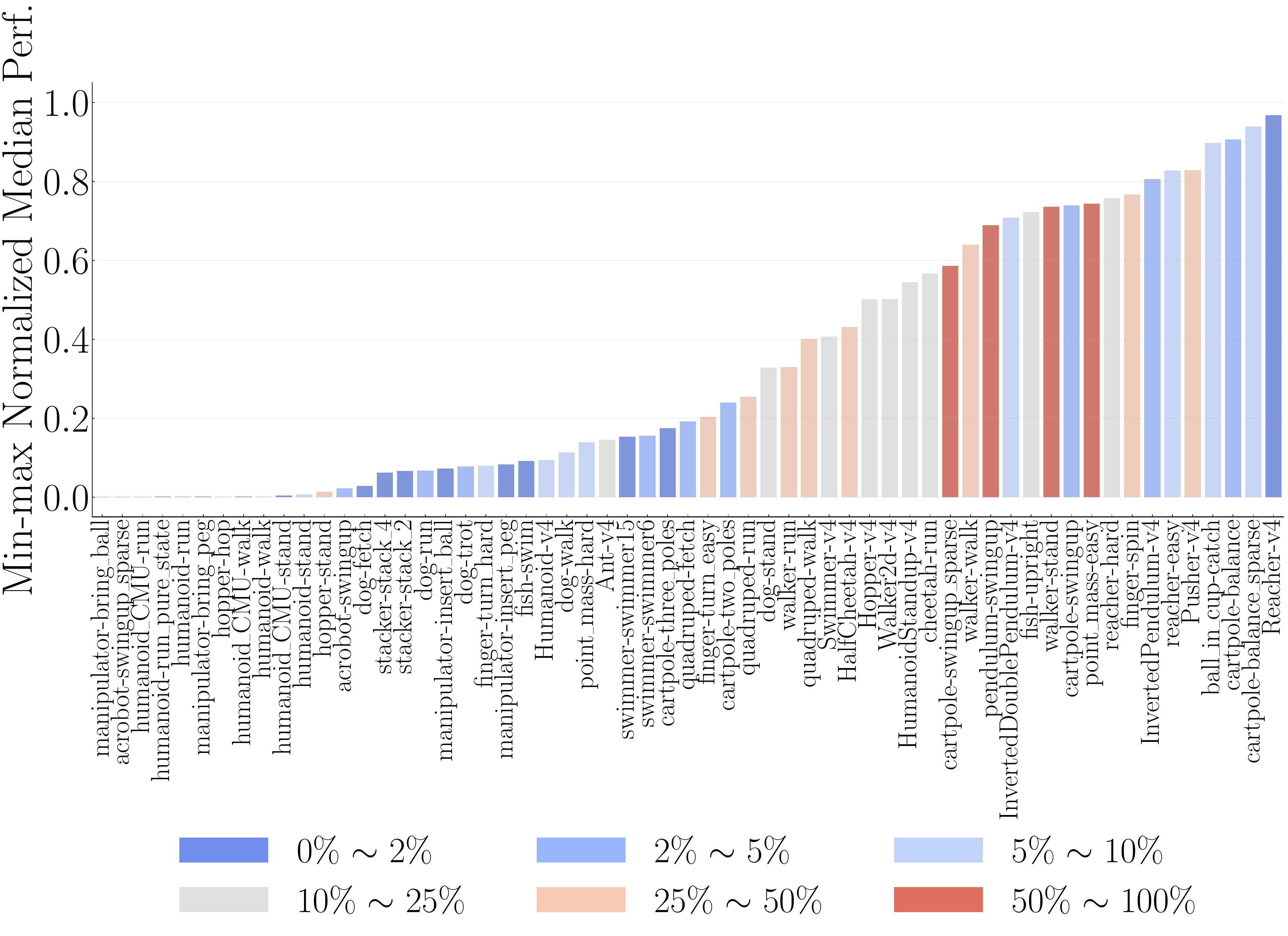}
        \caption{Default Median}
        \label{fig:ppo_comp_bar_plots:default:med}
    \end{subfigure}\\
    \begin{subfigure}[b]{0.4\textwidth}
        \centering
        \includegraphics[width=\textwidth]{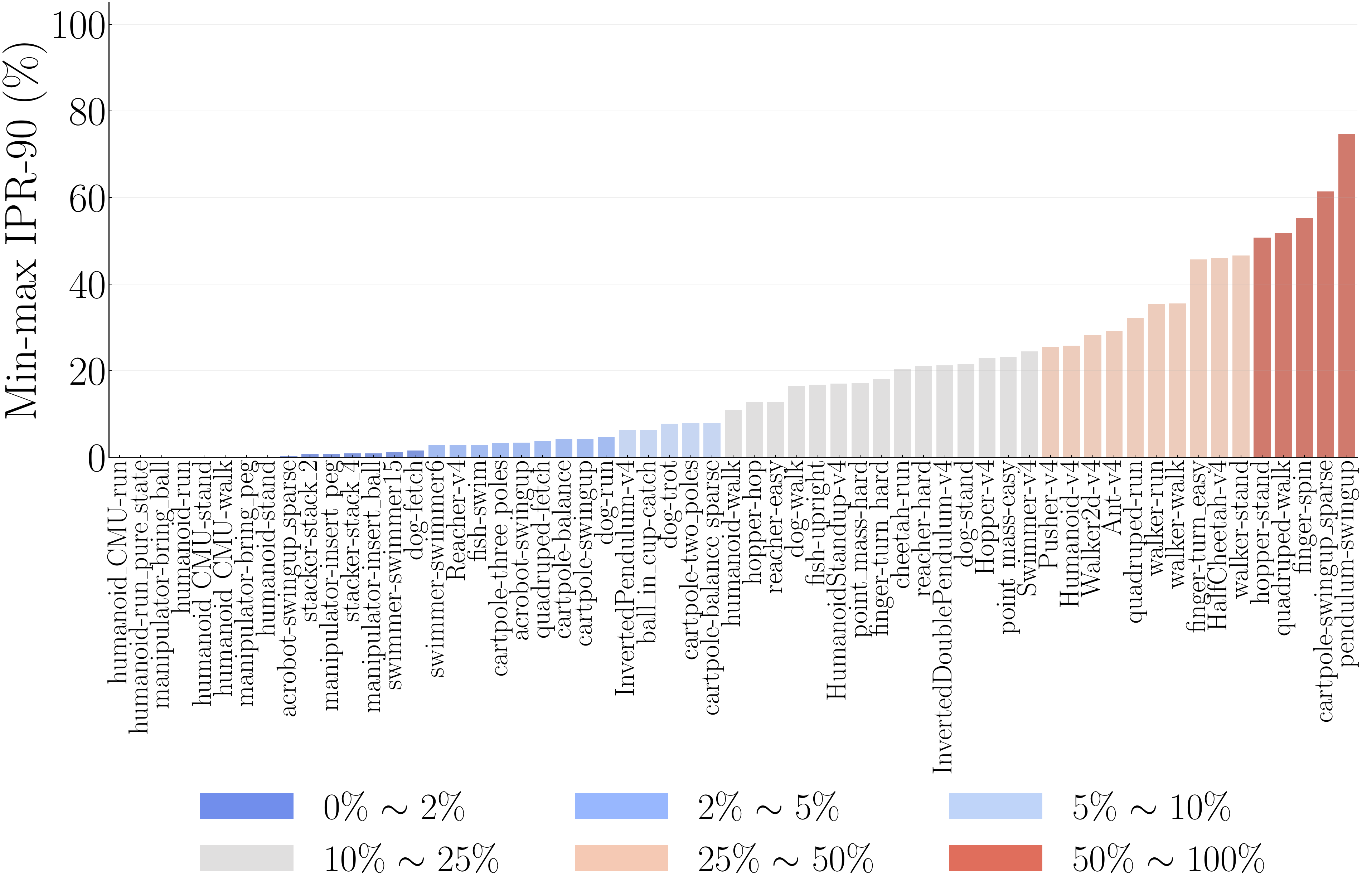}
        \caption{LayerNorm IPR-\(90\)}
        \label{fig:ppo_comp_bar_plots:layernorm:ipr}
    \end{subfigure}
    \hspace{0.05\textwidth}
    \begin{subfigure}[b]{0.4\textwidth}
        \centering
        \includegraphics[width=\textwidth]{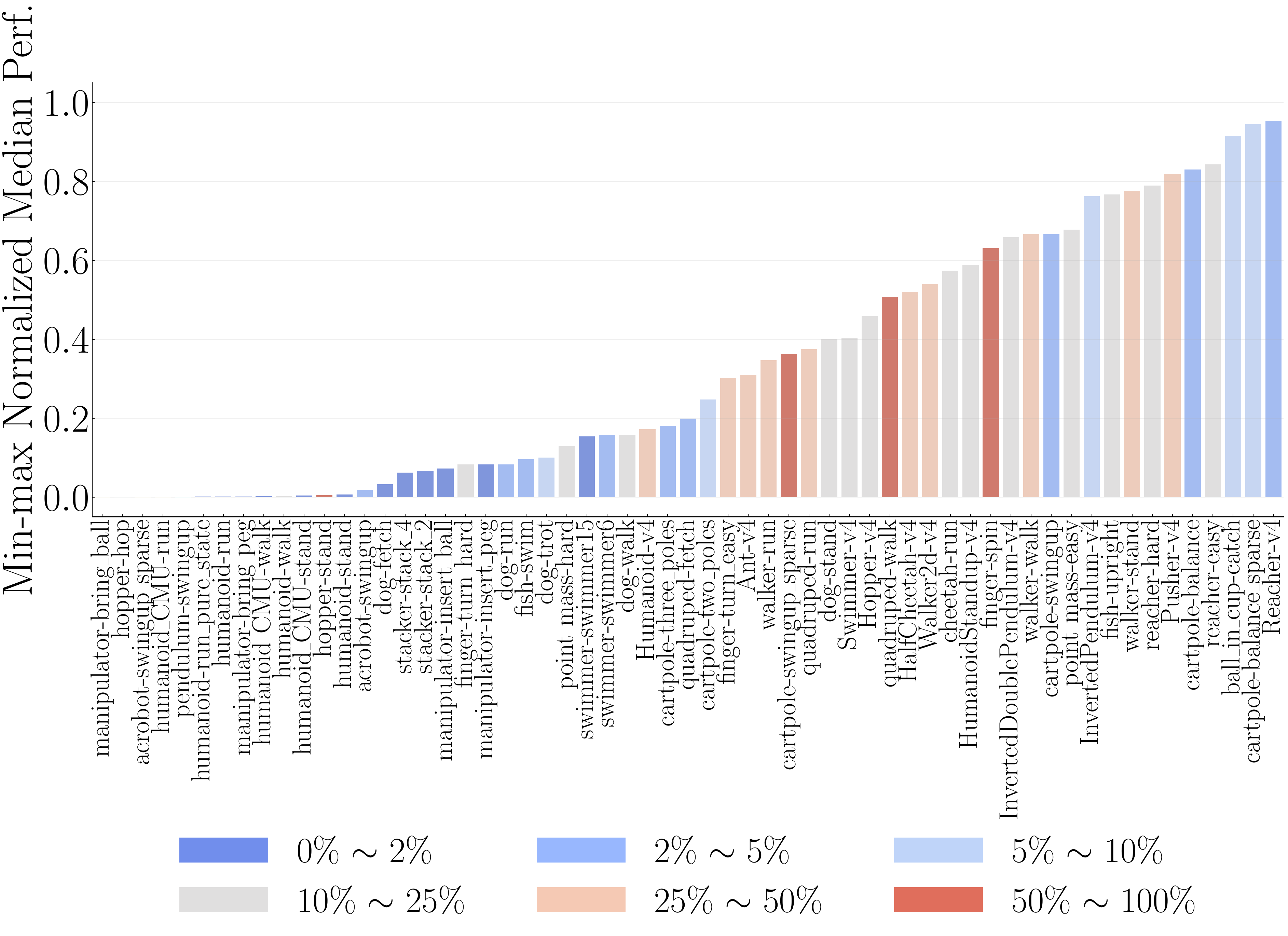}
        \caption{LayerNorm Median}
        \label{fig:ppo_comp_bar_plots:layernorm:med}
    \end{subfigure}\\
    \begin{subfigure}[b]{0.4\textwidth}
        \centering
        \includegraphics[width=\textwidth]{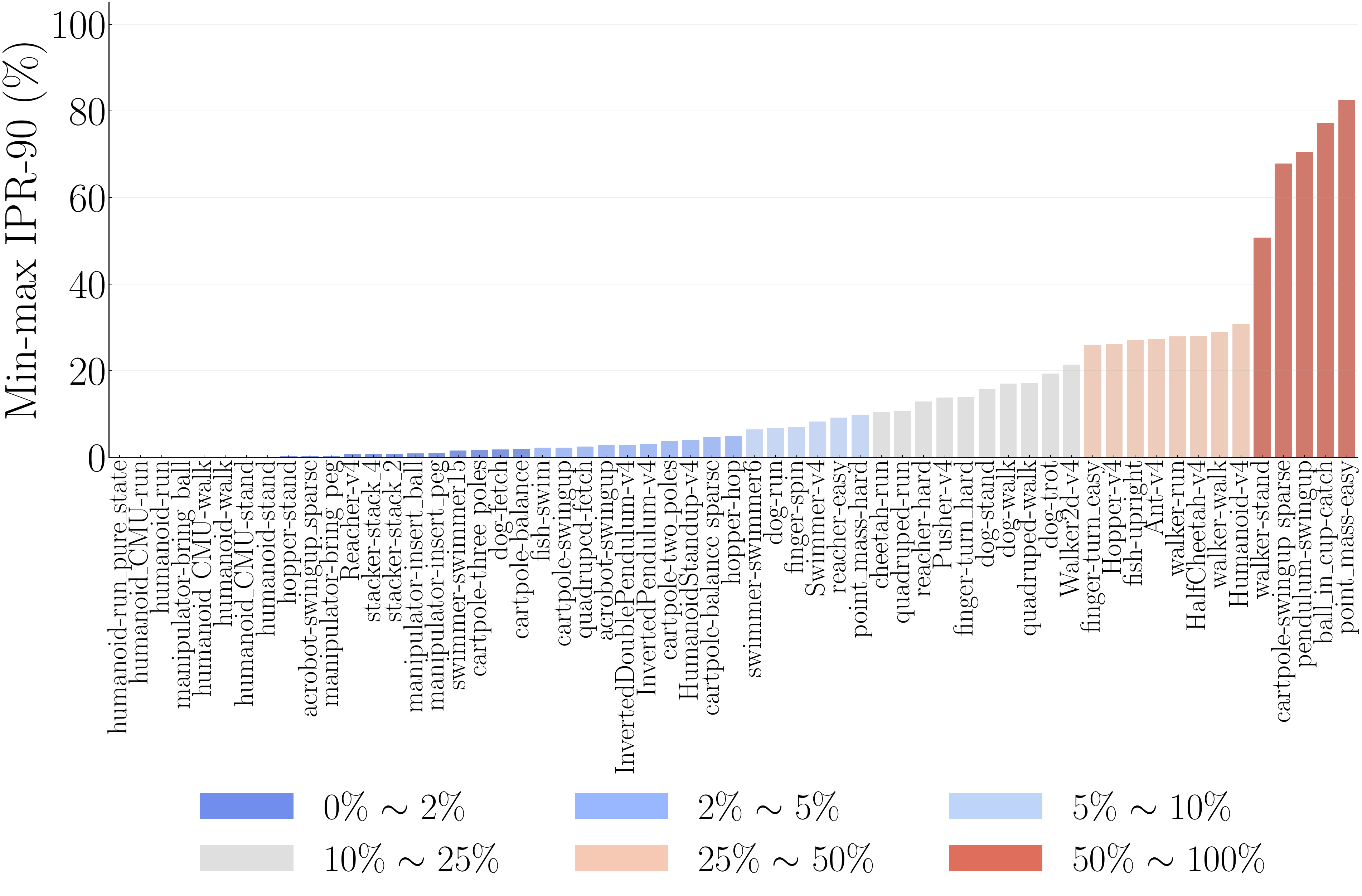}
        \caption{PNorm IPR-\(90\)}
        \label{fig:ppo_comp_bar_plots:pnorm:ipr}
    \end{subfigure}
    \hspace{0.05\textwidth}
    \begin{subfigure}[b]{0.4\textwidth}
        \centering
        \includegraphics[width=\textwidth]{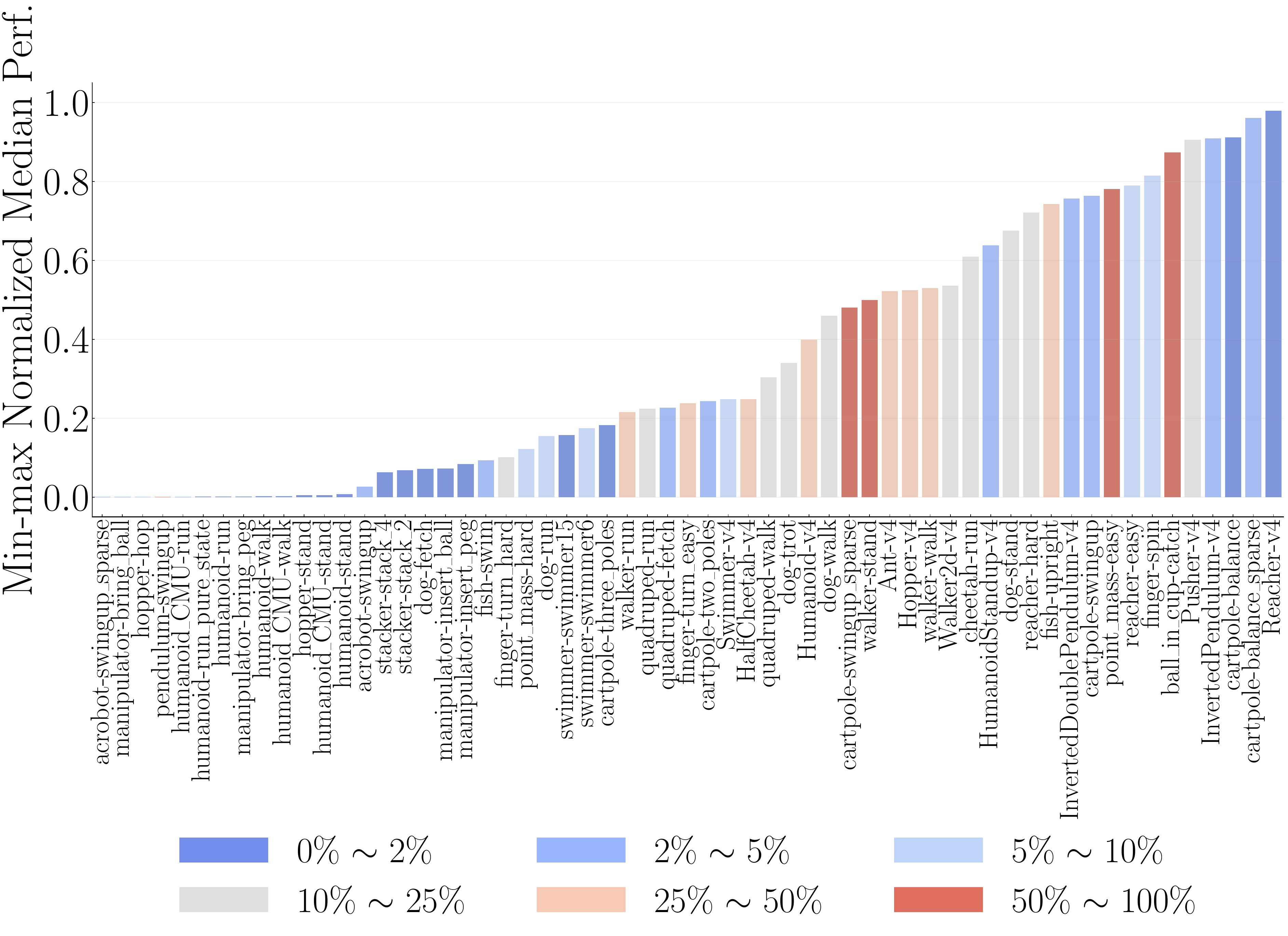}
        \caption{PNorm Median}
        \label{fig:ppo_comp_bar_plots:pnorm:med}
    \end{subfigure}\\
    \begin{subfigure}[b]{0.4\textwidth}
        \centering
        \includegraphics[width=\textwidth]{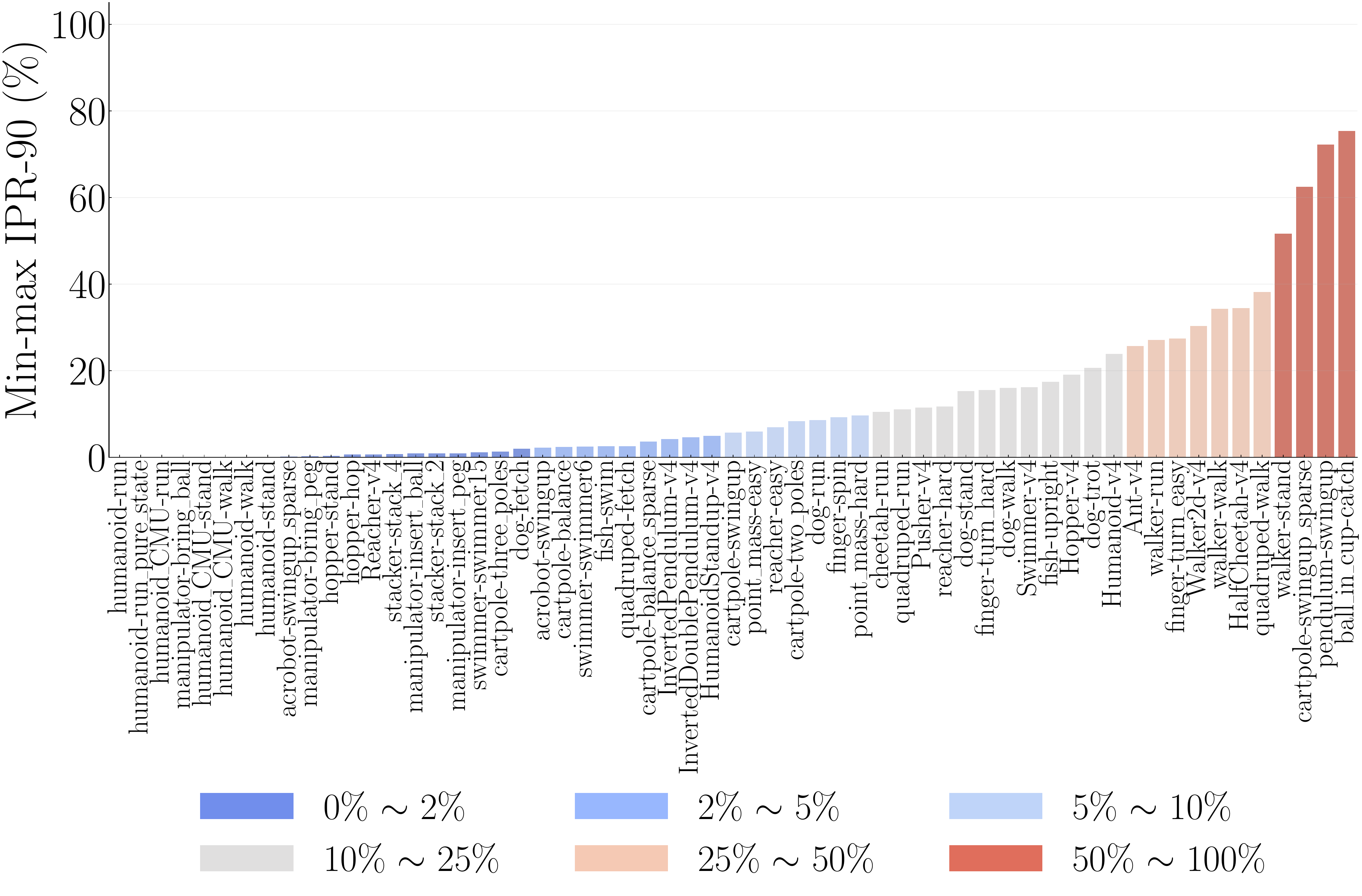}
        \caption{Normalized IPR-\(90\)}
        \label{fig:ppo_comp_bar_plots:lpnorm:ipr}
    \end{subfigure}
    \hspace{0.05\textwidth}
    \begin{subfigure}[b]{0.4\textwidth}
        \centering
        \includegraphics[width=\textwidth]{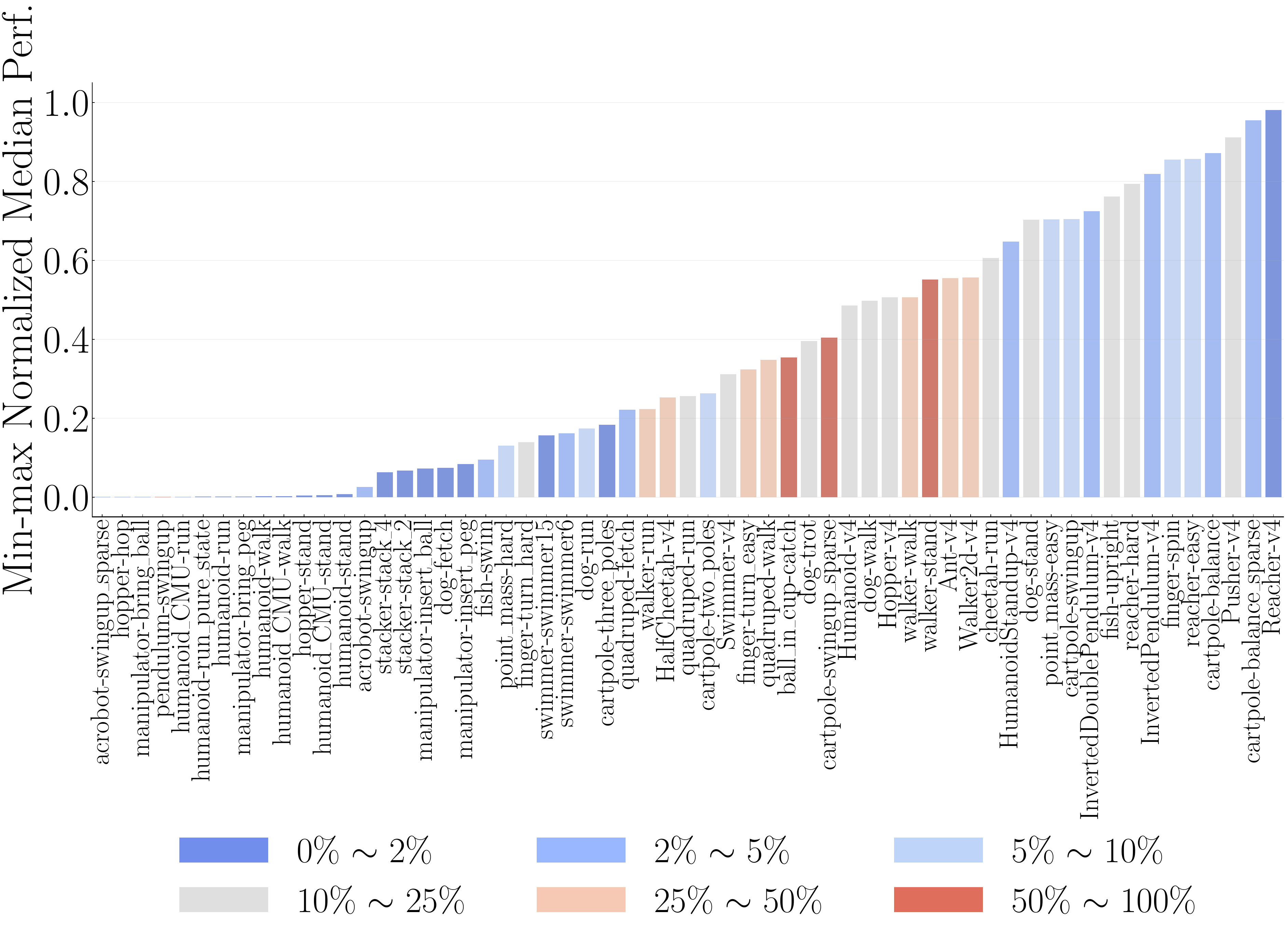}
        \caption{Normalized Median}
        \label{fig:ppo_comp_bar_plots:lpnorm:med}
    \end{subfigure}\\

    \caption
    {
        Comparison of performance variation and median bar plots for PPO with different normalization techniques.
    }
    \label{fig:ppo_comp_bar_plots}
\end{figure*}

\begin{figure*}[h]
    \centering
    \begin{subfigure}[b]{0.4\textwidth}
        \centering
        \includegraphics[width=\textwidth]{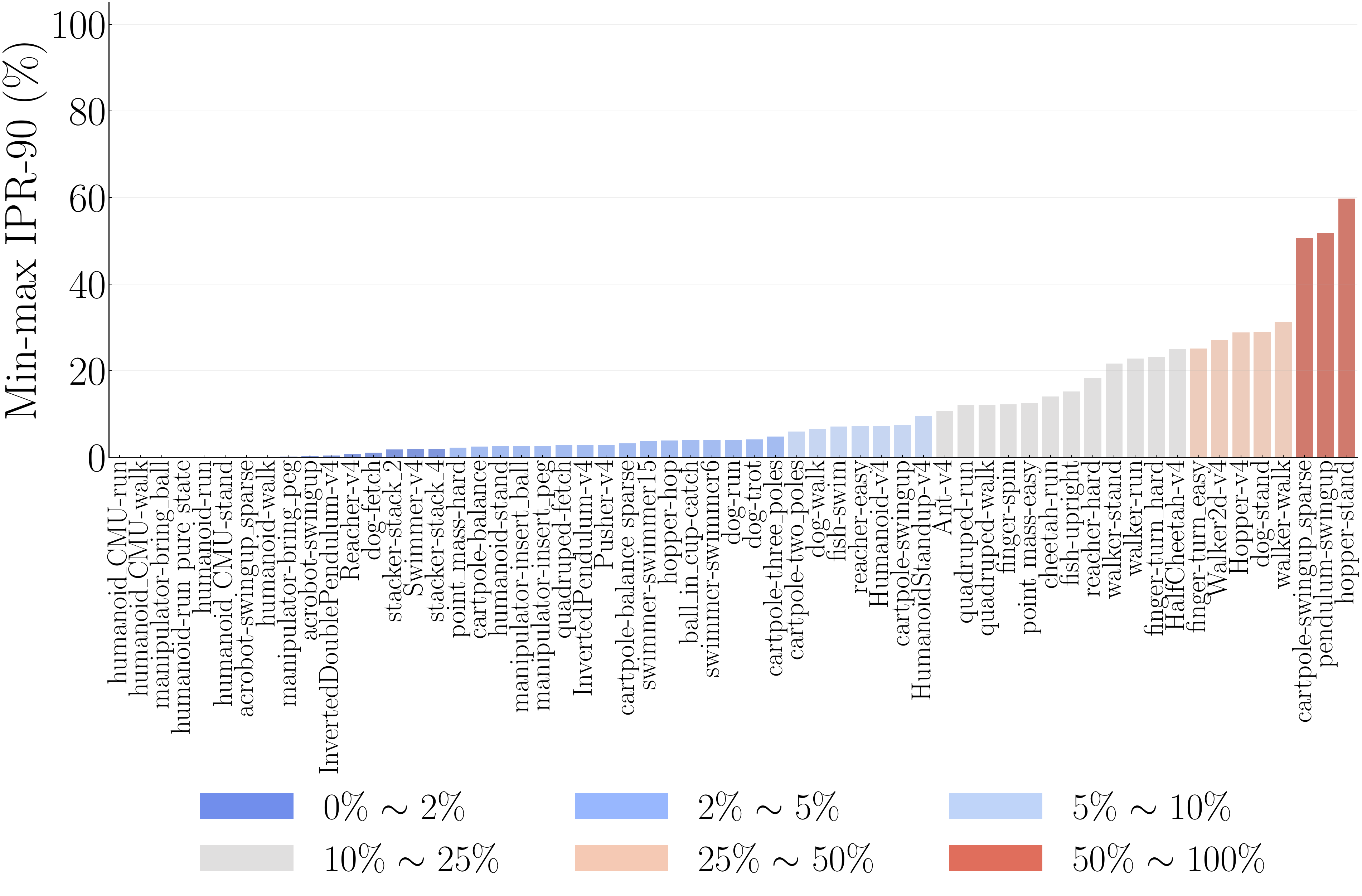}
        \caption{Default IPR-\(90\)}
        \label{fig:sac_comp_bar_plots:default:ipr}
    \end{subfigure}
    \hspace{0.05\textwidth}
    \begin{subfigure}[b]{0.4\textwidth}
        \centering
        \includegraphics[width=\textwidth]{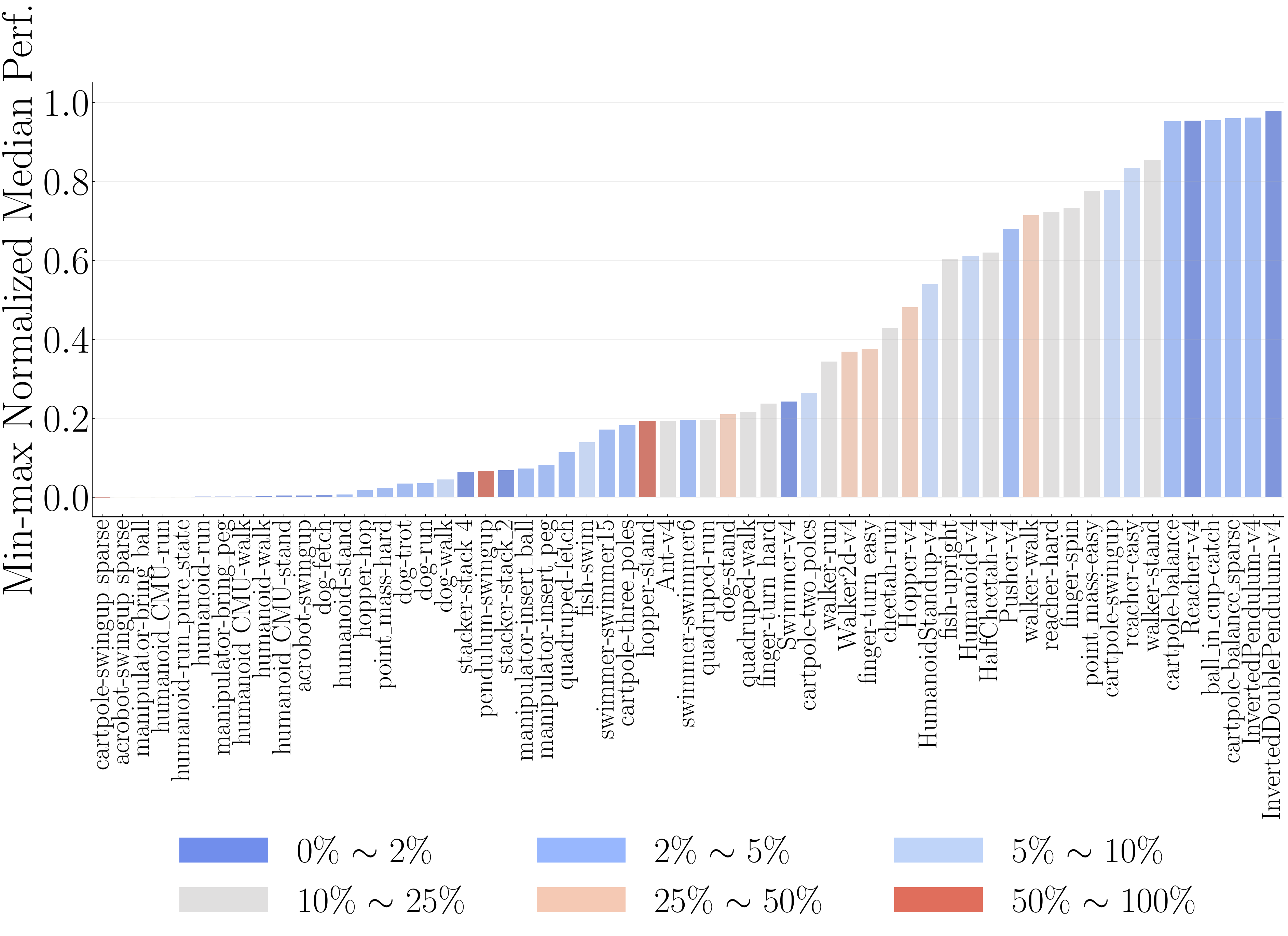}
        \caption{Default Median}
        \label{fig:sac_comp_bar_plots:default:med}
    \end{subfigure}\\
    \begin{subfigure}[b]{0.4\textwidth}
        \centering
        \includegraphics[width=\textwidth]{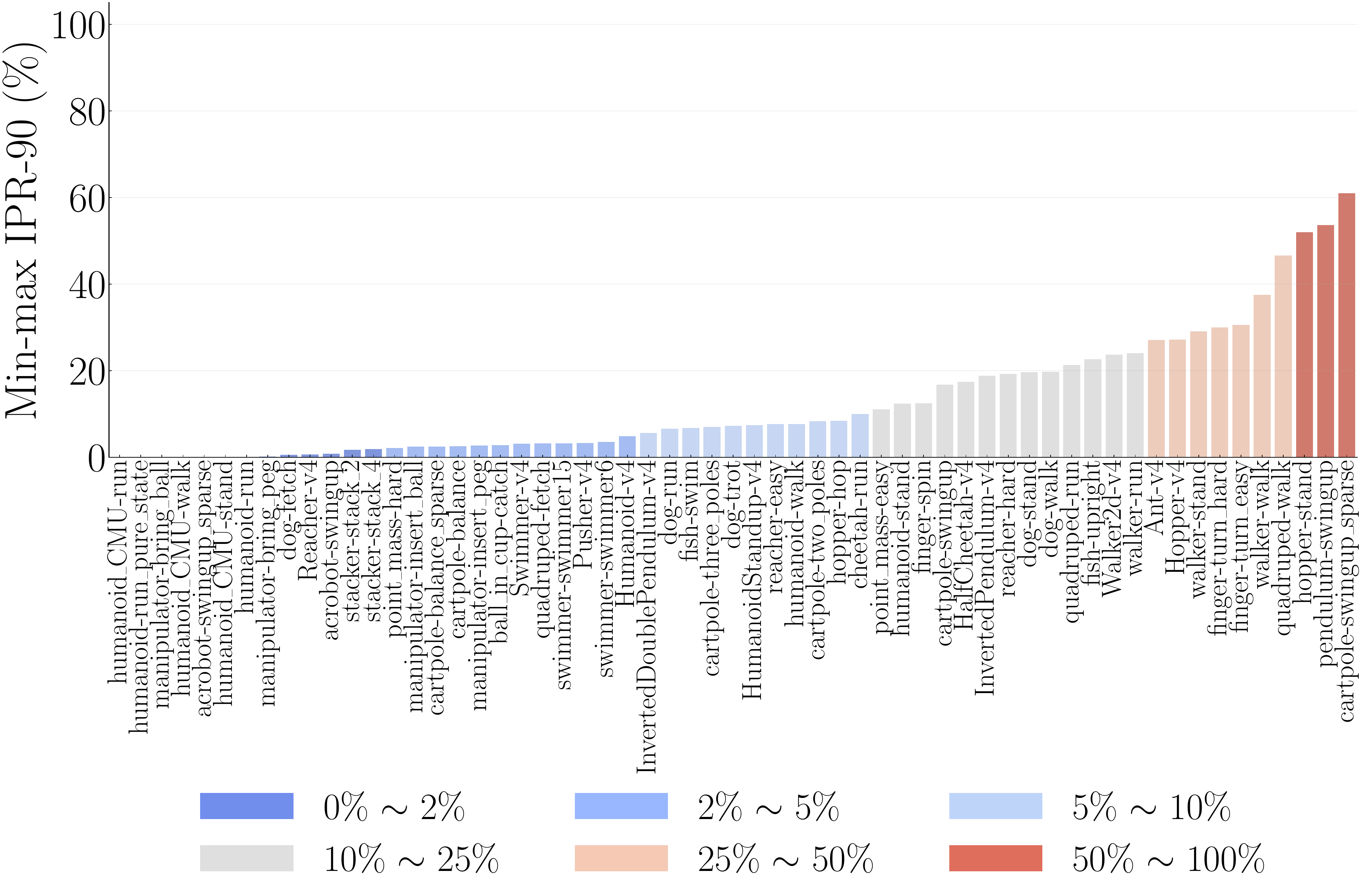}
        \caption{LayerNorm IPR-\(90\)}
        \label{fig:sac_comp_bar_plots:layernorm:ipr}
    \end{subfigure}
    \hspace{0.05\textwidth}
    \begin{subfigure}[b]{0.4\textwidth}
        \centering
        \includegraphics[width=\textwidth]{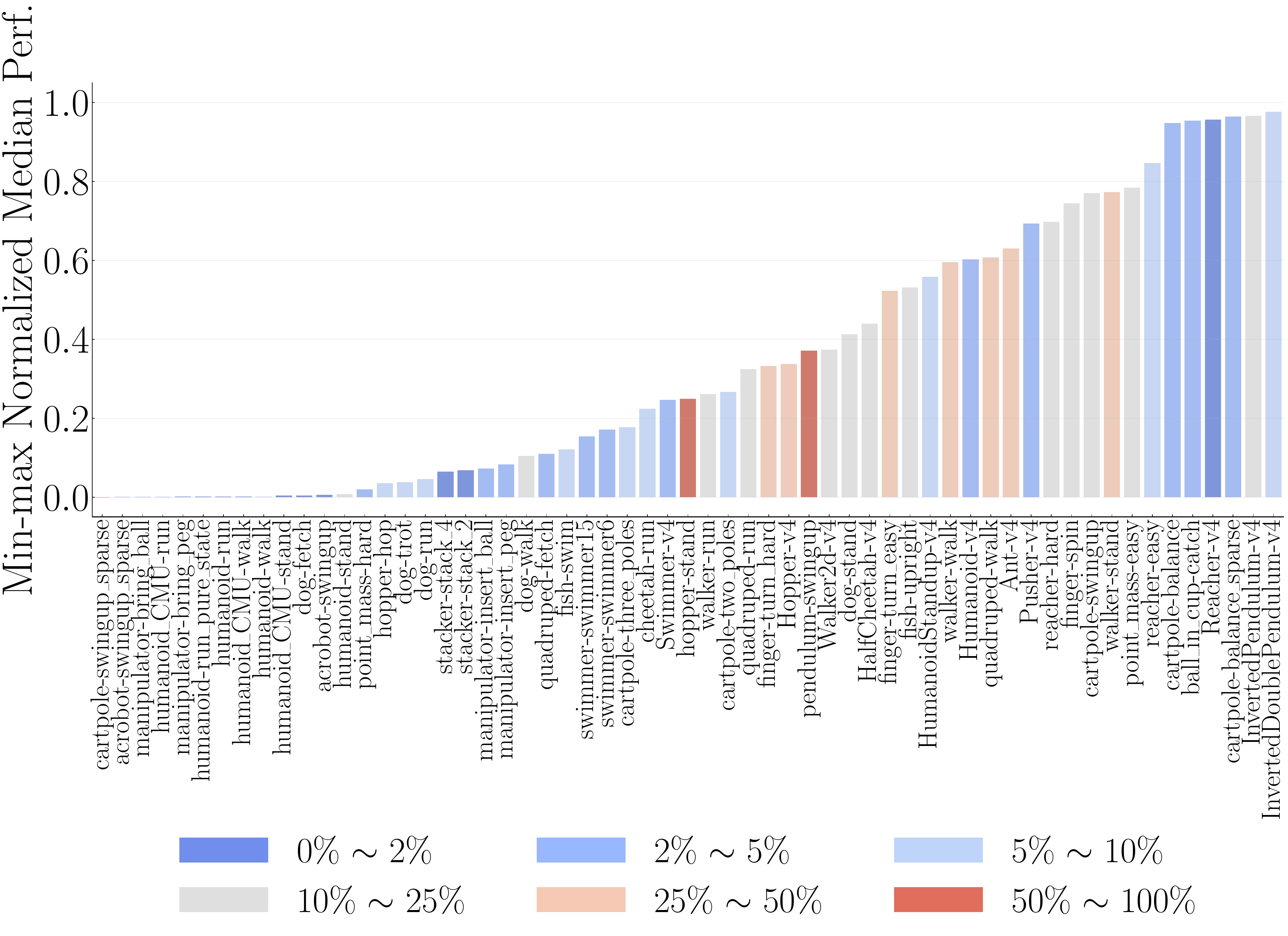}
        \caption{LayerNorm Median}
        \label{fig:sac_comp_bar_plots:layernorm:med}
    \end{subfigure}\\
    \begin{subfigure}[b]{0.4\textwidth}
        \centering
        \includegraphics[width=\textwidth]{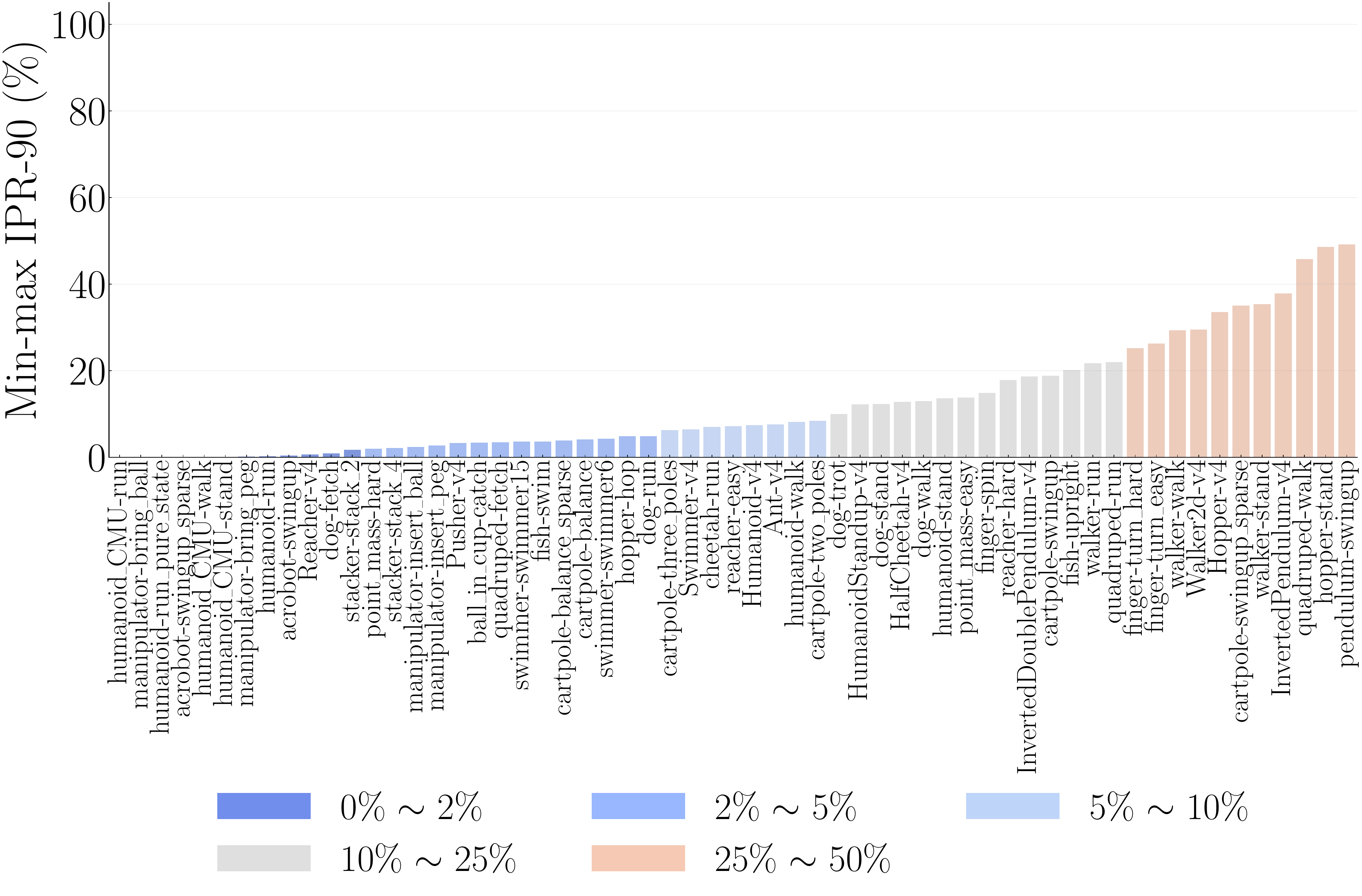}
        \caption{PNorm IPR-\(90\)}
        \label{fig:sac_comp_bar_plots:pnorm:ipr}
    \end{subfigure}
    \hspace{0.05\textwidth}
    \begin{subfigure}[b]{0.4\textwidth}
        \centering
        \includegraphics[width=\textwidth]{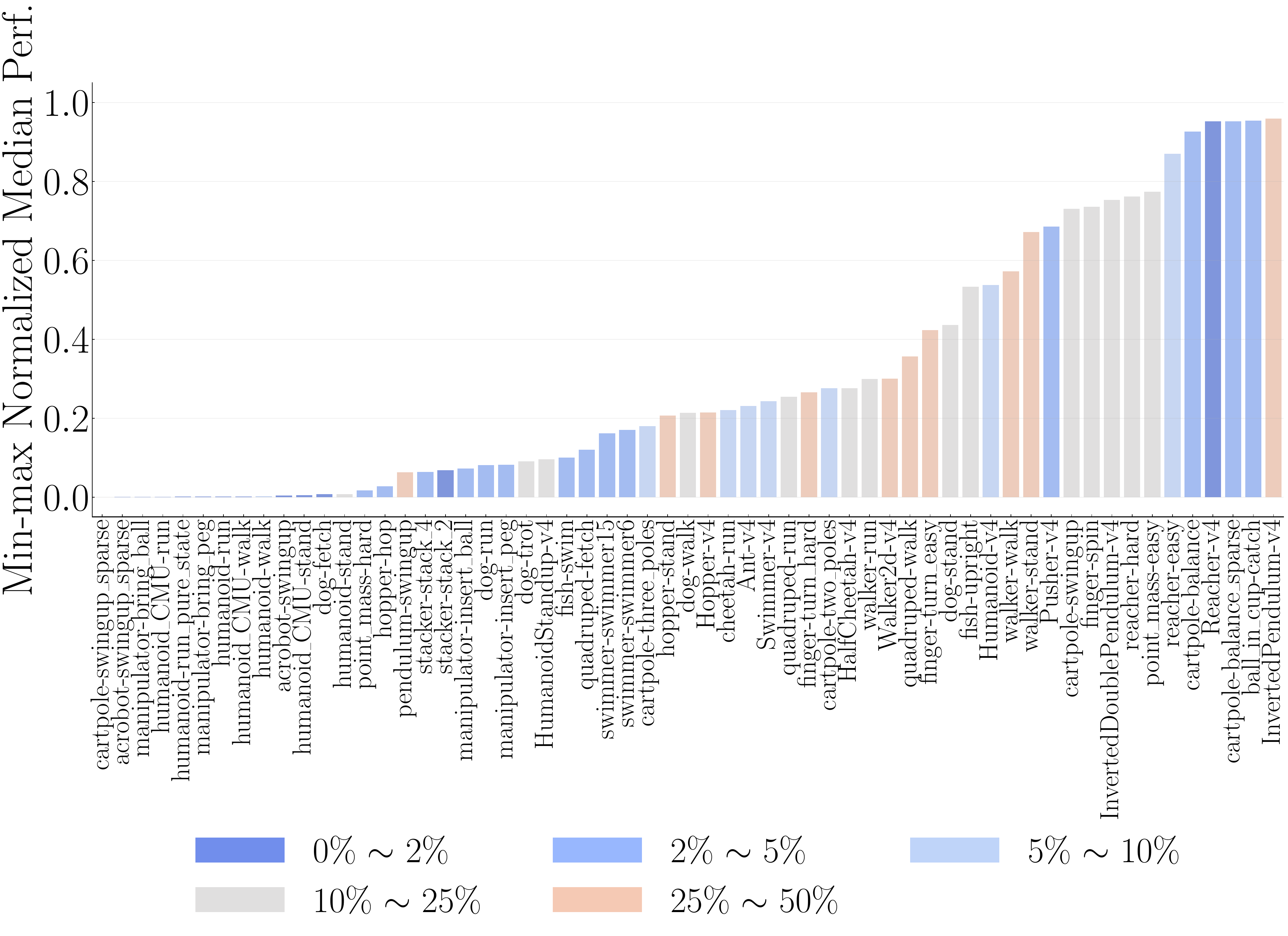}
        \caption{PNorm Median}
        \label{fig:sac_comp_bar_plots:pnorm:med}
    \end{subfigure}\\
    \begin{subfigure}[b]{0.4\textwidth}
        \centering
        \includegraphics[width=\textwidth]{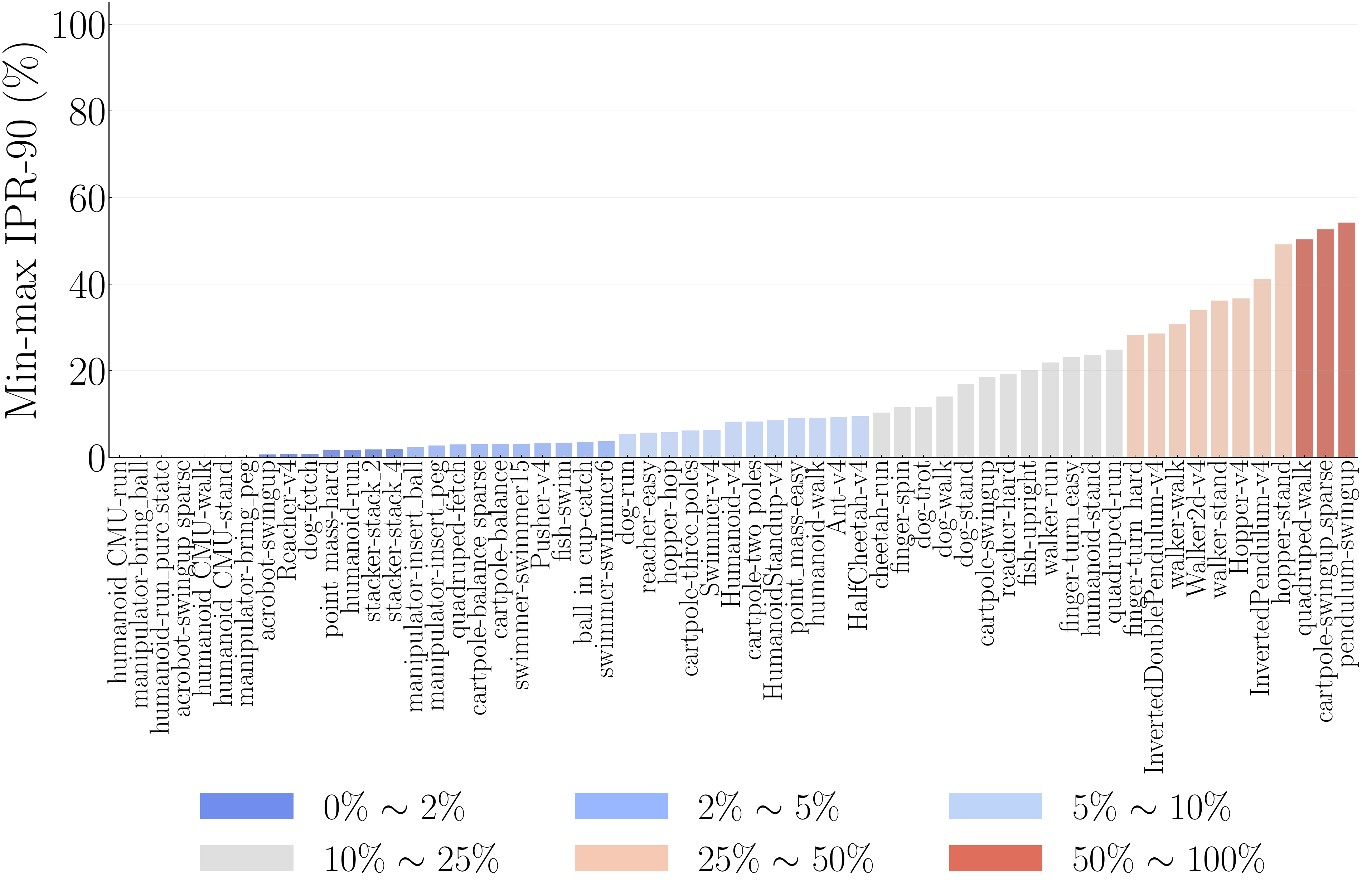}
        \caption{Normalized IPR-\(90\)}
        \label{fig:sac_comp_bar_plots:lpnorm:ipr}
    \end{subfigure}
    \hspace{0.05\textwidth}
    \begin{subfigure}[b]{0.4\textwidth}
        \centering
        \includegraphics[width=\textwidth]{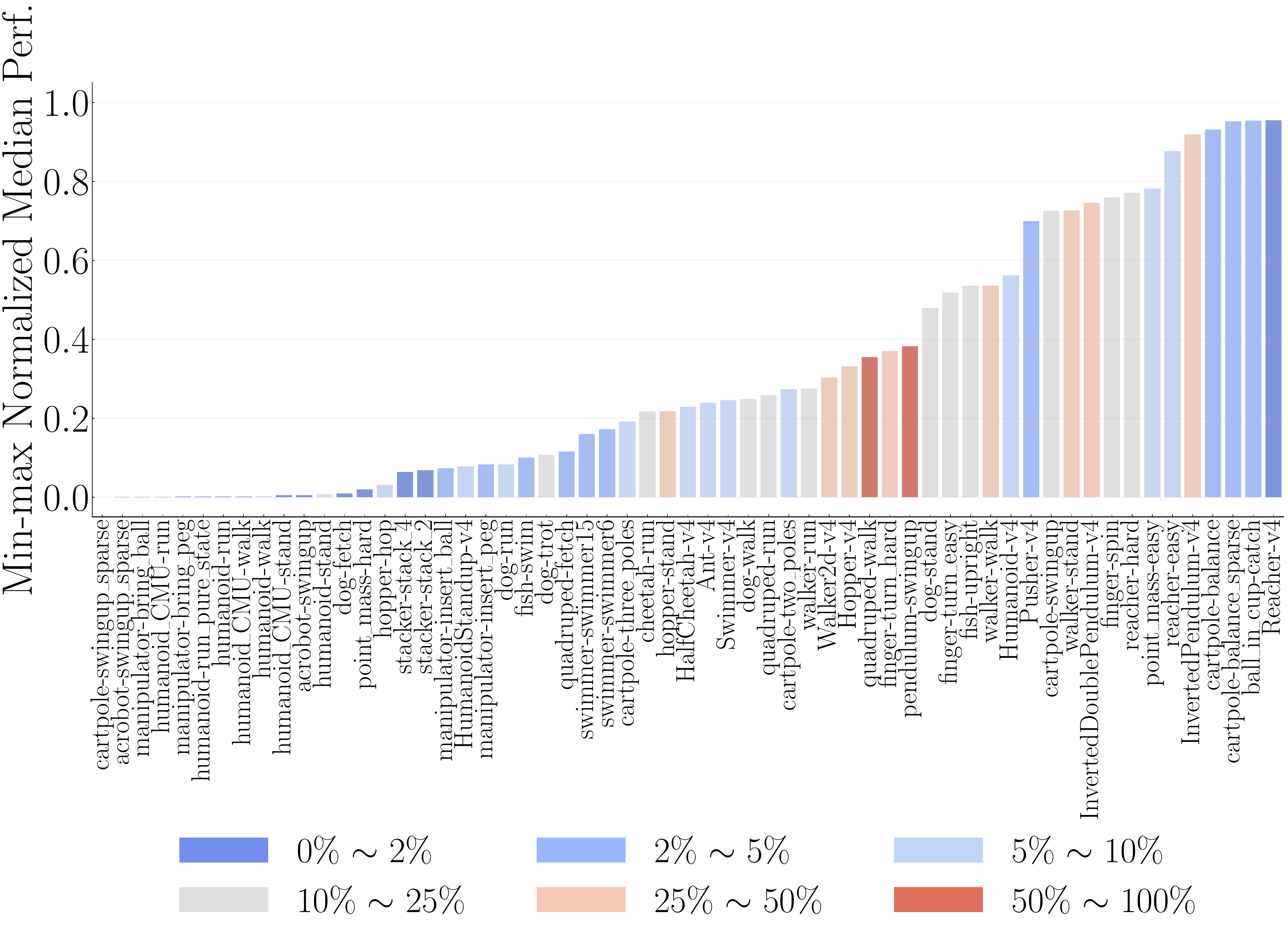}
        \caption{Normalized Median}
        \label{fig:sac_comp_bar_plots:lpnorm:med}
    \end{subfigure}\\

    \caption
    {
        Comparison of performance variation and median bar plots for SAC with different normalization techniques.
    }
    \label{fig:sac_comp_bar_plots}
\end{figure*}

\clearpage
\section{Comparison Plots of Learning Curves Before and After Applying Normalization Techniques}
\label{sec:appendix:use_case_comp}

\begin{figure*}[hbt]
    \centering
    \begin{subfigure}[b]{0.46\textwidth}
        \centering
        \includegraphics[width=\textwidth]{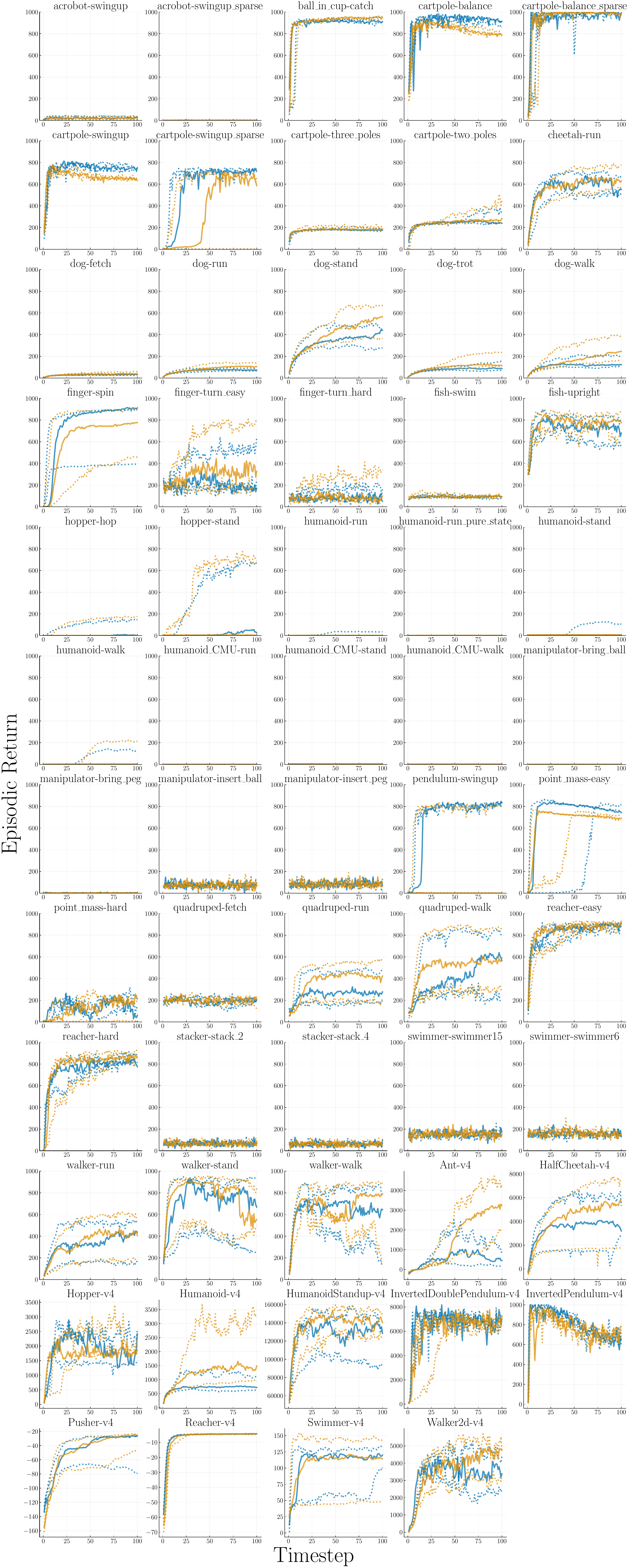}
        \caption{PPO}
        \label{fig:baseline_vs_layernorm:ppo}
    \end{subfigure}
    \hfill
    \begin{subfigure}[b]{0.46\textwidth}
        \centering
        \includegraphics[width=\textwidth]{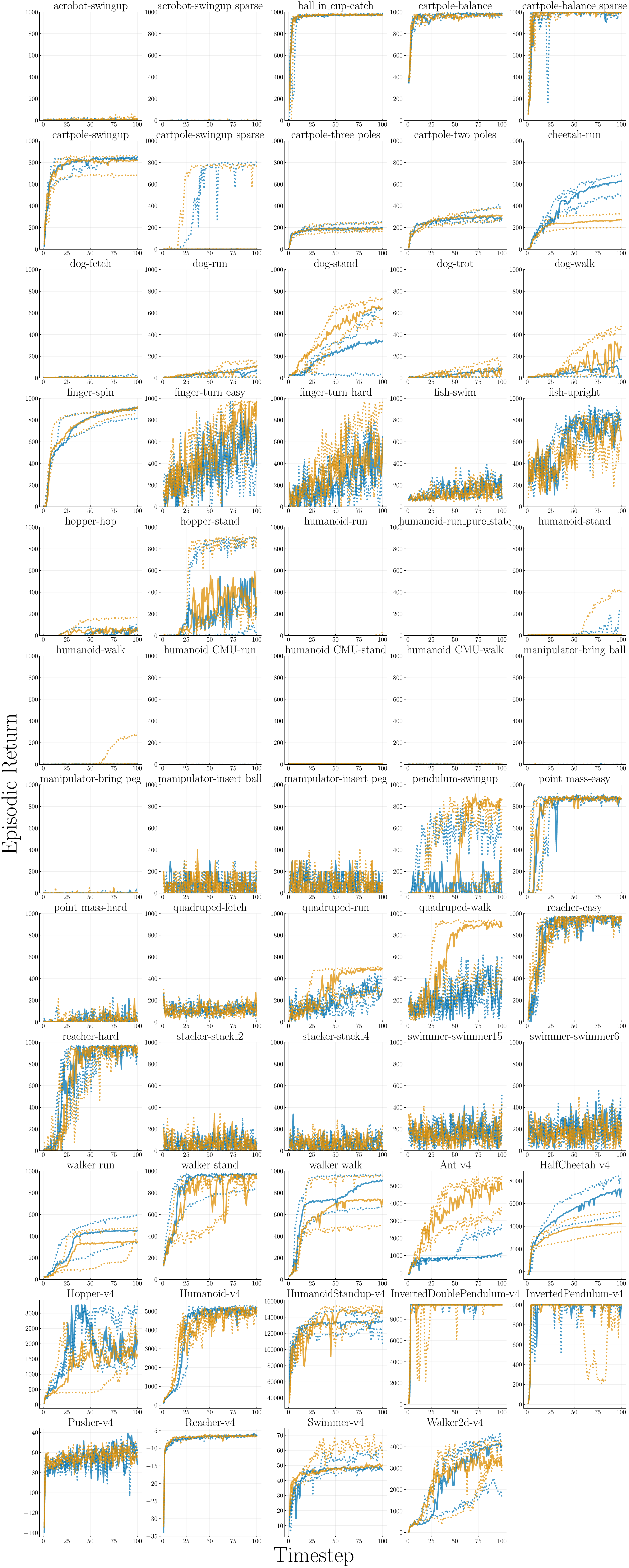}
        \caption{SAC}
        \label{fig:baseline_vs_layernorm:sac}
    \end{subfigure}

    \caption
    {
        Visual comparison of performance variation between PPO/SAC and LayerNorm PPO/SAC.
        Blue and orange curves represent the baseline and LayerNorm variants of each algorithm, respectively.
        Learning curves are visualized by RPH.
    }
    \label{fig:baseline_vs_layernorm}
\end{figure*}

\begin{figure*}[hbt]
    \centering
    \begin{subfigure}[b]{0.48\textwidth}
        \centering
        \includegraphics[width=\textwidth]{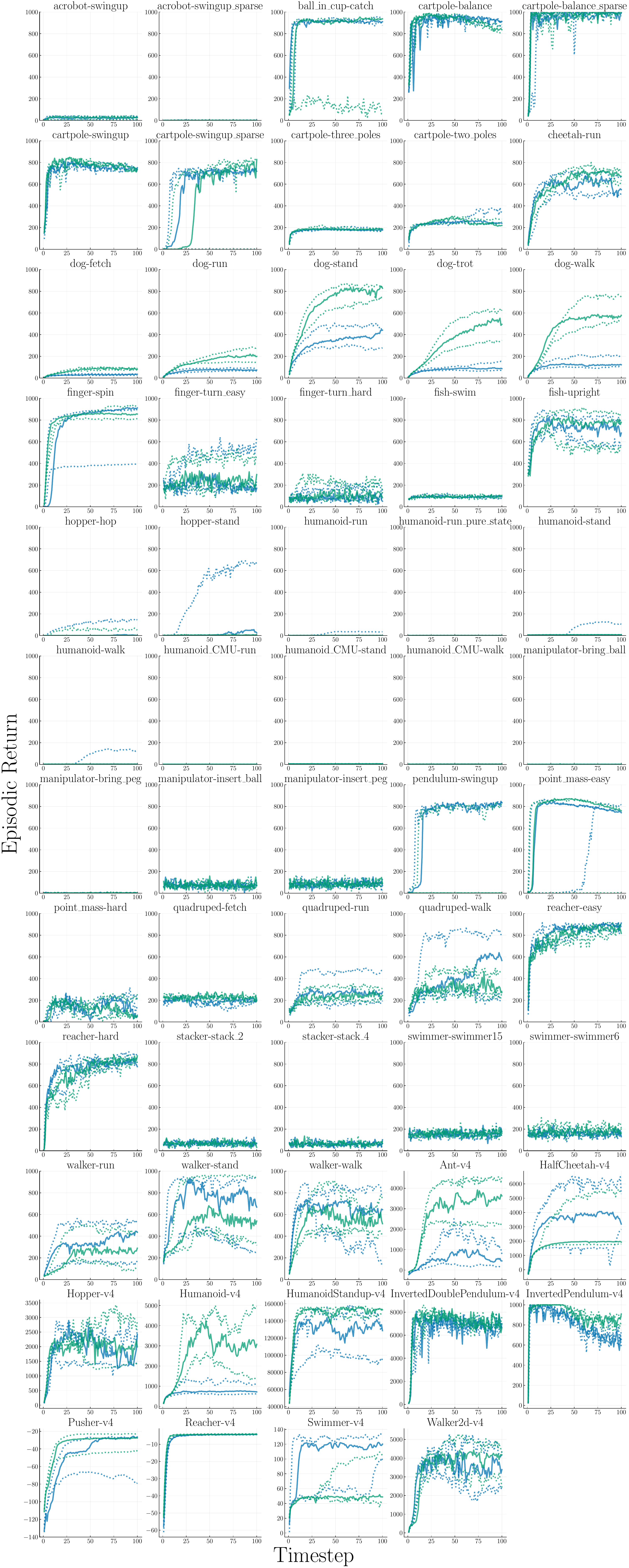}
        \caption{PPO}
        \label{fig:baseline_vs_pnorm:ppo}
    \end{subfigure}
    \hfill
    \begin{subfigure}[b]{0.48\textwidth}
        \centering
        \includegraphics[width=\textwidth]{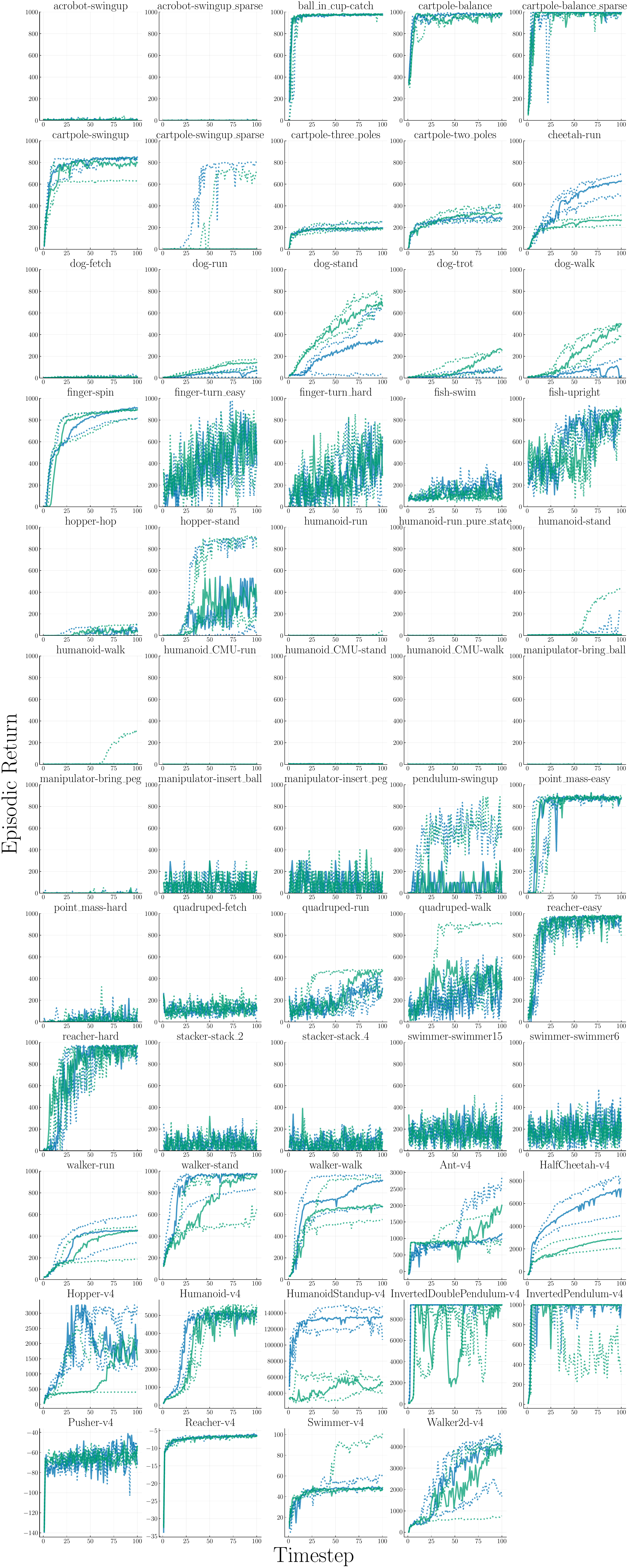}
        \caption{SAC}
        \label{fig:baseline_vs_pnorm:sac}
    \end{subfigure}

    \caption
    {
        Visual comparison of performance variation between PPO/SAC and PNorm PPO/SAC.
        Blue and green curves represent the baseline and PNorm variants of each algorithm, respectively.
        Learning curves are visualized by RPH.
    }
    \label{fig:baseline_vs_pnorm}
\end{figure*}

\begin{figure*}[hbt]
    \centering
    \begin{subfigure}[b]{0.48\textwidth}
        \centering
        \includegraphics[width=\textwidth]{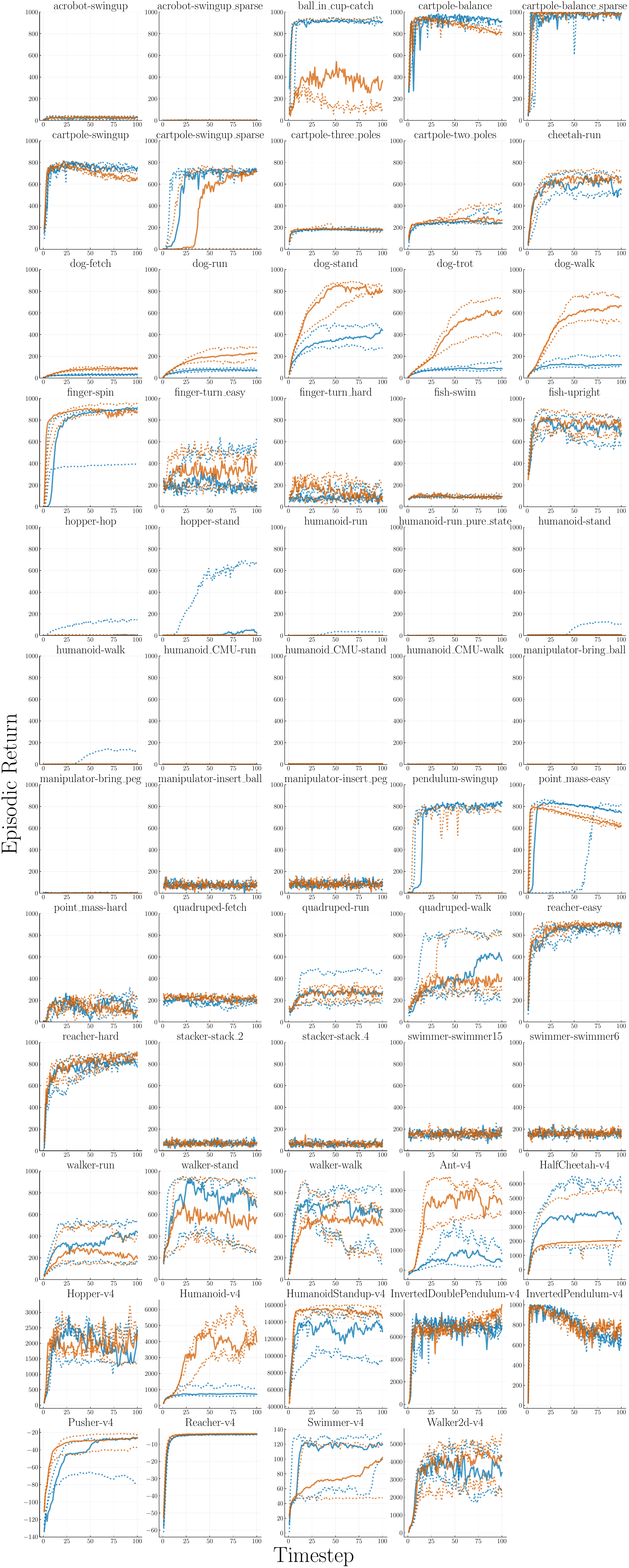}
        \caption{PPO}
        \label{fig:baseline_vs_lpnorm:ppo}
    \end{subfigure}
    \hfill
    \begin{subfigure}[b]{0.48\textwidth}
        \centering
        \includegraphics[width=\textwidth]{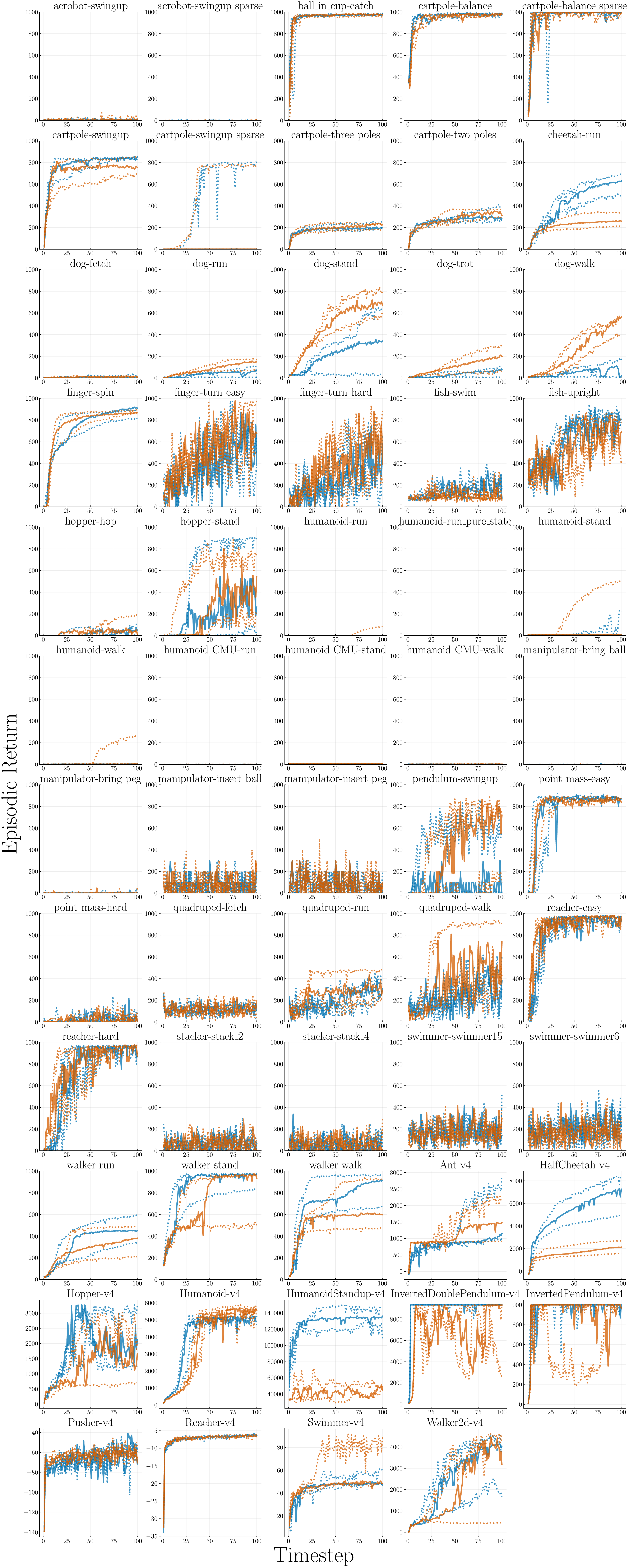}
        \caption{SAC}
        \label{fig:baseline_vs_lpnorm:sac}
    \end{subfigure}

    \caption
    {
        Visual comparison of performance variation between PPO/SAC and Normalized PPO/SAC.
        Blue and red curves represent the baseline and Normalized variants of each algorithm, respectively.
        Learning curves are visualized by RPH.
    }
    \label{fig:baseline_vs_lpnorm}
\end{figure*}

\clearpage
\section{Performance Variation Bar Plots with Different IPR Ranges}
\label{sec:appendix:multi_level_barplots}

\begin{figure*}[hbt]
    \centering
    \begin{subfigure}[b]{0.48\textwidth}
        \centering
        \includegraphics[width=\textwidth]{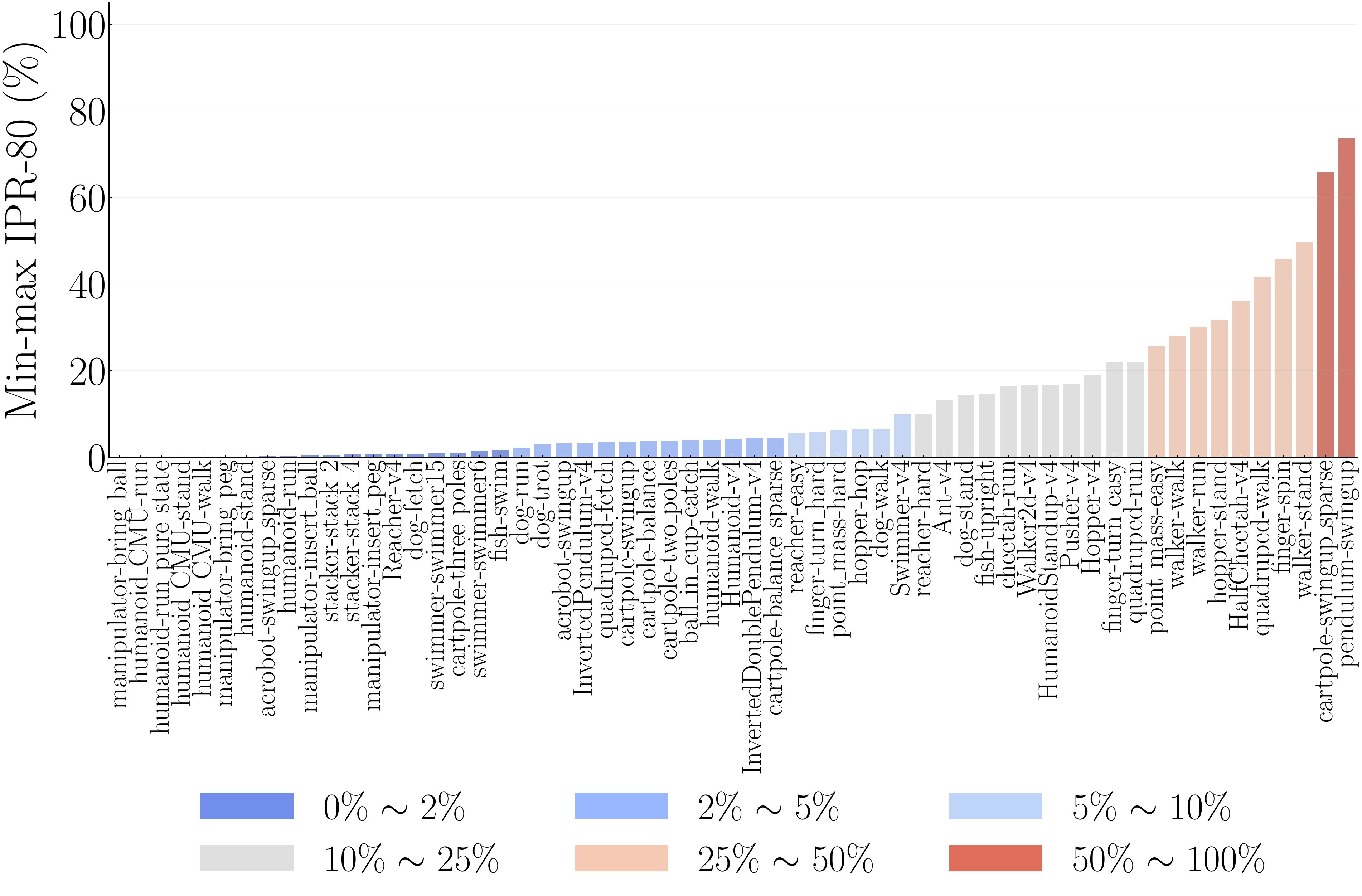}
        \caption{PPO IPR-\(80\)}
        \label{fig:multi_level_comp:ppo:80}
    \end{subfigure}
    \hfill
    \begin{subfigure}[b]{0.48\textwidth}
        \centering
        \includegraphics[width=\textwidth]{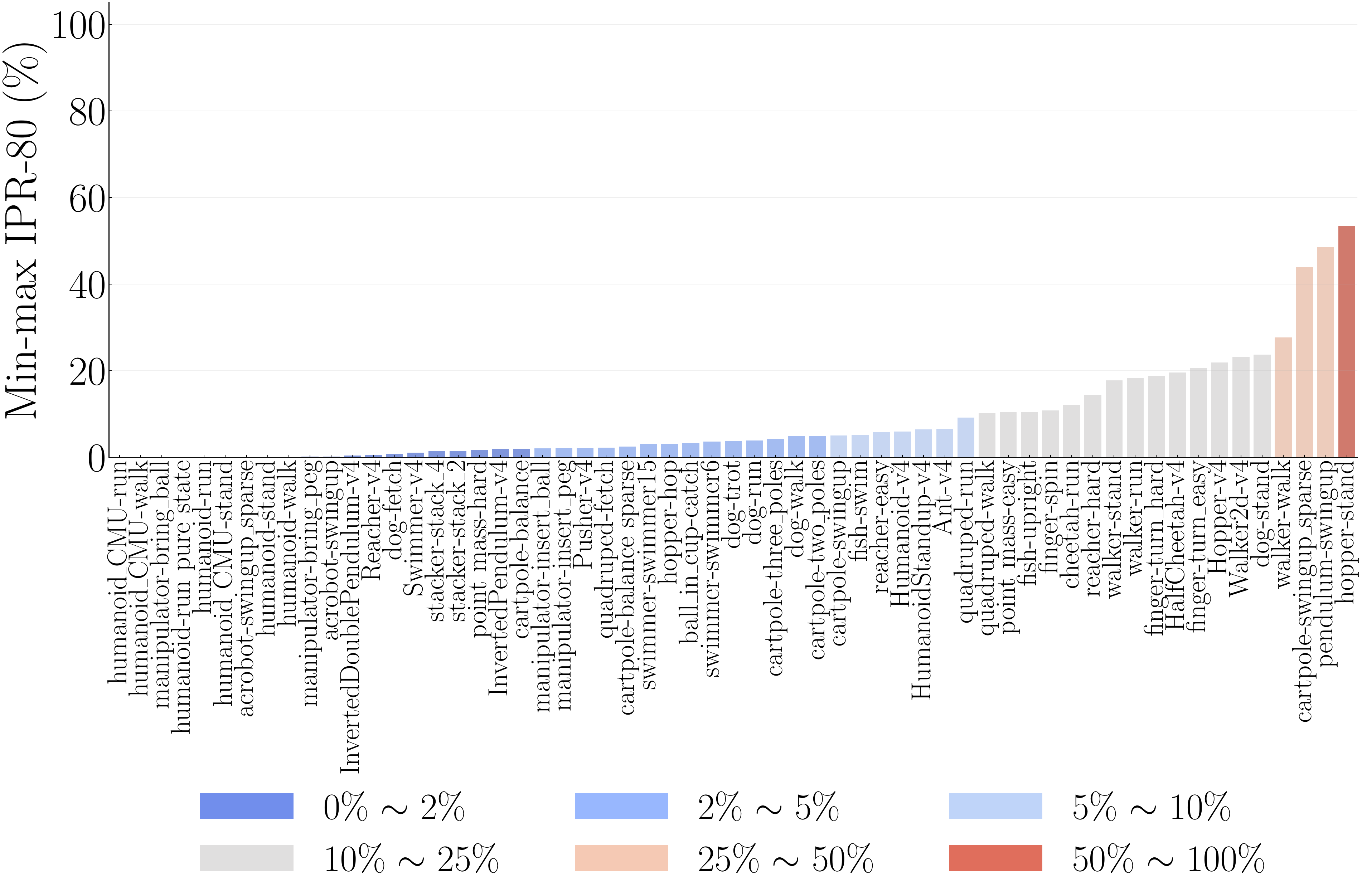}
        \caption{SAC IPR-\(80\)}
        \label{fig:multi_level_comp:sac:80}
    \end{subfigure}\\
    \begin{subfigure}[b]{0.48\textwidth}
        \centering
        \includegraphics[width=\textwidth]{figures/perf_var_bar_ppo_default_compact.pdf}
        \caption{PPO IPR-\(90\)}
        \label{fig:multi_level_comp:ppo:90}
    \end{subfigure}
    \hfill
    \begin{subfigure}[b]{0.48\textwidth}
        \centering
        \includegraphics[width=\textwidth]{figures/perf_var_bar_sac_default_compact.pdf}
        \caption{SAC IPR-\(90\)}
        \label{fig:multi_level_comp:sac:90}
    \end{subfigure}\\
    \begin{subfigure}[b]{0.48\textwidth}
        \centering
        \includegraphics[width=\textwidth]{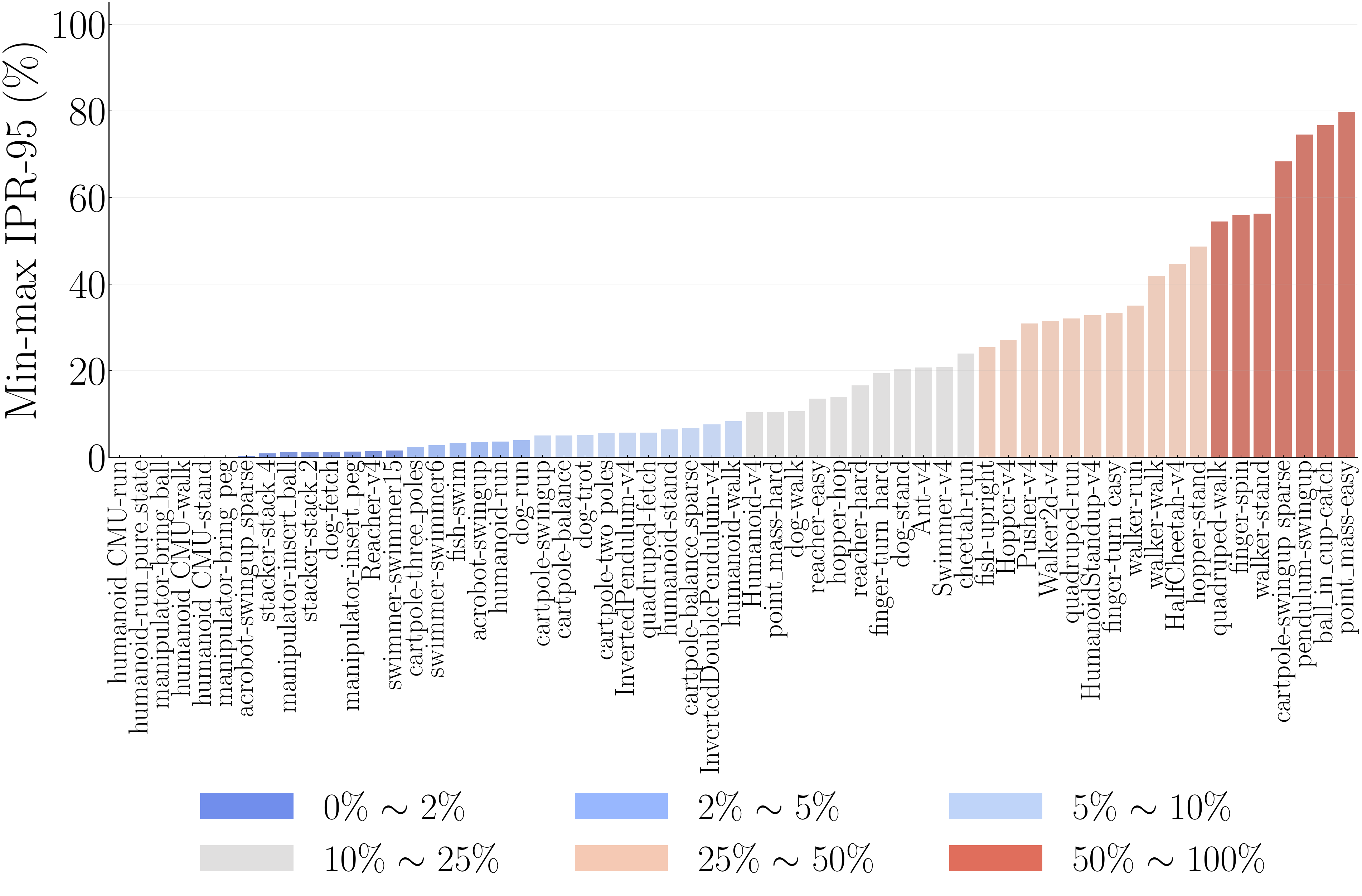}
        \caption{PPO IPR-\(95\)}
        \label{fig:multi_level_comp:ppo:95}
    \end{subfigure}
    \hfill
    \begin{subfigure}[b]{0.48\textwidth}
        \centering
        \includegraphics[width=\textwidth]{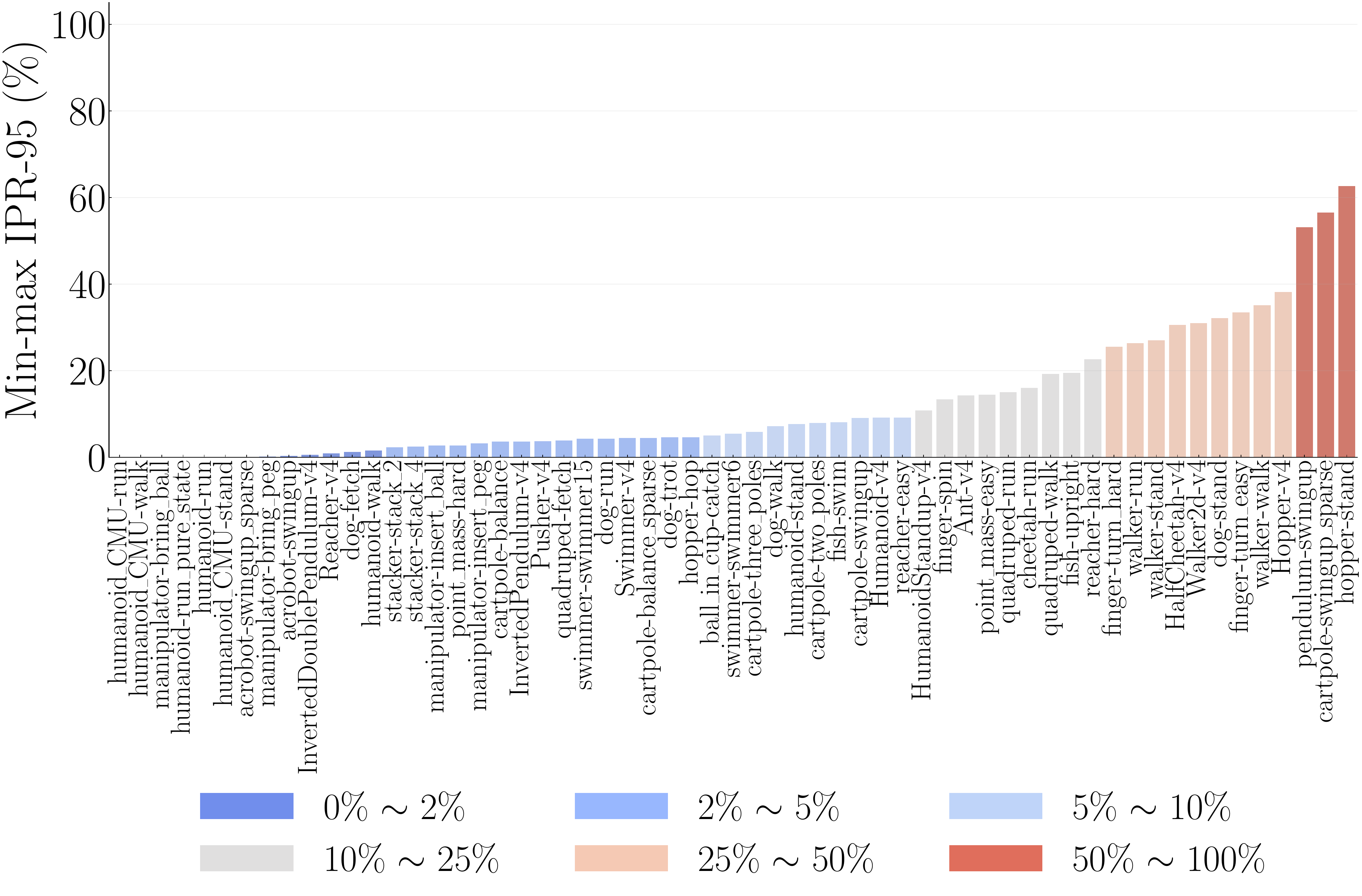}
        \caption{SAC IPR-\(95\)}
        \label{fig:multi_level_comp:sac:95}
    \end{subfigure}\\

    \caption
    {
        Comparison of IPR with a different range of central coverage.
        Different rates of central coverage result in different sensitivity against the tail behavior of performance distributions.
    }
    \label{fig:multi_level_comp}
\end{figure*}

\newpage
\section{Comparison of Learning Curves in Different Plotting Styles}
\label{sec:appendix:lc_comp_std}

\begin{figure*}[tbh]
    \centering
    \includegraphics[width=0.75\textwidth]{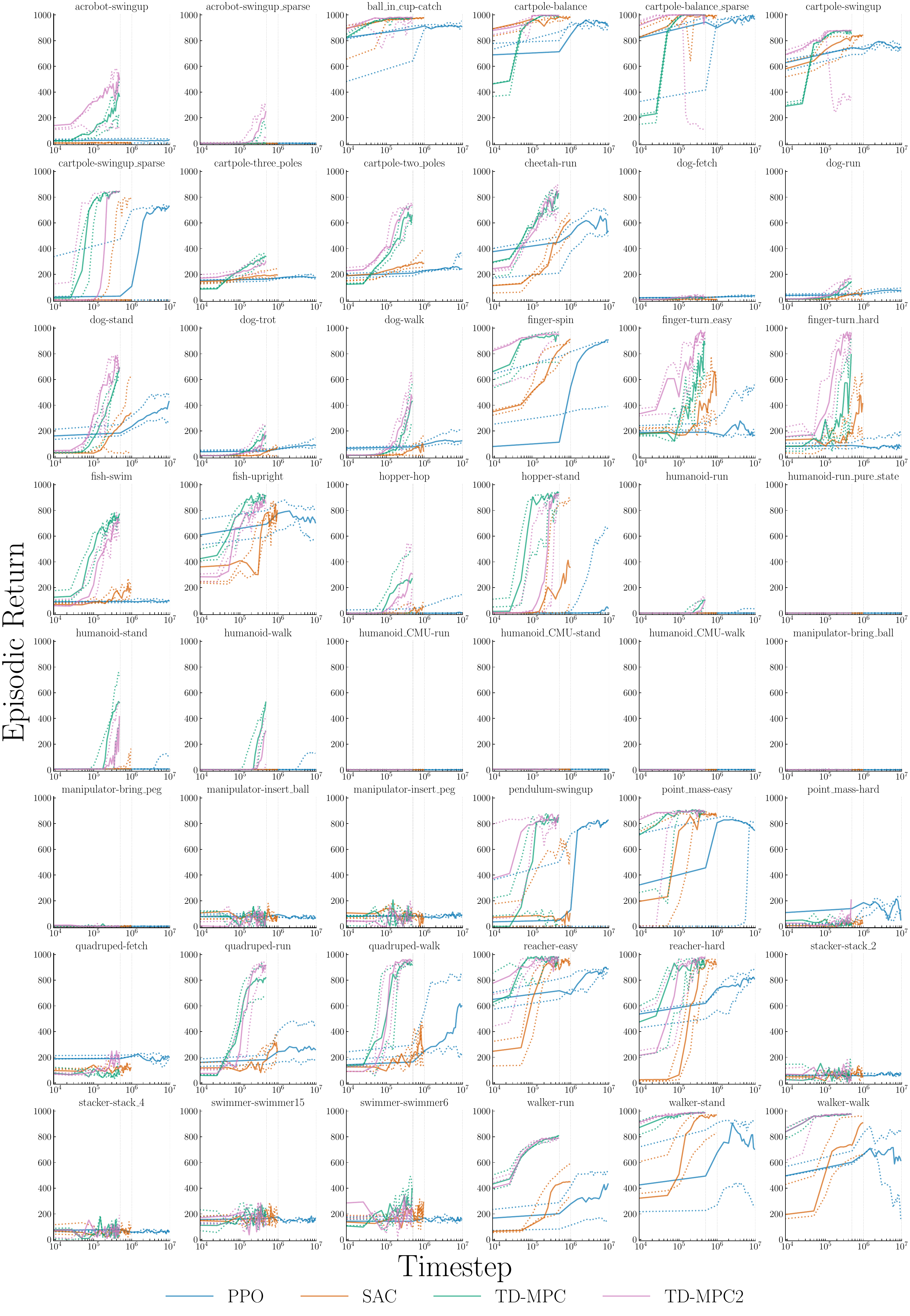}
    \caption{
        Comparison of learning curves of continuous control algorithms.
        Each subfigure shows the RPH learning curves of PPO, SAC, TD-MPC, and TD-MPC2 on a given task.
        For visual clarity, the learning curves are shown with RPH without non-highlighted curves, the number of bins is reduced to \(20\), and x-axis is log-scaled.
        TD-MPC/TD-MPC2 learns more rapidly than PPO/SAC in many tasks, while exhibiting less performance variation.
    }
    \label{fig:learning_curve_comparison_rph}
\end{figure*}

\begin{figure*}[tbh]
    \centering
    \includegraphics[width=0.75\textwidth]{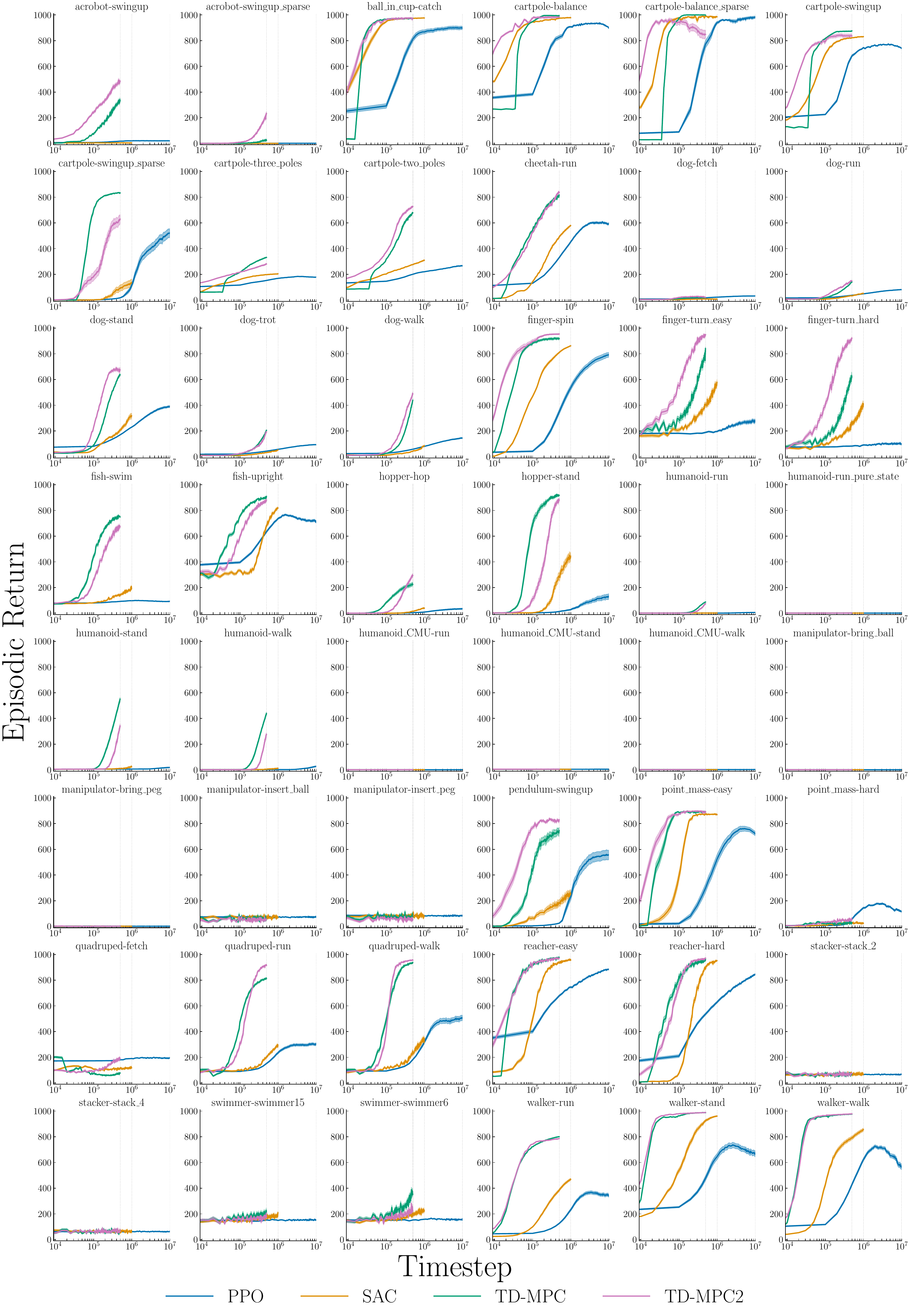}
    \caption{
        Learning curve comparison plot with standard error.
        Each subfigure shows the mean and standard error of the learning curves of PPO, SAC, TD-MPC, and TD-MPC2 on a given task.
        For visual clarity, x-axis is log-scaled.
        The standard errors are mostly imperceptible, which does not allow a visual discovery of TD-MPC/TD-MPC2 having tighter variability as in \Cref{fig:learning_curve_comparison_rph}.
    }
    \label{fig:learning_curve_comparison_stderr}
\end{figure*}

\begin{figure*}[tbh]
    \centering
    \includegraphics[width=0.75\textwidth]{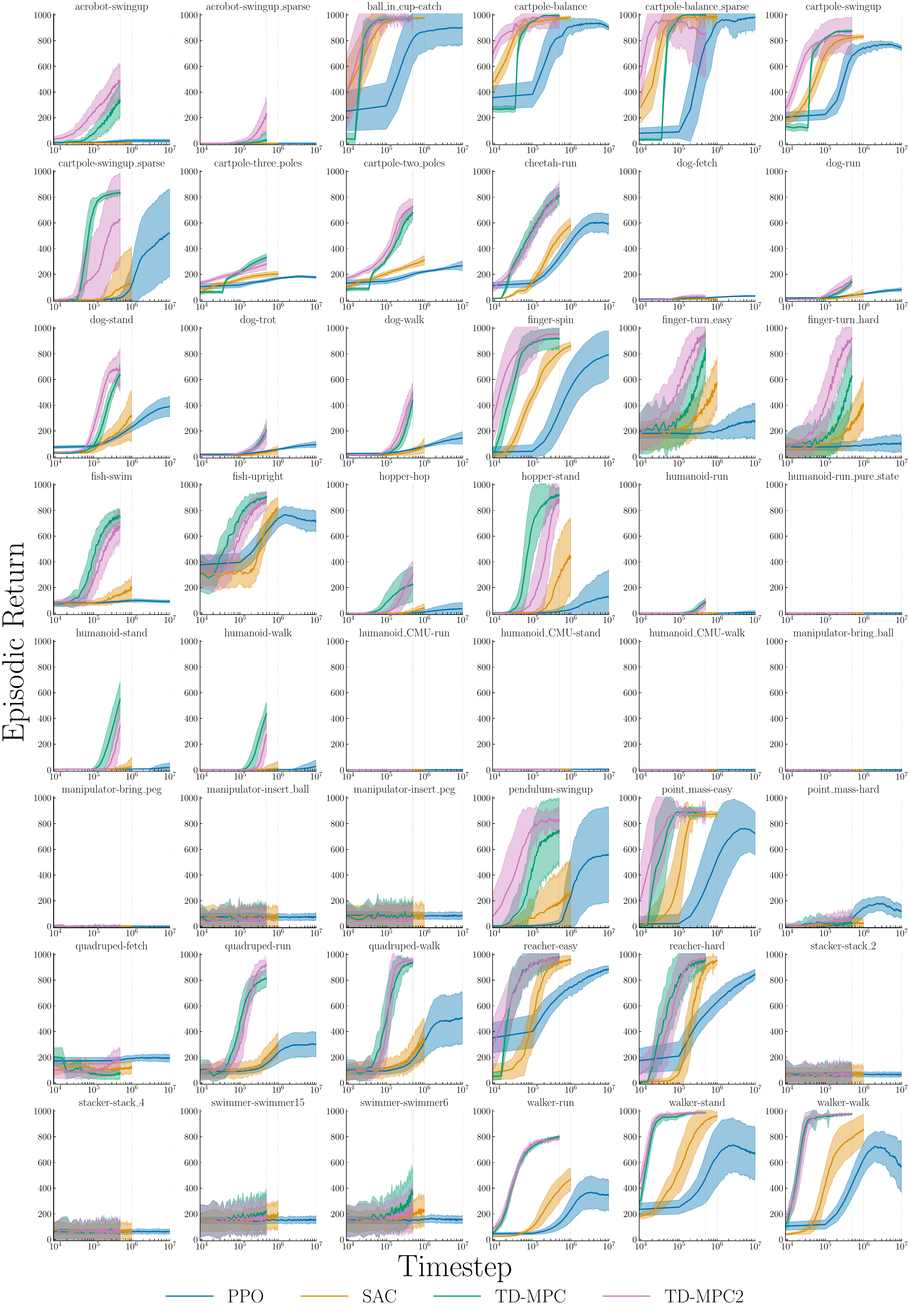}
    \caption{
        Learning curve comparison plot with standard deviation.
        Each subfigure shows the mean and standard deviation of the learning curves of PPO, SAC, TD-MPC, and TD-MPC2 on a given task.
        For visual clarity, the x-axis is log-scaled.
    }
    \label{fig:learning_curve_comparison_std}
\end{figure*}

\FloatBarrier

\newpage
\section{Environment Specification}
\label{sec:appendix:env_spec}

\begin{longtable}[htbp!]{ llcccc }
    \caption{List of tasks in this paper and their state/action dimensions, and min/max episodic returns. Episodic returns of ALE tasks are in human-normalized scale.} \\
    \toprule
    \textbf{Environment} & \textbf{Task}
    & \multicolumn{1}{c}{\(\dim(\mathcal{O})\)}
    & \multicolumn{1}{c}{\(\dim(\mathcal{A})\)}
    & \multicolumn{1}{c}{\textbf{Min.}}
    & \multicolumn{1}{c}{\textbf{Max.}} \\
    \midrule
    \endfirsthead

    \toprule
    \textbf{Environment} & \textbf{Task}
    & \multicolumn{1}{c}{\(\dim(\mathcal{O})\)}
    & \multicolumn{1}{c}{\(\dim(\mathcal{A})\)}
    & \multicolumn{1}{c}{\textbf{Min.}}
    & \multicolumn{1}{c}{\textbf{Max.}} \\
    \midrule
    \endhead
    MuJoCo & \texttt{Ant-v4} & 27 & 8 & -365 & 5787 \\
    MuJoCo & \texttt{HalfCheetah-v4} & 17 & 6 & -529 & 8718 \\
    MuJoCo & \texttt{Hopper-v4} & 11 & 3 & 13 & 3749 \\
    MuJoCo & \texttt{Humanoid-v4} & 376 & 17 & 96 & 6583 \\
    MuJoCo & \texttt{HumanoidStandup-v4} & 376 & 17 & 26492 & 216170\\
    MuJoCo & \texttt{InvertedDoublePendulum-v4} & 11 & 1 & 50 & 9355 \\
    MuJoCo & \texttt{InvertedPendulum-v4} & 4 & 1 & 7 & 1000 \\
    MuJoCo & \texttt{Pusher-v4} & 23 & 7 & -165 & -21 \\
    MuJoCo & \texttt{Reacher-v4} & 11 & 2 & -76 & -4 \\
    MuJoCo & \texttt{Swimmer-v4} & 8 & 2 & -52 & 347 \\
    MuJoCo & \texttt{Walker2d-v4} & 17 & 6 & -1 & 6704 \\
    DMC & \texttt{acrobot-swingup} & 6 & 1 & 0 & 1000 \\
    DMC & \texttt{acrobot-swingup\_sparse} & 6 & 1 & 0 & 1000 \\
    DMC & \texttt{ball\_in\_cup-catch} & 8 & 2 & 0 & 1000 \\
    DMC & \texttt{cartpole-balance} & 5 & 1 & 0 & 1000 \\
    DMC & \texttt{cartpole-balance\_sparse} & 5 & 1 & 0 & 1000 \\
    DMC & \texttt{cartpole-swingup} & 5 & 1 & 0 & 1000 \\
    DMC & \texttt{cartpole-swingup\_sparse} & 5 & 1 & 0 & 1000 \\
    DMC & \texttt{cartpole-three\_poles} & 11 & 1 & 0 & 1000 \\
    DMC & \texttt{cartpole-two\_poles} & 8 & 1 & 0 & 1000 \\
    DMC & \texttt{cheetah-run} & 17 & 6 & 0 & 1000 \\
    DMC & \texttt{dog-fetch} & 232 & 38 & 0 & 1000 \\
    DMC & \texttt{dog-run} & 223 & 38 & 0 & 1000 \\
    DMC & \texttt{dog-stand} & 223 & 38 & 0 & 1000 \\
    DMC & \texttt{dog-trot} & 223 & 38 & 0 & 1000 \\
    DMC & \texttt{dog-walk} & 223 & 38 & 0 & 1000 \\
    DMC & \texttt{finger-spin} & 9 & 2 & 0 & 1000 \\
    DMC & \texttt{finger-turn\_easy} & 12 & 2 & 0 & 1000 \\
    DMC & \texttt{finger-turn\_hard} & 12 & 2 & 0 & 1000 \\
    DMC & \texttt{fish-swim} & 24 & 5 & 0 & 1000 \\
    DMC & \texttt{fish-upright} & 21 & 5 & 0 & 1000 \\
    DMC & \texttt{hopper-hop} & 15 & 4 & 0 & 1000 \\
    DMC & \texttt{hopper-stand} & 15 & 4 & 0 & 1000 \\
    DMC & \texttt{humanoid-run} & 67 & 21 & 0 & 1000 \\
    DMC & \texttt{humanoid-run\_pure\_state} & 55 & 21 & 0 & 1000 \\
    DMC & \texttt{humanoid-stand} & 67 & 21 & 0 & 1000 \\
    DMC & \texttt{humanoid-walk} & 67 & 21 & 0 & 1000 \\
    DMC & \texttt{humanoid\_CMU-run} & 137 & 56 & 0 & 1000 \\
    DMC & \texttt{humanoid\_CMU-stand} & 137 & 56 & 0 & 1000 \\
    DMC & \texttt{humanoid\_CMU-walk} & 137 & 56 & 0 & 1000 \\
    DMC & \texttt{manipulator-bring\_ball} & 44 & 5 & 0 & 1000 \\
    DMC & \texttt{manipulator-bring\_peg} & 44 & 5 & 0 & 1000 \\
    DMC & \texttt{manipulator-insert\_ball} & 44 & 5 & 0 & 1000 \\
    DMC & \texttt{manipulator-insert\_peg} & 44 & 5 & 0 & 1000 \\
    DMC & \texttt{pendulum-swingup} & 3 & 1 & 0 & 1000 \\
    DMC & \texttt{point\_mass-easy} & 4 & 2 & 0 & 1000 \\
    DMC & \texttt{point\_mass-hard} & 4 & 2 & 0 & 1000 \\
    DMC & \texttt{quadruped-fetch} & 90 & 12 & 0 & 1000 \\
    DMC & \texttt{quadruped-run} & 78 & 12 & 0 & 1000 \\
    DMC & \texttt{quadruped-walk} & 78 & 12 & 0 & 1000 \\
    DMC & \texttt{reacher-easy} & 6 & 2 & 0 & 1000 \\
    DMC & \texttt{reacher-hard} & 6 & 2 & 0 & 1000 \\
    DMC & \texttt{stacker-stack\_2} & 49 & 5 & 0 & 1000 \\
    DMC & \texttt{stacker-stack\_4} & 63 & 5 & 0 & 1000 \\
    DMC & \texttt{swimmer-swimmer15} & 61 & 14 & 0 & 1000 \\
    DMC & \texttt{swimmer-swimmer6} & 25 & 5 & 0 & 1000 \\
    DMC & \texttt{walker-run} & 24 & 6 & 0 & 1000 \\
    DMC & \texttt{walker-stand} & 24 & 6 & 0 & 1000 \\
    DMC & \texttt{walker-walk} & 24 & 6 & 0 & 1000 \\
    ALE & \texttt{BattleZone-v5} & (84, 84, 1) & 18 & -0.01 & 1.47 \\
    ALE & \texttt{DoubleDunk-v5} & (84, 84, 1) & 18 & -2.53 & 9.68 \\
    ALE & \texttt{NameThisGame-v5} & (84, 84, 1) & 18 & -0.29 & 2.26 \\
    ALE & \texttt{Phoenix-v5} & (84, 84, 1) & 18 & -0.11 & 4.41 \\
    ALE & \texttt{Qbert-v5} & (84, 84, 1) & 18 & -0.01 & 1.45 \\
    \bottomrule
    \label{table:rl_tasks}
\end{longtable}

\end{document}